\def\ARXIV{1}

\documentclass[12pt]{article}
\usepackage[margin=1in]{geometry}

\usepackage{microtype}
\usepackage{graphicx}
\usepackage{subfigure}
\usepackage{booktabs} % for professional tables
\usepackage{tcolorbox} % for professional tables
\usepackage{natbib}

\usepackage{hyperref}

\ifnum\ARXIV=0
\usepackage{minitoc}
\else
\usepackage[tight]{minitoc}
\fi

\usepackage{amsmath,amsthm,amsfonts,nicefrac,amssymb,bm}

\usepackage[capitalize]{cleveref}

\usepackage{morenotations2}

\ifnum\ARXIV=0
\usepackage{minitoc}
\else
\usepackage[tight]{minitoc}
\fi

\usepackage[toc,page,header]{appendix}

\usepackage{cleveref}
\usepackage{rotating}

\renewcommand\cite[1]{\citep{#1}}

\begin{document}

\doparttoc
\faketableofcontents




% TODO
%\begin{icmlauthorlist}
%\icmlauthor{}{anu}
%\icmlauthor{}{goor}
%\icmlauthor{Sanmi Koyejo}{ill}
%\icmlauthor{}{goor,tel}
%\icmlauthor{}{goor}
%\icmlauthor{Richard Nock}{goor,anu}
%\icmlauthor{}{d61,anu}
%\icmlauthor{}{anu}
%\end{icmlauthorlist}

\title{\papertitle}
\author{
    Alexander Soen\textsuperscript{\dag} \quad
    Ibrahim Alabdulmohsin\textsuperscript{\ddag} \quad
    Sanmi Koyejo\textsuperscript{\textasteriskcentered, \ddag} \\
    Yishay Mansour\textsuperscript{\ddag, $\diamond$} \quad
    Nyalleng Moorosi\textsuperscript{\ddag} \quad
    Richard Nock\textsuperscript{\ddag, \dag} \\
    Ke Sun\textsuperscript{\dag, $\circ$} \quad
    Lexing Xie\textsuperscript{\dag} \\
    \\[1pt]
    Australian National University\textsuperscript{\dag} \quad 
    Google Research\textsuperscript{\ddag} \\
    Stanford University\textsuperscript{\textasteriskcentered} \quad 
    Tel Aviv University\textsuperscript{$\diamond$} \\
    Data61/CSIRO\textsuperscript{$\circ$}
}

\date{}

\maketitle




\begin{abstract}
We introduce a new family of techniques to post-process (``wrap") a black-box classifier in order to reduce its bias. Our technique builds on the recent analysis of improper loss functions whose optimization can correct any \textit{twist} in prediction, unfairness being treated as a twist. In the post-processing, we learn a wrapper function which we define as an $\alpha$-tree, which modifies the prediction. We provide two generic boosting algorithms to learn $\alpha$-trees. We show that our modification has appealing properties in terms of composition of $\alpha$-trees, generalization, interpretability, and KL divergence between modified and original predictions. We exemplify the use of our technique in three fairness notions: conditional value-at-risk, equality of opportunity, and statistical parity; and provide experiments on several readily available datasets.
\end{abstract}
\section{Introduction}
\label{introduction}

% Add "black-box" into the first para
The social impact of Machine Learning (ML) has seen a dramatic increase over the past decade -- enough so that the bias of model outputs must be accounted for alongside accuracy \citep{alAN,hpsEO,zvggFC}. Considering the various number of fairness targets \citep{mmslgAS} and the energy and CO2 footprint of ML \citep{mSD,sgmEA}, the combinatorics of training accurate \emph{and} fair models is non-trivial. This is especially so given the inherit incompatibilities of fairness constraints~\citep{kleinberg2017inherent} and the underlying tension of satisfying fairness whilst maintaining accuracy. One trend in the field "decouples" the two constraints by post-processing \emph{pretrained} (accurate) models to achieve fairer outputs~\citep{zvggFC}. Post-processing may be the only option if we have no access to the model's training data / algorithm / hyperparameters (etc.).

\newcommand{\des}[1]{\textbf{(#1)}}
Within the post-processing approach, three trends have emerged: learning a new fair model close to the black-box, tweaking the output subject to fairness constraints, and exploiting sets of classifiers. % ($\S$ \ref{sec-rel}).
If the task is class probability estimation \citep{rwID}, the estimated black-box is an accurate but potentially unfair posterior $\posunfair : \mathcal{X} \rightarrow [0,1]$ which neither can be opened nor trained further. The goal is then to learn a fair posterior $\posfair$ from it. In addition to the black-box constraint, a number of \emph{desiderata} can be considered for post-processing.
Ideally in correcting a black-box, we would want the approach to have \des{flexibility} in satisfying substantially different fairness criteria, \des{proximity} of the learnt \( \posfair \) to the original \( \posunfair \), and meaningful \des{composability} properties if, \emph{e.g.}, $\posfair$ was later treated as a new black-box to be post-processed. To facilitate a specific style of correction, we may also want the representation of the correction to facilitate \des{explainability} for auditing the post-processing procedure and bounds on the increased model \des{complexity} of the final classifier \( \posfair \). In training the correction, algorithmically we would also want guarantees for \des{convergence}.

\begin{figure}
    \centering
    \includegraphics[trim=35bp 602bp 505bp 61bp,clip,width=\textwidth]{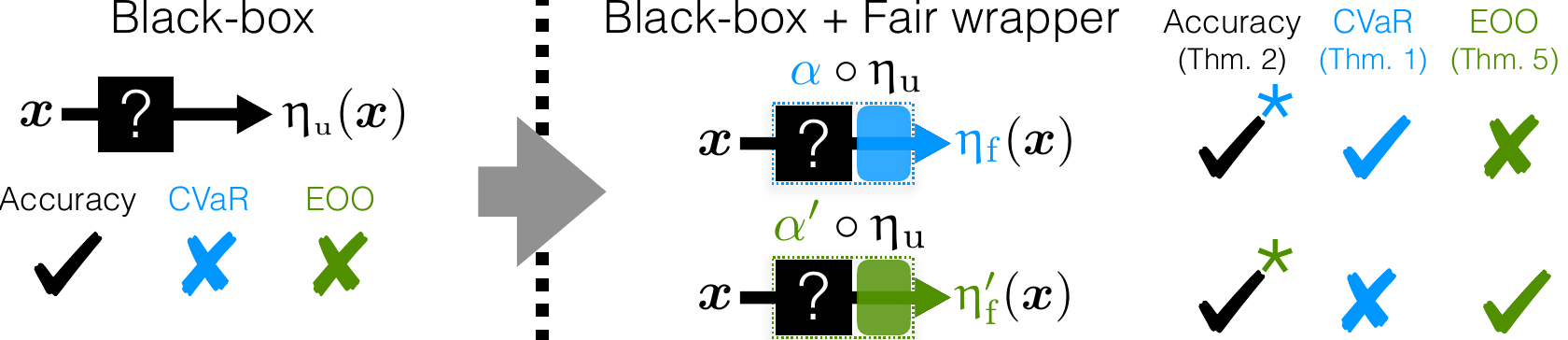}%
    \caption{Summary of using different \( \alpha \)-correction wrappers to obtain different fairness criteria guarantees. See \cref{sec-formal,sec-fairness} for full details on guarantees.}
    \label{fig:teaser}
    \fnegativespace
\end{figure}

\textbf{Our contribution} satisfies the aforementioned desiderata in its correction, representation, and algorithmic guarantees.
By leveraging recent theory in \emph{im}proper loss functions, we utilize a \emph{universal} correction of black-box posteriors defined by the \( \alpha \)-loss.
%by \emph{untwisting} the black-box.
%By treating unfairness as a twist, we utilize recently introduced theory to define a correction of black-box posteriors through the minimization of specific \emph{im}proper losses.
This allows for a flexible correction which yields convenient divergence bounds between $\posfair$ and $\posunfair$, a convenient form for the Rademacher complexity of the class of $\posfair$, and a simple composability property. Representation-wise, the corrections we learn are easy-to-interpret tree-shaped functions that we define as \textit{$\alpha$-trees}. Algorithmically speaking, we provide two formal boosting algorithms to learn $\alpha$-trees building upon seminal results \citep{kmOT}.
We demonstrate our algorithm for conditional value-at-risk (\(\cvar{}\)) \citep{wmFR}, equality of opportunity (EOO), and statistical parity; as depicted in \cref{fig:teaser}. Experiments are provided against five baselines on readily available datasets. All proofs and more experiments are in an Appendix denoted as \supplement.
%To exemplify the algorithm, we first present an ``instantiation'' of our approach which considers the conditional value-at-risk fairness criteria. We then specify the components of the algorithm which can be ``swapped'' to satisfy other fairness criteria. In total, we demonstrate our algorithm for conditional value-at-risk, equality of opportunity, and statistical parity; as depicted in \cref{fig:teaser}. Experiments are provided against five baselines on readily available datasets. All proofs and many more experiments are in an Appendix denoted \supplement.

\section{Related Work}
\label{sec-rel}
Post-processing models to achieve fairness is one of three different categories in tackling the ML + fairness challenge \citep[Section 6.2]{zvggFC}. Although other notions exist, \emph{e.g.} individual fairness \citep{dwork2012fairness}, we limit our analysis to group fairness, which concerns itself with ensuring that statistics of sub-populations are similar.
We further segment this cluster into three subsets: (I) approaches learning a new model with two constraints: being close to the pretrained model and being fair \citep{kgzMB,pmsyPP,wrcOS,yang2020fairness}; (II) approaches biasing the output of the pretrained model at classification time, modifying observations for fairer outcomes \citep{alAN,hpsEO,lrbsvpBM,mwTC,wgosLN,yang2020fairness}; and (III) techniques consisting of exploiting sets of models to achieve fairness \citep{diklDC}. None of these approaches formulates substantial guarantees on all of the desiderata in the introduction. Some bring contributions with the \des{flexibility} of being applicable to more than two fairness notions \citep{cpfghAD,wrcOS,diklDC,yang2020fairness}. Two of which provide the convenience of analytic conditions on new fairness notions to fit in the approach \citep{wrcOS,diklDC}. However, for all of them, the algorithmic price-tag is unclear \citep{cpfghAD,diklDC} or heavily depends on convex optimization routines \citep{wrcOS}. \citet{alAN,yang2020fairness} provide strong guarantees regarding \textbf{(proximity)}, w.r.t. \textit{consistency and generalization}. To our knowledge, our approach of correcting prediction unfairness through improper losses \citep{nssBP} is new.

%\todo
%{TODO: insert a table summarizing these? columns: diserderatas, rows: key clusters/work above}

%\cite{lrbsvpBM}: 1 fairness model, no formal guarantees on b, c, e, f. d unclear

%\cite{pmsyPP}: 1 general fairness model on Lipschitz continuity, no formal guarantees on b, c, d, e, f. d linked to graph laplacian

%\cite{wrcOS}: general to fairness constraints expressible as linear constraints (SP, equalized odds), guarantees on a, b. c linked to ADMM / convex optimisation. d depends on model trained. no formal guarantees on e, f.

%\cite{kgzMB}: 1 fairness model, guarantees on b. no formal guarantees on c, e, f. d unclear.

%\cite{hpsEO}: 1 fairness model, guarantees on b. a, d unclear. no guarantees on c, e, f.

%\cite{alAN}: 1 fairness model (+tweak for a second one), guarantees on b, c, f. no guarantees on e. d unclear.

%\cite{diklDC}: general to fairness constraints meeting a specific loss formulation (demographic parity). guarantees on b, f. no guarantees on c, e. d unclear

%\cite{mwTC}: 2 fairness models. guarantees on b. c linked to plugin algorithm. no guarantees on e, f. d unclear

%\cite{wgosLN}: 1 fairness model. shows limits on \cite{hpsEO} -- pb with constraints, size of hypothesis space for black box, loss minimized (if not strictly convex). Strong guarantees on b, f. c, d, e, irrelevant

%\cite{cpfghAD}: 3 fairness models. guarantees on b. no guarantee on c, e, f. d unclear
%
\section{Setting and Motivating Example}
\label{sec-example}
%
%\todo{Minimal notation to introduce CVaR and a working alpha-tree}
%
Let $\mathcal{X}$ be a domain of observations, $\mathcal{Y} \defeq \{-1,1\}$ be labels and $S$ is a sensitive attribute in $\mathcal{X}$. We assume that the modalities of $S$ induce a partition of $\mathcal{X}$. We further let \( \meas{D} \) denote the joint measure over \( \mathcal{X} \times \mathcal{Y} \), \( \meas{M} \) denote the marginal measure over \( \mathcal{X} \), and \( \prior \defeq \pr[\Y = 1] \) being the prior. We denote conditioning of \( \meas{M} \) through a subscript, \emph{e.g.}, \( \meas{M}_{s} \) for \( s \in S \) denotes \( \meas{M} \) conditioned on a sensitive attribute subgroup \( \SSS = s \). We leave the \( \sigma \)-algebra to be implicit (which is assumed to be the same everywhere). 
As is often assumed in ML,
%We assume that
sampling is i.i.d.; we make no notational distinction between empirical and true measures to simplify exposition -- most of our results apply for both.% This setup can be contextualized as a \emph{binary task} \citep[Section 4]{rwID}, see the \supplement.

\begin{figure}[!t]
    \centering
    \includegraphics[trim=0pt 528pt 217pt 0pt,clip,width=\textwidth]{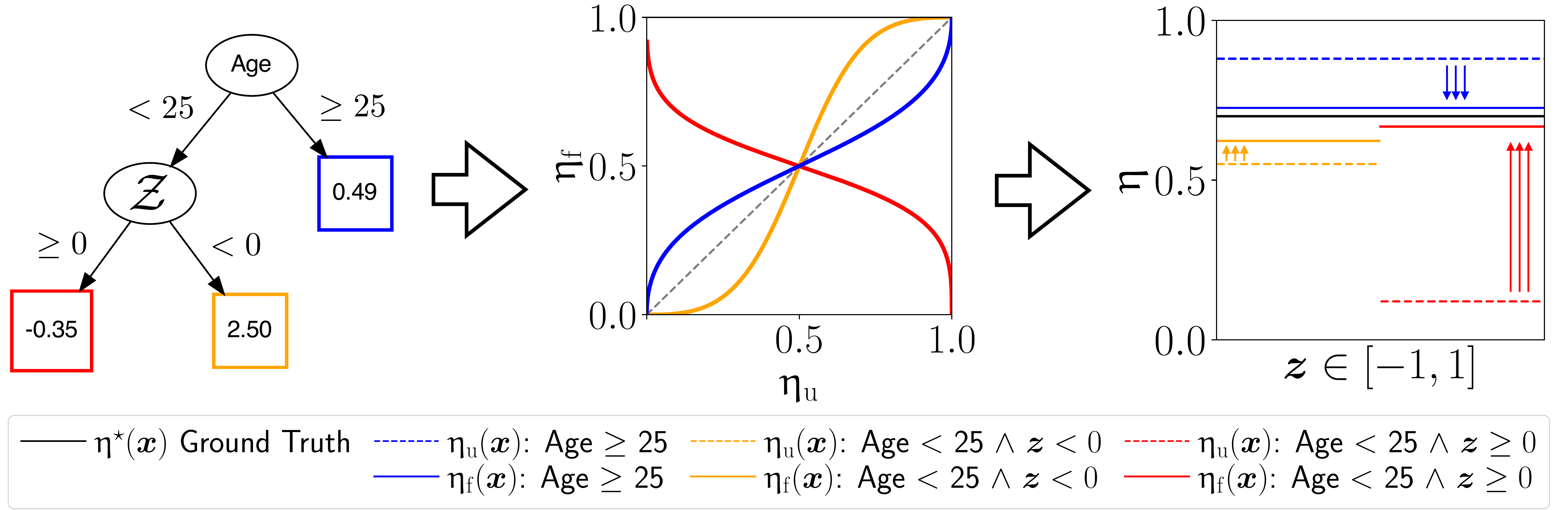}%
    \caption{%
        Improving \( \cvar{} \) for a toy hiring task with \(\alpha\)-trees.
        %Using \( \alpha \)-trees to improve \( \cvar{} \) for a toy hiring dataset.
        An \( \alpha\)-tree (left) transforms the posterior via \eqref{ourposfair} (middle);
        resulting in an input-dependent fairness correction of the posterior \( \posunfair\) (right).
%         correction of posterior values to improve the fairness criteria (right).
    }
    \label{fig:alpha-trees}
    \fnegativespace
\end{figure}

%Following class probability estimation (CPE), we aim to find a function \( \posterior \in [0,1]^{\mathcal{X}} \) to estimate the Bayes posterior \( \bayespo(\ve{x}) = \pr[\Y=1 | \X = \ve{x}] \).  Typically in binary classification, to fit a posterior \( \posterior \), a model aims to minimize the \emph{(total) risk}
Consider the task of learning a function \( \posterior \in [0,1]^{\mathcal{X}} \) to estimate the true posterior \( \bayespo(\ve{x}) = \pr[\Y=1 | \X = \ve{x}] \) in binary classification.
For instance, we may want to predict the probability of hiring an applicant for a company.
%We assume that \( \posterior(.) \) is learned through empirical risk minimization.
Given the pointwise loss \( \poirisk(\posterior(\X),\bayespo(\X)) \), which determines the loss of a single example, the \emph{(total) risk} is defined as 
%We note that in most circumstances \( \bayespo(.) \) can be readily approximated with empirical approximations. In such a task, the estimator \( \posterior \) is learned by minimizing the \emph{(total) risk} of a loss
%
\begin{eqnarray}\label{eq:risk_unfair}
    \totrisk(\posterior; \meas{M}, \bayespo) & \defeq &  \expect_{\X \sim \meas{M}} \left[\poirisk(\posterior(\X),\bayespo(\X))\right]
\end{eqnarray}
%where \( \poirisk(\posterior(\X),\bayespo(\X)) \) determines the loss of a single example 
(with slight abuse of notation). A low risk corresponds to good classification performance. In this paper, we consider the risk determined by the \emph{log-loss}:
\begin{equation}
  \totrisk(\posterior; \meas{M}, \bayespo)
  \defeq
  \mathop{\expect}_{\X \sim \meas{M}}
  \left[ 
  \bayespo(\X)\cdot - \log \posterior(\X) + (1-\bayespo(\X))\cdot - \log (1-\posterior(\X)) \right].\label{lossgen}
\end{equation}
%
% For examples, one can take \( \poirisk(u, v) = - v \cdot \log u - (1-v) \cdot \log (1-u) \) to give the log-loss. % (as we will do).%  In our work, we take \( L \) to be the log-loss.
%
%
We now consider a simple fairness problem, centered around the example of \cref{fig:alpha-trees}. %, with an example centered around \cref{fig:alpha-trees}. Suppose that 
Suppose that we are given a black-box \( \posunfair \) which predicts hiring probabilities without considering fairness. Although the minimized total risk of \eqref{eq:risk_unfair} might be small, there can be discrepancies in the performance between different subgroups.
Instead of considering the total risk, the predictive performance of specific subgroups can be examined through the \emph{subgroup risk} \( \totrisk(\posunfair; \meas{M}_{s}, \bayespo) \), for a subgroup \( s \in S \). For instance, we might want to examine the discrepancy of subgroup risks among the age of applications. A natural fairness task would be to improve the worst performing subgroup, say \( s_{\mathrm{w}} \in S \). 
%Then in our hiring task example, by picking age groups as our subgroups (those above and below \( 25 \) years old for simplicity), we want to improve the worst performing age subgroup \( \totrisk(\posunfair; \meas{M}_{s_{\mathrm{w}}}, \bayespo) \).

%Thus, for a black-box \( \posunfair \) which is learned without considering fairness, we want to improve \( \totrisk(\posunfair; \meas{M}_{s_{\mathrm{w}}}, \bayespo) \).

%To motivate our approach, we consider the simple fairness problem of improving the worse performing subgroup of an unfair black-box, say \( s_{\mathrm{w}} \in S \). The performance of each subgroup can simply be defined as the \emph{subgroup risk} \( \totrisk(\posunfair; \meas{M}_{s}, \bayespo) \), for \( s \in S \).
%
%Suppose that \( \posunfair \), our black-box, is a posterior learned by minimizing \eqref{eq:risk_unfair} without considering the fairness problem.

To post-process unfairness, we want to learn a function \( \alpha: \mathcal{X} \to \mathbb{R}\) which ``wraps'' \( \posunfair \) and \emph{lowers} the worst subgroup risk.
% of \( \posunfair \).
We propose the following ``wrapping'', inspired by improper loss functions \citep{nssBP}
%
%Now consider a simplified fairness problem where we wish to improve performance (decreasing subgroup risk) of the worst performing (highest subgroup risk) subgroup of \( \posunfair \), say \( s_{\mathrm{w}} \). In the post-processing setting, we wish to find a mapping \( \posunfair \mapsto \posfair \) with \( \totrisk(\posunfair; \meas{M}_{s_{\mathrm{w}}}, \bayespo) > \totrisk(\posfair; \meas{M}_{s_{\mathrm{w}}}, \bayespo) \). Our proposed correction comes from the pointwise minimizer of the improper \emph{\( \alpha \)-loss} \citep{nssBP}:
%
\begin{eqnarray}
\posfair(\ve{x}) & \defeq & \frac{\posunfair(\ve{x})^{\alpha(\ve{x})}}{\posunfair(\ve{x})^{\alpha(\ve{x})}+(1-\posunfair(\ve{x}))^{\alpha(\ve{x})}} \; \in [0,1].\label{ourposfair}
\end{eqnarray}
%where the function \( \alpha(.) \) is to be learned.
%
Notice when \( \alpha(\ve{x}) = 1 \) the resulting posterior is the original \( \posunfair(\ve{x}) \). 
%
%A key proper of this correction is its \emph{twist-properness}, which is an universality property of the mapping: by fixing \( \ve{x} \) in \eqref{ourposfair}, an appropriately chosen \( \alpha \)-value can map any \( \posunfair(\ve{x}) \) to any value in \( [0, 1] \). Thus an appropriately learned \( \alpha(.) \) can post-process posterior \( \posunfair \) to become more fair.
%Importantly, \eqref{ourposfair} is flexible enough to correct any potential unfairness in \( \posunfair \).
Importantly, \eqref{ourposfair} is flexible enough to transform any input black-box \( \posunfair \) to any needed \( \posfair \). Looking at \cref{fig:alpha-trees} (left and middle), intuitively by setting different \( \alpha(\ve{x}) \) values, \eqref{ourposfair} {\color{orange} ``sharpens''} (yellow, \( \alpha > 1\)), {\color{blue}``dampens''} (blue, \( 0 < \alpha < 1 \)), or {\color{red}``polarity reverses''} (red, \( \alpha < 0 \)) the original posterior \( \posunfair \).
To improve fairness, we need a combination of these corrections to accommodate different subsets of the input domain (thus learning \( \alpha(.) \) as a function).
We specifically learn \( \alpha(.) \) to be a tree structure, which allows an interpretable correction alongside other formal properties (\cref{sec-formal}).
%for a boosting compliant boosting algorithm and other formal properties~\cref{sec-formal}. %\todo{why a tree?, also include forward pointer to the next Sec}
\begin{definition}\label{def-tree}
  An \textbf{$\alpha$-tree} is a rooted, directed binary tree, with internal nodes labeled with observation variables. Outgoing arcs are labeled with tests over the nodes' variable. Leaves are real valued. $\leafset(\alphatree)$ is the leafset of $\alpha$-tree $\alphatree$. An \( \alpha\)-tree induces a correction over posteriors as per \eqref{ourposfair} with \( \alpha = \alphatree \).
\end{definition}

Our fairness problem is now learning an \( \alpha \)-tree \( \posterior \) which provides a corresponding correction that improves the worst subgroup risk, \emph{i.e.}, \( \totrisk(\posunfair; \meas{M}_{s_{\mathrm{w}}}, \bayespo) > \totrisk(\posfair; \meas{M}_{s_{\mathrm{w}}}, \bayespo) \). The entirety of \cref{fig:alpha-trees} presents such a process. In this hiring task example, the ground truth hiring rate is constant \emph{w.r.t.} the inputs, \( \bayespo(\ve{x}) = 0.7 \). Despite this, the black-box \( \posunfair(.) \) unfairly depends on the age of applicants and incorrectly depends on a noise feature \( \mathcal{Z} \). By choosing \( \alpha(\ve{x}) \) as per \cref{fig:alpha-trees} (left), the correction
%``sharpens'', ``dampens'', and ``polarity reverses'' 
changes
\( \posunfair \) to be closer to \( \bayespo \), improving the risk of the worst subgroup (alongside the other subgroup in this example): the worse-case loss improves from \( 1.09 \) to \( 0.62 \).

Although the fairness criteria discussed might be considered simple, the procedure of iteratively minimizing the worse performing subgroup can be used to improve the \emph{conditional value-at-risk} (lower is better) fairness criteria \citep{wmFR}:
\begin{eqnarray} \label{eq:cvar}
    %\cvar{\beta}(\posfair) &\defeq& {\expect}_{\SSS \sim \meas{M}_S} [\totrisk(\posfair; \meas{M}_{\SSS}, \bayespo) \vert \totrisk(\posfair; \meas{M}_{\SSS}, \bayespo)  \geq L_\beta],
    \cvar{\beta}(\posfair) &\defeq& {\expect}_{\SSS} [\totrisk(\posfair; \meas{M}_{\SSS}, \bayespo) \mid \totrisk(\posfair; \meas{M}_{\SSS}, \bayespo)  \geq L_\beta],
\end{eqnarray}
where $L_\beta$ is the risk value for the $\beta$ quantile among subgroups, which is user-defined.
The difference is that \( \cvar{\beta} \) not only considers the worse case subgroup, but all subgroups above the \( L_{\beta} \) risk value (let these subgroups be \(\mathcal{S}_{\beta}\)). One can simply iteratively improve all \( s \in \mathcal{S}_{\beta} \), as done in the example of \cref{fig:alpha-trees}. Indeed, in the example, \( \cvar{\beta} \) with \( \beta = 0.9 \) improves from \( 1.09 \) to \( 0.62 \) (which is equivalent to worse-case loss in this case).

\begin{figure}[!t]
\begin{minipage}{0.55\linewidth}
  \vspace{-\fboxsep}
  \begin{algorithm}[H]
\caption{\topdown($\meas{M}_{\mbox{\tiny{\textup{t}}}},\postarget, \alphatree_0, B$)}\label{alg:topdown}
\begin{algorithmic}
  \STATE  \textbf{Input} mixture $\meas{M}_{\mbox{\tiny{\textup{t}}}}$, posterior $\postarget$, $\alpha$-tree $\alphatree_0$, $B\in \mathbb{R}_{+*}$;
  \STATE Step 1: $\alphatree \leftarrow \alphatree_0$;
  \STATE Step 2 : \textbf{while} stopping condition not met \textbf{do}
  \STATE  \hspace{0.2cm} Step 2.1 : pick leaf $\leaf^\star \in \leafset(\alphatree)$;  // \emph{i.e.} heaviest leaf
  \STATE  \hspace{0.2cm} Step 2.2 : $h^\star \leftarrow \arg\min_{h\in \mathcal{H}} \entropy(\alphatree(\leaf^\star, h); \meas{M}_{\mbox{\tiny{\textup{t}}}}, \postarget)$;
 \STATE  \hspace{0.2cm} Step 2.3 :  $\alphatree \leftarrow \alphatree(\leaf^\star, h^\star)$; // split using $h^\star$ at $\leaf^\star$
 \STATE  Step 3 : label leaves \( \forall \leaf \in \leafset(\alphatree) \):
 \negativespace
 \begin{eqnarray}
    \alphatree(\leaf) &\defeq& \normalizedlogit\left(\frac{1+\edge(\meas{M}_{\leaf}, \postarget)}{2}\right), \;\; \textrm{// \(\alpha\)-value}\label{alphatreeoutputCONS}
  \end{eqnarray}%
 \STATE \textbf{Output} $\alphatree$;
\end{algorithmic}
\end{algorithm}
\end{minipage}%
\hfill%
\begin{minipage}{0.4\linewidth}
\begin{figure}[H]
  \centering
  \includegraphics[trim=25bp 520bp 450bp 25bp,clip,width=\textwidth]{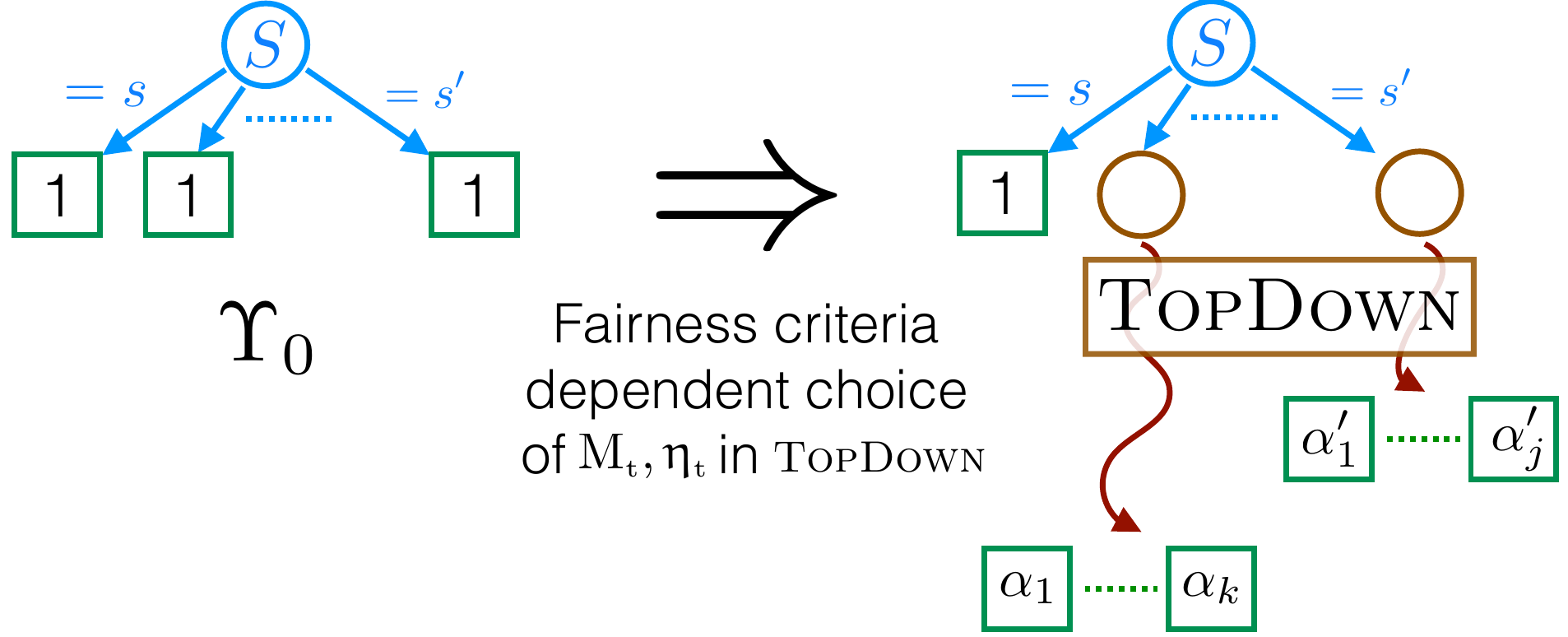}%
    \caption{Picking $\alphatree_0$ a stump on the fairness attribute allows to finely tune growths of sub-$\alpha$-trees to the fairness criterion at hand.}
    \label{fig:bigalpha}
    \vspace{-0.5cm}
\end{figure}
\end{minipage}%  
    \fnegativespace
\end{figure}
\section{Growing Alpha-Trees}%
\label{sec:growing_alpha_trees}%
We now introduce the procedure to grow an \( \alpha \)-tree via a boosting algorithm \topdown, \cref{alg:topdown}. \topdown~can be thought of as a generalization of the standard decision tree induction used for classification \citep{kmOT}.
We first introduce relevant concepts from decision tree induction to explain \topdown. We contextualize \topdown through its application in improving the \( \cvar{} \) criteria.%, as per the prior section.
%
%To introduce our approach, we return to the motivating example of improving \( \cvar{} \) by minimizing the worse performing subgroup's risk \( \totrisk(\posfair; \meas{M}_{s_{\mathrm{w}}}, \bayespo) \). 

We first introduce a technical assumption for the black-box \( \posunfair \) to be post-processed:
\begin{assumption}\label{assumptionBUNF}
  The black-box prediction is bounded away from the extremes: $\exists B>0$ such that
  \begin{eqnarray}
    \mathrm{Im} (\posunfair) \subseteq \mathbb{I} & \defeq & \left[({1+\exp (B)})^{-1}, ({1+\exp (-B)})^{-1}\right] \quad \mbox{(a.s.)} \label{eqclipSUP}.
    %\mathrm{Im} (\posunfair) \subseteq \mathbb{I} & \defeq & \left[{1}\big/({1+\exp (B)}), {1}\big/({1+\exp (-B)})\right] \quad \mbox{(a.s.)} \label{eqclipSUP}.
  \end{eqnarray}
\end{assumption}
Compliance with Assumption \ref{assumptionBUNF} can be done by clipping the black-box's output with a user-fixed $B$ or making sure it is calibrated and then finding $B$.

\csubsec{Entropy-based updates for fairness}
An important component in standard decision tree induction is the \emph{edge} function, which measures the \emph{label purity} (proportion of positive examples) of a decision tree node. We introduce a generalization which considers the \emph{alignment purity} of a black-box.
\begin{definition}\label{def:edge}
    Let $\logit(u) \defeq \log(u/(1-u))$ the logit of $u \in (0,1)$ and $\normalizedlogit(u) \defeq \logit(u)/B$ a normalization which satisfies $\normalizedlogit(\mathbb{I}) = [-1,1]$.
    %The edge of the normalized logit given $\meas{M}$ and $\postarget$ is defined as,
    The \textbf{alignment edge} of $\meas{M}_{\mbox{\tiny{\textup{t}}}}$ and $\postarget$ is defined as,
    \begin{eqnarray}
        \edge(\meas{M}_{\mbox{\tiny{\textup{t}}}}, \postarget) & \defeq & \expect_{(\X, \Y) \sim \measdtarget} \left[\Y \normalizedlogit(\posunfair(\X))\right],\label{defEDGEGEN}
    \end{eqnarray}%
    where \( \measdtarget \) is the joint measure induced by \( \meas{M}_{\mbox{\tiny{\textup{t}}}} \) and \( \postarget \).
    With Assumption~\ref{assumptionBUNF}, $\edge(\meas{M}_{\mbox{\tiny{\textup{t}}}}, \postarget) \in [-1,1]$.
%    This edge satisfies $\edge(\meas{M}, \postarget) \in [-1,1]$ when Assumption \ref{assumptionBUNF} is satisfied.
\end{definition}

By replacing the normalized logit \( \normalizedlogit \) with a constant \( 1 \), \eqref{defEDGEGEN} reduces to a measure of label purity used in regular classification. In our case, \eqref{defEDGEGEN} measures how well the black-box \( \posunfair \) ``aligns'' with the true labels \( \Y \) through the logit. This also takes into account the ``confidence'' of the black-box's predictions: for a high alignment purity, predictions not only need be correct but also to be highly confident (\( \posunfair \) close to the endpoints of \(\mathbb{I}\)).
Similar to how the splits of a decision tree classifier are determined by the entropy of a tree's label purity, an \( \alpha \)-tree splits based on its alignment entropy.

\begin{definition}\label{defENTROPYA}
    Given an $\alpha$-tree $\alphatree$ with leafset $\leafset$, when Assumption \ref{assumptionBUNF} is satisfied, the \textbf{entropy of} $\alphatree$ is:
    \begin{eqnarray}
        \entropy(\alphatree; \meas{M}_{\mbox{\tiny{\textup{t}}}}, \postarget) & \defeq & \expect_{\leaf \sim \meas{M}_{\leafset(\alphatree)}} \left[\entropy_1(\leaf; \meas{M}_{\mbox{\tiny{\textup{t}}}}, \postarget)\right],\label{defBINENT}
    \end{eqnarray}%
    where $\entropy(q) \defeq -q \log(q) - (1-q)\log(1-q)$,
    $\entropy_1(\leaf; \meas{M}_{\mbox{\tiny{\textup{t}}}}, \postarget) \defeq \entropy\left((1+\edge(\meas{M}_\leaf, \postarget))/2\right)$, $\meas{M}_\leaf$ is $\meas{M}_{\mbox{\tiny{\textup{t}}}}$ conditioned to leaf $\leaf \in \leafset(\alphatree)$, and $\meas{M}_{\leafset(\alphatree)}$ is a measure induced on $\leafset(\alphatree)$ by leaf weights on $\meas{M}_{\mbox{\tiny{\textup{t}}}}$.
    %$\entropy_1(\leaf; \meas{M}_{\mbox{\tiny{\textup{t}}}}, \postarget) \defeq \entropy\left((1+\edge(\meas{M}_\leaf, \postarget))/2\right)$, $\meas{M}_\leaf$ is $\meas{M}_{\mbox{\tiny{\textup{t}}}}$ conditioned to leaf $\leaf \in \leafset(\alphatree)$, and $\meas{M}_{\leafset(\alphatree)}$ is a measure induced on $\leafset(\alphatree)$ by the leaves' weights on $\meas{M}_{\mbox{\tiny{\textup{t}}}}$.
\end{definition}

%\eqref{defENTROPYA} directly leads to an upperbound of a risk value, which for \( \cvar{} \) can be used to upperbound subgroup risks.

\begin{theorem}\label{thm:entropyupper}
    For any \( \meas{M}_{\mbox{\tiny{\textup{t}}}}, \postarget \),
    let \( \alphatree \)'s leaves follow  \eqref{alphatreeoutputCONS}. Then 
    %If the \( \alpha \)-tree leaves follow \eqref{alphatreeoutputCONS}, then
    %
    \( \totrisk(\posfair; \meas{M}_{\mbox{\tiny{\textup{t}}}}, \postarget) \leq  \entropy(\alphatree; \meas{M}_{\mbox{\tiny{\textup{t}}}}, \postarget) \).
    %holds for \( \alphatree \) and its corresponding \( \posfair \),
%    \begin{eqnarray}
%        \totrisk(\posfair; \meas{M}, \postarget) & \leq  & \entropy(\alphatree; \meas{M}, \postarget). \label{totriskB1}
%    \end{eqnarray}
\end{theorem}

\cref{alg:topdown} can now be explained by repeatedly leveraging \cref{thm:entropyupper}.
Suppose that we have a hypothesis set of possible splits \( \mathcal{H} \) to grow our \( \alpha \)-tree. Denote \( \alphatree(\leaf, h) \) as the \( \alpha \)-tree \( \alphatree \) split at leaf \( \leaf \in \leafset(\alphatree) \) using test \( h \). The inner loop within Step 2 is the process of finding the best possible leaf splits 
which helps to minimize the \( \alpha \)-tree's entropy and to reduce
%which minimize the \( \alpha \)-tree's entropy, which in turn reduces
the risk as per \cref{thm:entropyupper}. The \( \alpha \)-values of \eqref{alphatreeoutputCONS} calculated in step 3 are those used to ensure \cref{thm:entropyupper} holds.
By setting \( \meas{M}_{\mbox{\tiny{\textup{t}}}} \leftarrow \meas{M}_{s_{\mathrm{w}}} \) and \( \postarget \leftarrow \bayespo \), 
\cref{alg:topdown} improves \( \cvar{} \) by iteratively improving the \( \alpha\)-tree's worst subgroup entropy \( \entropy(\alphatree; \meas{M}_{s_{\mathrm{w}}}, \bayespo) \), which as a surrogate improves the worst subgroup risk \( \totrisk(\posunfair; \meas{M}_{s_{\mathrm{w}}}, \bayespo) \). To accommodate for different \( \beta \) quantiles values for \( \cvar{} \), \topdown can be run repeatedly (replacing the initial input tree \( \alphatree_{0} \)) to progressively improve all \( s\in \mathcal{S}_{\beta}\). Hence, to reduce $\cvar{}$, we basically

\begin{tcolorbox}[colframe=blue,boxrule=0.5pt,arc=4pt,left=6pt,right=6pt,top=2pt,bottom=4pt,boxsep=0pt]
  \begin{equation} \label{topdown_cvar}
    \textrm{
    use
    \topdown~with $\meas{M}_{\mbox{\tiny{\textup{t}}}} \leftarrow \meas{M}_s$ $(s \in \mathcal{S}_\beta)$ and $\postarget \leftarrow \bayespo$.}
    \tag{\(\cvar{}\)}
%\topdown~with $\meas{M} \leftarrow \meas{M}_s$ $(s \in \mathcal{S}_\beta)$ and $\postarget \leftarrow \bayespo$,
    \end{equation}
\end{tcolorbox}%
As alluded to by the notation used to instantiate \topdown, the inputs of the algorithm can be instantiated to optimize for fairness criteria beyond \( \cvar{} \). This is discussed in \cref{sec-fairness}. In the usual ML setting, \( \meas{M}_{\mbox{\tiny{\textup{t}}}}, \postarget \) can be \textit{estimated} from a training sample (see \cref{sec-experiments}).
%For example, in \eqref{topdown_cvar} the edge can be calculated with samples of \( (\X, \Y) \sim \meas{D} \) with \( \SSS = s \).

\csubsec{Initialization}
In the procedure of improving \( \cvar{} \), the worst subgroups can be iteratively improved. However, we also need to make sure that improvement of a subgroup does not adversely affect another subgroup (which could potentially harm \( \cvar{} \) instead). As such, we introduce an additional structure to the \( \alpha \)-tree \( \alphatree \) by tweaking the initial tree structure \( \alphatree_{0} \) used in \topdown.
Since the fairness attribute \( \SSS \) partitions the dataset, a convenient choice of initializing the \( \alpha \)-tree is to split by the subgroup modalities, as depicted in \cref{fig:bigalpha}. As such, we grow separate sub-\(\alpha\)-trees for each of the sensitive modalities. For \( \cvar{} \), this allows the subgroup risk of individual subgroups to be tweaked without adversely affecting other subgroups.

\section{Formal Properties}
\label{sec-formal}

We move to the formal properties of our approach. We first detail the background of improper loss functions which motivates our correction given in \cref{ourposfair}. We then present the formal properties of this correction. The useful properties of having \( \alpha(.) \) represented by a tree structure is then discussed. Finally, we present a convergence analysis of \cref{alg:topdown} and an alternative boosting scheme.

%\todo{There are 2 version of this subsection, change the variable below to 0 or 1}

% Can we generalize approach beyond CVaR?
\paragraph{Can \( \posfair \) as per \eqref{ourposfair} correct (any) potential unfairness? Yes.} 
In short, this comes from recent theory in \emph{improper loss functions} for class probability estimation (CPE) \citep{nssBP}. We are interested in the pointwise minimizer (eventually set-valued) of:
\begin{eqnarray}
    \pseudob(\posterior) & \defeq & \arg\inf_{u\in[0,1]} L(u,\posterior).
\end{eqnarray}
Dubbed as the \emph{Bayes tilted estimate} of a loss \(\loss\) \citep{nssBP}, \( \pseudob(\posterior) \) is the set of optimal ``responses'' given a ground truth (pointwise) posterior \( \posterior \). Common loss functions are \emph{proper}: the ground truth value \( \posterior \in \pseudob(\posterior) \) is an optimal response. However, in the case where \( \posterior \) cannot be trusted (for instance when it is \emph{unfair}), we may not want to default to imitating \( \posterior \). In addition we also want to make sure that the Bayes tilted estimate can fit to any desired (in our context, \emph{fair}) target. The so-called $\alpha$-loss \( \loss^{\alpha} \), which generalizes the (proper) log-loss, is a good candidate parameterized by a variable \( \alpha \). Its Bayes tilted estimate is the pointwise version of \eqref{ourposfair}, for \( \alpha \not \in \{0, \infty\}\) and \( \posterior \neq 1/2 \):
\begin{eqnarray}
\apseudob(\posterior) & = & \left\{\posterior^\alpha/(\posterior^\alpha+(1-\posterior)^\alpha)\right\} .\label{eqCORRECT}
\end{eqnarray}
Importantly for \( \alpha \)-losses, for any $\posterior \not\in \{0,1/2,1\}$ and any $\posterior' \in (0,1)$, there exists $\alpha \in \mathbb{R}$ in \eqref{eqCORRECT} such that $\apseudob(\posterior) = \{\posterior'\}$. This property, called \emph{twist-properness} \citep{nssBP}, allows for any pointwise correction. By extending \( \alpha \) to a function (of \( \ve{x} \in \mathcal{X} \), as per \eqref{ourposfair}), twist-properness ensures that given an initial unfair posterior an appropriately learned \( \alpha(.) \) can correct any unfairness. This allows for \des{flexible} fairness post-processing -- different \( \alpha \) functions can be learned for different criteria (\emph{i.e.}, \cref{fig:teaser}).

%\todo{Not sure where the correct spot to talk about this is}

%\subsection{Choice of Log-Loss}
\paragraph{Why use the Log-Loss?}
As per \cref{sec-example}, we minimize the log-loss. We choose the log-loss for two reasons: \ding{172} it is strictly proper and so minimizing \( \totrisk(\posfair; \meas{M}_{\mbox{\tiny{\textup{t}}}}, \postarget) \) (\emph{i.e.}, via \topdown) ``pushes'' $\posfair$ towards target $\postarget$; and \ding{173} it is the $\alpha$-loss for $\alpha = 1$, so we are guaranteed that for the minimizer
$\posfair \rightarrow \posunfair \iff \alpha \rightarrow 1$.
%\emph{loss is perfectly minimized} $\iff \alpha \rightarrow 1$. 
With alternative (\emph{i.e.} non-strictly proper) loss, we might have only ``$\Rightarrow$''.
%$\posfair \rightarrow \posunfair \iff \alpha \rightarrow 1$. Otherwise taking another loss, we might have only $\Leftarrow$.
%As mentioned in \cref{sec-example}, we choose to minimize the log-loss. We choose the log-loss for two reasons: first it is strictly proper and so minimizing \( \totrisk(\posfair; \meas{M}, \postarget) \) (\emph{i.e.}, via \topdown) ``pushes'' $\posfair$ towards target $\postarget$; and second it is also the $\alpha$-loss for $\alpha = 1$, so we are guaranteed that $\posfair \rightarrow \posunfair$ \textit{is equivalent to} $\alpha \rightarrow 1$. Otherwise we might have only $\Leftarrow$.

%\paragraph{How close is \( \posfair \) to \( \posunfair \)?}
\paragraph{Do we have guarantees on some proximity of \( \posfair \) with respect to \( \posunfair \)? Yes, with light assumptions.}
%To examine the \des{proximity} guarantees, we move from considering pointwise posteriors back to functions over \( \mathcal{X} \).
We examine the \des{proximity} of black-box and post-processed posteriors with the KL divergence \citep{anMO}:
\begin{eqnarray}
 %\kl(\posunfair, \posfair; \meas{M}) & \defeq & \expect_{(\X, \Y)\sim \measdunfair} \left[\log \left(\frac{\dmeasdunfair((\X,\Y))}{\dmeasdfair((\X,\Y))}\right) \right],\label{defKL}
 \kl(\posunfair, \posfair; \meas{M}) & \defeq & \expect_{(\X, \Y)\sim \measdunfair} \left[\log \left({\dmeasdunfair((\X,\Y))} \big/ {\dmeasdfair((\X,\Y))}\right) \right],\label{defKL}
\end{eqnarray}
where $\measdunfair, \measdfair$ are the product measures defined from $\meas{M}$ and their respective posteriors. To bound the proximity \eqref{defKL}, we present setting \textbf{(S1)}.
\begin{enumerate}
\item [\textbf{(S1)}] Assumption \ref{assumptionBUNF} holds for some $0<B\leq 3$ and function $\alpha$ satisfies $|\alpha(\ve{x})-1| \leq 1/B$ (a.s.).
\end{enumerate}
This setting lead to the following \emph{data independent} proximity bound.
\begin{theorem}\label{cor-distortion-to-blackbox}
  For any \( \meas{M} \),
  \textup{\textbf{(S1)}} implies
  \( 
  \kl(\posunfair, \posfair; \meas{M}) \leq \pi^2/(6\cdot (2 + \exp (B) + \exp (-B))) \).
\end{theorem}
As an example, fix $B=3$ for \textbf{(S1)}. In this case, we want $\alpha(.) \in [2/3, 4/3]$ (a.s.) which is a reasonable sized interval centered at 1. The clamped black-box posterior's interval is approximately $[0.04, 0.96]$, which is quite flexible and the distortion is upperbounded as $\kl(\posunfair, \posfair; \meas{M}) \leq 7.5E-2$.

%\paragraph{How do composed corrections interact?}
\paragraph{Is the composition of transformations meaningful? Yes.}
The analytical form in \eqref{ourposfair} brings the following easy-to-check \textbf{(composability)} property.
\begin{lemma}\label{lemcomposability}
The composition of any two wrapping transformations $\posunfair \raisebox{-3pt}{ $\stackrel{\alpha}{\mapsto}$ } \posfair \raisebox{-3pt}{ $\stackrel{\alpha'}{\mapsto}$ } \posfair'$ following \eqref{ourposfair} is equivalent to the single transformation $\posunfair \raisebox{-3pt}{ $\stackrel{\alpha \cdot \alpha'}{\mapsto}$ } \posfair'$.
%The composition of any two wrapping transformations $\posunfair \raisebox{-3pt}{ $\stackrel{\alpha}{\mapsto}$ } \posfair \raisebox{-3pt}{ $\stackrel{\alpha'}{\mapsto}$ } \posfair'$ following \eqref{ourposfair} is equivalent to the single transformation $\posunfair \raisebox{-3pt}{ $\stackrel{\alpha \cdot \alpha'}{\mapsto}$ } \posfair'$.
\end{lemma}
This gives an \emph{invertibility} condition -- wrapping $\posfair$ with $\alpha' = 1/\alpha$ recovers the original black-box \( \posunfair \).

%\paragraph{\todo{Question about explainability? Maybe add this to the fairness criteria section (as a catch-all for "socially minded problems")}}
%
%\todo{TODO}

\paragraph{Given some capacity parameter for \( \posunfair \), can we easily compute that of \( \posfair \)? Yes, \emph{e.g.}, for decision trees.} 
Such a question is particularly relevant for generalization. As we are using the log-loss \eqref{lossgen}, a relevant capacity notion to assess the uniform convergence of risk minimization for the whole wrapped model is the Rademacher \textbf{(complexity)} \citep{bmRA}.
We examine the following set of functions:
\begin{eqnarray}
\mathcal{H}_{\mbox{\tiny{f}}} & \defeq & \left\{ \posfair : \posfair(\ve{x}) \textrm{ given by \eqref{ourposfair} with } \alpha, \posunfair; \forall (\alpha, \posunfair)\right\},\label{defHR}
\end{eqnarray}
%\begin{eqnarray}
%\mathcal{H}_{\mbox{\tiny{f}}} & \defeq & \left\{\alpha(\ve{x}) \cdot \log\left(\frac{\posunfair(\ve{x})}{1-\posunfair(\ve{x})}\right) : \forall (\alpha, \posunfair)\right\},\label{defHR}
%\end{eqnarray}
%
where we assume known the set of functions from which $\posunfair$ was trained.
We now assume we have a $m$-training sample $\mathcal{S} \defeq \{(\ve{x}_i, y_i)\sim \meas{D}\}_{i=1}^{m} $. The empirical Rademacher complexity of a set of functions $\mathcal{H}$ from $\mathcal{X}$ to $\mathbb{R}$, $\radcomp_{\mathcal{S}}(\mathcal{H}) \defeq \expect_{\ve{\sigma}} \sup_{h \in \mathcal{H}} \expect_{i} [\sigma_i h(\ve{x}_i)]$
(sampling uniform with $\sigma_i \in \{-1,1\}$), is a capacity parameter that yields efficient control of uniform convergence when the loss used is Lipschitz \citep[Theorem 7]{bmRA}, which is the case of the log-loss.
%(which we shall use).
%\todo{
%{%
To see how the $\alpha$-tree affects the Rademacher complexity of classification using $\posfair$ instead of $\posunfair$, suppose real-valued prediction based on $\posunfair$ is achieved via logit mapping, $\logit \circ \posunfair$ \eqref{defHR}.
%}%
Such mappings are common for decision trees \citep{ssIB}.
\begin{lemma} \label{lem:complexity}
  Suppose $\{\posunfair\}$ is the set of decision trees of depth $\leq d$ and denote $\radcomp_{\mathcal{S}}(\textsc{dt}(d))$ the empirical Rademacher complexity of decision trees of depth $\leq d$ \citep{bmRA} and $d'$ the maximum depth allowed for $\alpha$-trees. Then we have for $\mathcal{H}_{\mbox{\tiny{\textup{f}}}}$ in \eqref{defHR}: $\radcomp_{\mathcal{S}}(\mathcal{H}_{\mbox{\tiny{\textup{f}}}}) \leq \radcomp_{\mathcal{S}}(\textsc{dt}(d+d'))$.
\end{lemma}%
The proof is straightforward once we remark that elements in $\mathcal{H}_{\mbox{\tiny{f}}}$ can be represented as decision trees, where we plug at each leaf of $\posunfair$ a copy of the $\alpha$-tree $\alphatree$.

%\paragraph{What are the convergence properties of \cref{alg:topdown}?}
\paragraph{Does \cref{alg:topdown} have any convergence properties? Yes, it is a boosting algorithm.}
Following a similar blueprint to classical decision tree induction, it comes with no surprise that \topdown~can achieve boosting compliant convergence. To show that \topdown~is a boosting algorithm, we need a \emph{Weak Hypothesis Assumption (WHA)}, which postulates informally that each chosen split brings a small edge over random splits for a tailored distribution.
\begin{definition}\label{balancedDist}
  Let $\leaf \in \leafset(\alphatree)$ and ${\measdtarget}_\leaf$ be the product measure on $\mathcal{X} \times\mathcal{Y}$ conditioned on $\leaf$. The \textbf{balanced} product measure ${\rebmeasdtarget}_\leaf$ at leaf $\leaf$ is defined as ($z \defeq (\ve{x},y)$ for short): 
  \begin{eqnarray}
{\rebmeasdtarget}_\leaf (z) & \defeq & \frac{1 - \edge(\meas{M}_\leaf, \postarget) \cdot y \normalizedlogit(\posunfair(\ve{x}))}{1 - \edge(\meas{M}_\leaf, \postarget)^2} \cdot {\measdtarget}_\leaf (z).\label{defDMEASTARGET2}
  \end{eqnarray}
\end{definition}
We check that $\int_{\leaf} \mathrm{d} {\rebmeasdtarget}_\leaf = 1$ because of Def.~\ref{def:edge}. Our balanced distribution is named after \citet{kmOT}'s: ours indeed generalizes theirs. 
The key difference comes from the change in setting, where we consider the alignment purity of a leaf and not its label purity. The ``\textit{fairness-free case}'' where $\normalizedlogit(.)$ is replaced by constant $1$ yields the original balanced distribution \citep{kmOT}. We now state our WHA.
\begin{assumption}\label{defWHA}
  %Let $h : \mathcal{X} \rightarrow \{-1,1\}$ be the function splitting leaf $\leaf$, and let $\upgamma > 0$. We say that $h$ \textbf{$\upgamma$-witnesses} the Weak Hypothesis Assumption (\textbf{WHA}) at $\leaf$ iff
    \begin{flushleft}
    Let $h : \mathcal{X} \rightarrow \mathcal{Y}$ be the function splitting leaf $\leaf$. For $\upgamma > 0$, then \textbf{$h$ $\upgamma$-witnesses} the Weak Hypothesis Assumption (\textbf{WHA}) at $\leaf$ iff
    \begin{minipage}{.4\linewidth}%
        \begin{equation*}
        \mbox{\bf (i)} \; \bigg\vert\mathop{\expect}_{(\X, \Y) \sim {\rebmeasdtarget}_\leaf} \left[\Y \normalizedlogit(\posunfair(\X)) \cdot h(\X)\right]\bigg\vert \geq \upgamma;
        \end{equation*}
    \end{minipage}%
    \begin{minipage}{.6\linewidth}
        \begin{equation*}
        \quad \mbox{\bf (ii)} \; \edge(\meas{M}_\leaf, \postarget) \cdot \mathop{\expect}_{(\X, \Y) \sim {\measdtarget}_{\leaf}} \left[(1-\normalizedlogit^2(\posunfair(\X)))\cdot h(\X)\right] \leq 0.
        \end{equation*}
    \end{minipage}
    \end{flushleft}%
\end{assumption}
Intuitively, (i) gives a condition on the split \( h \)'s correlation with unfair posterior \( \posunfair \) agreement with labels and (ii) does the same for the confidence of \( \posunfair \) predictions. Similarly to the balanced distribution, in the fairness-free case where we only care about the label purity of splits, the WHA simplified to that of \citet{kmOT}'s -- \emph{i.e.}, only (i) matters.
A further discussion on our balanced distribution and WHA is in the \supplement, Section \ref{sec-wha-disc}. We now state \topdown's boosting compliant \des{convergence}.
\begin{theorem}\label{thm-boosting}
 Suppose (a) Assumption \ref{assumptionBUNF} holds, (b) we pick the heaviest leaf to split at each iteration in Step 2.1 of \topdown and (c) $\exists \upgamma > 0$ such that each split $h^\star$ (Step 2.2) in $\alphatree$ $\upgamma$-witnesses the WHA. Then there exists a constant $c>0$ such that $\forall \varepsilon > 0$, if the number of leaves of $\alphatree$ satisfies $|\leafset(\alphatree)| \geq \left(1/\varepsilon\right)^{c \log \left(\frac{1}{\varepsilon}\right)/\upgamma^2}$,
then $\posfair$ crafted from \eqref{ourposfair} using \topdown's $\alphatree$ achieves $\totrisk(\posfair; \meas{M}_{\mbox{\tiny{\textup{t}}}}, \postarget)  \leq \varepsilon$.
\end{theorem}
%
%The proof of Theorem \ref{thm-boosting} follows from Theorem \ref{thm:entropyupper} and adapting the convergence proof for binary entropy and decision trees \citep{kmOT}. The full proof is in \supplement, Section \ref{proof-thm-boosting}. 

\begin{figure}[t]
  \centering
  %\begin{tabular}{cc}
    \includegraphics[trim=0bp 400bp 560bp 20bp,clip,height=0.28\textwidth]{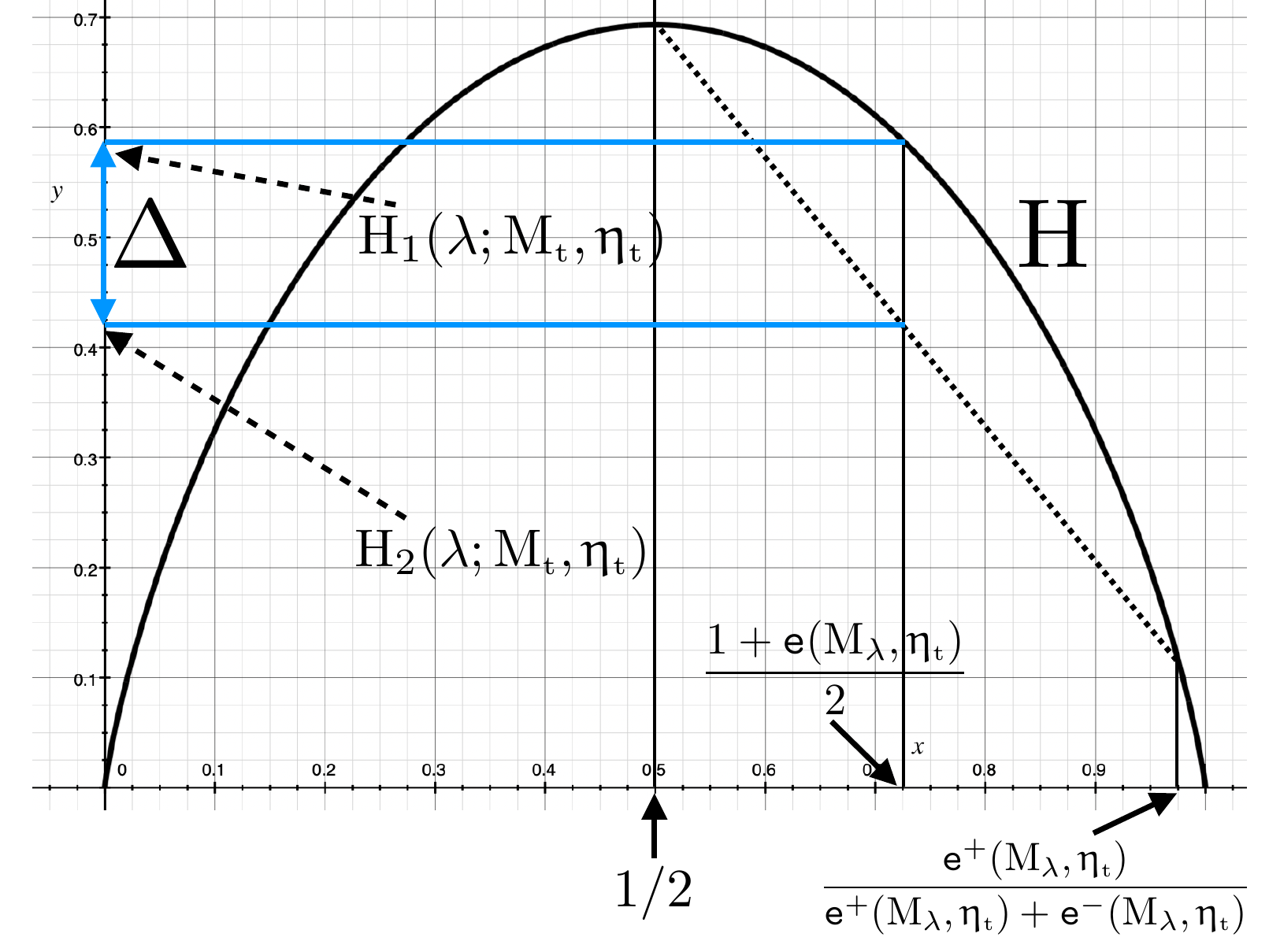}%
  %  &
    \hspace{2cm}
    \includegraphics[trim=5bp 455bp 690bp 15bp,clip,height=0.27\textwidth]{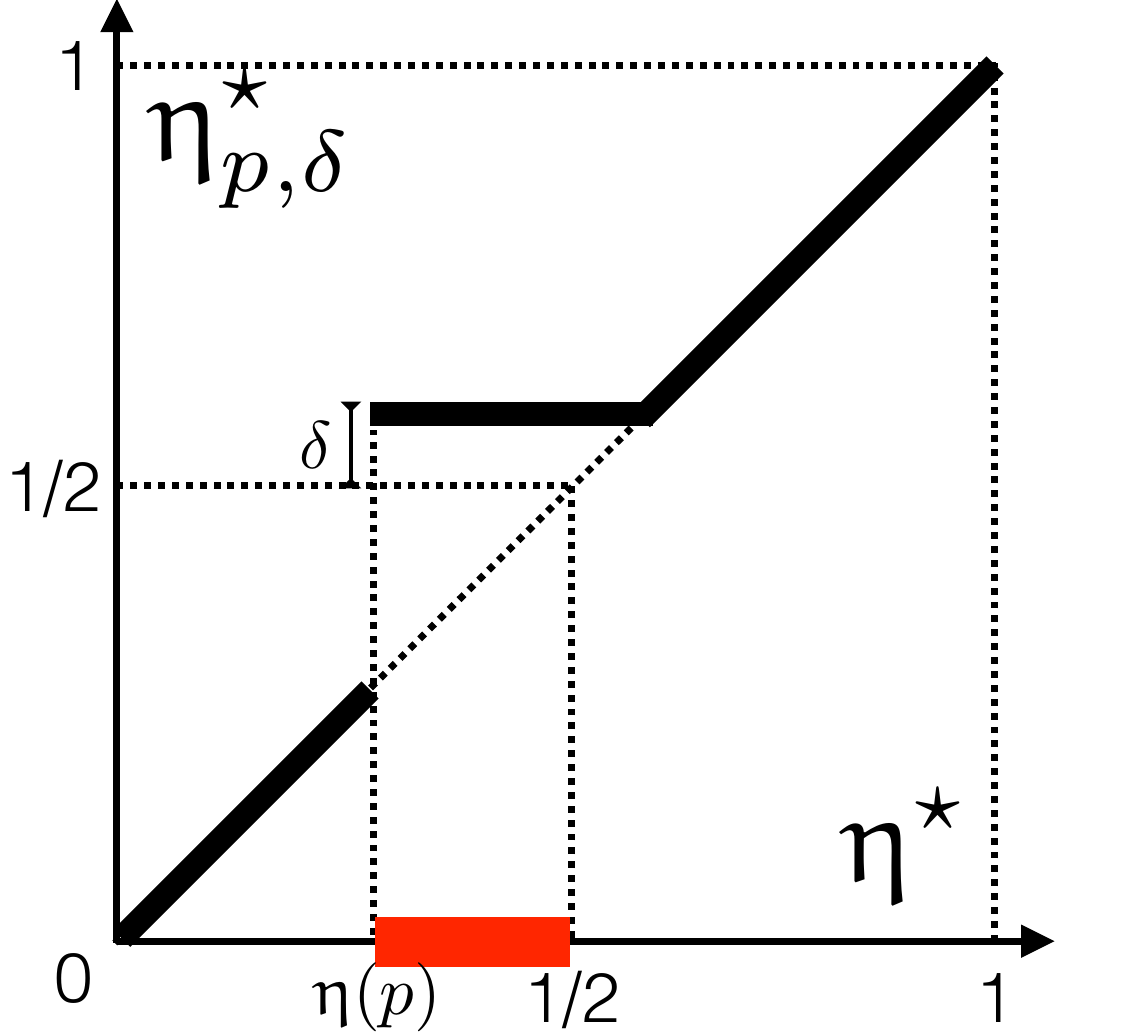}%
  %  \end{tabular}
     \caption{\textit{Left}: Difference between the per-leaf bounds on risk $\totrisk(\posunfair; \meas{M}_{\mbox{\tiny{\textup{t}}}}, \postarget)$ using \eqref{defBINENT} and \cref{thm:entropyupper} (conservative scoring) and \eqref{totriskB2} (audacious scoring). Details in the proof of Lemma \ref{lem-entropies}. \textit{Right}: A representation of the $(p,\delta)$-pushup of $\bayespo$, where $\eta(p) \defeq \inf \bayespo(\mathcal{X}_p) < 1/2$ (Def. \ref{def-pushup}). All posteriors in $[\eta(p), 1/2 + \delta]$ are mapped to $1/2 + \delta$; others do not change. New posterior $\bayespo_{p,\delta}$ eventually reduces the accuracy of classification for observations whose posterior lands in the thick red interval ($x$-axis).}
    \label{fig:eta-diff-eoo}
    \fnegativespace
   \end{figure}%

\paragraph{Are there alternative ways of growing \( \alpha\)-trees? Yes.}
Let us call \textit{conservative} the scoring scheme in \eqref{alphatreeoutputCONS}. There is an alternative scoring scheme, which can lead to substantially larger corrections in absolute values, hence the naming, and yields better entropic bounds for the $\alpha$-tree.
\begin{definition}\label{audaciousScore}
  For any mixture $\meas{M}_{\mbox{\tiny{\textup{t}}}}$ and posteriors $\posunfair, \postarget$, let \( \edge^+(\meas{M}_{\mbox{\tiny{\textup{t}}}}, \postarget) \) and \( \edge^-(\meas{M}_{\mbox{\tiny{\textup{t}}}}, \postarget) \) be defined by
  \begin{eqnarray}
    \edge^{\pm}(\meas{M}_{\mbox{\tiny{\textup{t}}}}, \postarget) & \defeq & {\expect}_{(\X, \Y) \sim \measdtarget} \left[\max\{0, \pm \Y  \normalizedlogit(\posunfair(\X))\}\right]. \label{defEDGEGEPNMF}
  \end{eqnarray}
%\begin{minipage}{.47\linewidth}
%  \begin{equation}
%    \hspace{-0.1cm}\edge^+(\meas{M}, \postarget) \defeq \hspace{-0.3cm}\mathop{\expect}_{(\X, \Y) \sim \measdtarget} \hspace{-0.2cm}\left[\max\{0,\Y  \normalizedlogit(\posunfair(\X))\}\right];\hspace{-0.9cm}\label{defEDGEGEPMF}
%  \end{equation}
%\end{minipage}%
%\begin{minipage}{.53\linewidth}
%  \begin{equation}
%    \edge^-(\meas{M}, \postarget) \defeq -\hspace{-0.3cm}\mathop{\expect}_{(\X, \Y) \sim \measdtarget} \hspace{-0.2cm}\left[\min\{0,\Y  \normalizedlogit(\posunfair(\X))\}\right].\label{defEDGEGENMF}
%  \end{equation}
%\end{minipage}
%
The \textbf{audacious} scoring schemes at the leaves of the $\alpha$-tree replaces \eqref{alphatreeoutputCONS} in Step 3 by:
%\negativespace
  \begin{eqnarray*}
\alphatree(\leaf) & \defeq & \normalizedlogit\left(\frac{\edge^+(\meas{M}_{\leaf}, \postarget)}{\edge^+(\meas{M}_{\leaf}, \postarget) + \edge^-(\meas{M}_{\leaf}, \postarget)}\right), \quad \forall \leaf \in \leafset(\alphatree).
%\negativespace
  \end{eqnarray*}
\end{definition}
\begin{theorem}\label{thm-alternative-alpha-in-alpha-tree}
 Suppose Assumption \ref{assumptionBUNF} holds and let $\entropy_2(q) \defeq \entropy(q) / \log 2$ ($\in [0,1]$), $H$ being defined in Definition \ref{defENTROPYA}.  For any leaf $\leaf \in \leafset(\alphatree)$, denote for short:
 \begin{eqnarray*}
    \entropy_2(\leaf; \meas{M}_{\mbox{\tiny{\textup{t}}}}, \postarget) & \defeq & \log(2) \cdot \left( 1 + (\edge^+_{\leaf} + \edge^-_{\leaf}) \cdot \left( \entropy_2  \left( \frac{\edge^+_{\leaf}}{\edge^+_{\leaf}+\edge^-_{\leaf}} \right)  -  1 \right)\right),
  \end{eqnarray*}%
  where let $\edge^b_{\leaf} \defeq \edge^b(\meas{M}_{\leaf}, \postarget), \forall b \in \{+,-\}$. Using audacious scoring, we get instead of
  \cref{thm:entropyupper}:
  %\eqref{totriskB1}:
  \begin{eqnarray}
  \totrisk(\posfair; \meas{M}_{\mbox{\tiny{\textup{t}}}}, \postarget) & \leq  & \expect_{\leaf \sim \meas{M}_{\leafset(\alphatree)}} \left[\entropy_2(\leaf; \meas{M}_{\mbox{\tiny{\textup{t}}}}, \postarget)\right]. \label{totriskB2}
%\negativespace
\end{eqnarray}
\end{theorem}
While upperbounds in \cref{thm:entropyupper}
%\eqref{totriskB1}
and \cref{thm-alternative-alpha-in-alpha-tree} may look incomparable, it takes a simple argument to show that \eqref{totriskB2} is never worse and can be much tighter.
\begin{lemma}\label{lem-entropies}
$\forall \alpha$-tree $\alphatree$, $\expect_{\leaf \sim \meas{M}_{\leafset(\alphatree)}} \left[\entropy_2(\leaf; \meas{M}_{\mbox{\tiny{\textup{t}}}}, \postarget)\right] \leq \entropy(\alphatree; \meas{M}_{\mbox{\tiny{\textup{t}}}}, \postarget)$.
\end{lemma}
%\negativespace
It thus comes at no surprise that using the audacious scoring also results in a boosting result for \topdown~guaranteeing the same rates as in Theorem \ref{thm-boosting}. It also takes a simple picture to show that the per-leaf slack in Lemma \ref{lem-entropies} can be substantial, a slack which can be represented using a simple picture, see Figure \ref{fig:eta-diff-eoo} (left), following from the use of Jensen's inequality in the Lemma's proof.

\paragraph{As audacious scoring is better boosting-wise, is conservative scoring useful? Yes.}
%\paragraph{Since audacious scoring is better, boosting-wise, than conservative scoring, is conservative scoring useful? Yes.}
If we only cared about accuracy, we would barely have any reason to use the conservative correction. Even thinking about generalization, the Rademacher complexity of decision trees is a function of their depth so faster the convergence, the better \citep[Section 4.1]{bmRA}. Adding fairness substantially changes the picture: some constraints, like equality of opportunity (Section \ref{sec-fairness}) can antagonize accuracy to some extent. In such a case, using the conservative correction can keep posteriors $\posunfair$ and $\postarget$ close enough (\cref{cor-distortion-to-blackbox}%
%/ \cref{thm-distortion-to-blackbox}
) so that fairness can be achieved without substantial sacrifice on accuracy.
\section{Fairness and Societal Considerations}
\label{sec-fairness}
In this section, we present the fairness guarantees \topdown~can achieve. In particular, we provide a discussion about how \cref{thm-boosting} can guarantee minimization of the \( \cvar{} \) criteria. Furthermore, we provide alternative inputs to \topdown~which allows for EOO to be targeted as a fairness criteria. In the \supplement Section \ref{sec-stat-par}, we further present a treatment of statistical parity. Lastly, we discuss how using \( \alpha \)-trees provides explainable corrections and how utilization of the sensitive attribute (as per \cref{fig:bigalpha}) can be circumvented.

\csubsec{Guarantees on CVaR}
As discussed in previous sections, one way to improve the \( \cvar{} \) fairness criteria (as per \eqref{eq:cvar}) is to focus optimization on the worst treated subgroups. Given a specified quantile group \( \beta \) and the set of worse subgroups \( \mathcal{S}_{\beta} \), we can repeat \eqref{topdown_cvar} until \( \cvar{\beta}(\posfair) \) gets below a threshold or (more specifically) its worst tread group gets a risk below a threshold (\emph{i.e.}, a stopping criterion). Importantly, \cref{thm-boosting} provides a guarantee: to ensure \( \cvar{\beta} \) is below \( \varepsilon \), we simply need to boost for \( \vert S \vert \) times the tree size bound \( \vert \leafset(\alphatree) \vert \) given in \cref{thm-boosting}.

\csubsec{Guarantees on EOO}
EOO requires to smooth discrimination within an ``advantaged'' group, modeled by the label $\Y=1$ \citep{hpsEO}. We say that $\posfair$ achieves $\varepsilon$-equality of opportunity iff a mapping $\predfair$ of $\posfair$ to $\mathcal{Y}$ (\textit{e.g.} using the sign of its logit \( \logit \)) satisfies
\begin{eqnarray}
    \max_{s\in S} \mathop{\pr}_{\X \sim \meas{P}_s} \left[\predfair(\X) = 1 \right] - \min_{s\in S} \mathop{\pr}_{\X \sim \meas{P}_{s}} \left[\predfair(\X) = 1 \right] &\leq& \varepsilon\label{eqEOO},
\end{eqnarray}
where $\meas{P}_s$ is the positive observations' measure conditioned to value $\SSS = s$ for the sensitive attribute.
It is clear that EOO can be antagonistic to accuracy: the rate of advantage in the data \( \meas{D} \) may not be equal among the subgroups. As such, unlike \( \cvar{} \), we do not want to target the Bayes posterior \( \postarget \not\leftarrow \bayespo \) for EOO. Instead, we target a skewed posterior which aims to improve the least advantaged subgroup, \emph{i.e.,} increasing
%
%EOO can be antagonistic with the fitting of $\posfair$ to $\bayespo$: if that latter one is close to zero in a subgroup and close to one in another one, then better fittings on $\posfair$ can arbitrarily increase the LHS in \eqref{eqEOO}. To cope with this issue, we do not pick $\postarget \leftarrow \bayespo$ as in $\cvar{}$, but rather skew the posterior for a subset of observations.
$s^\circ \in \arg\min_{s\in S} \pr_{\X \sim \meas{P}_s} \left[\predfair(\X) = 1 \right]$.
Our strategy consists of picking a target posterior which skews part of the original \( \bayespo \) to be more advantaged, thus reducing the LHS of \eqref{eqEOO} until \eqref{eqEOO} is satisfied%
%in skewing the target posterior for $S = s^\circ$ so that for a subset of the subgroup, it becomes $>1/2$. 
%A convenient use of \topdown~then yields more positive classifications for $S = s^\circ$ -- thus a more fair outcome -- and thus a reduction of LHS in \eqref{eqEOO} until \eqref{eqEOO} is satisfied
%\footnote{A symmetric strategy holds if one instead wants to \textit{reduce} $\pr_{\X \sim \meas{P}_{s^*}} \left[\predfair(\X) = 1 \right]$ ($s^* \in \arg\max_{s\in S} \pr_{\X \sim \meas{P}_s} \left[\predfair(\X) = 1 \right]$).
%Choosing one strategy depends on the application: if positive class implies money spending (\textit{e,g,} for loan prediction), then our strategy implies spending more money to achieve fairness, while the latter one reduces the amount of money lent to achieve fairness.}.
\footnote{If we instead \textit{reduce} $\arg\max_{s\in S} \pr_{\X \sim \meas{P}_s} \left[\predfair(\X) = 1 \right]$ we get a symmetric strategy.
%Choosing one strategy depends on the application
The application informs which to use: if positive class implies money spending (\textit{e.g.} loan prediction), then our strategy implies spending more money; while the latter aims to reduce money lent to achieve fairness.}.
For this, we create a $(p,\delta)$-\textit{pushup} of $\bayespo$, defined in \supplement \cref{sec:sup_eoo}.
%\begin{definition}\label{def-pushup}
%    Fix $p\in [0,1]$ and let $\mathcal{X}_p$ be a subset of $\mathcal{X}$ such that (i) $\inf \bayespo(\mathcal{X}_p) \geq \sup \bayespo(\mathcal{X}\backslash \mathcal{X}_p)$ and (ii) $\int_{\mathcal{X}_p} \dmeas{M} = p$. For any $\delta \geq 0$, the $(p,\delta)$-pushup of $\bayespo$, $\bayespo_{p,\delta}$, is the posterior defined as $\bayespo_{p,\delta} = \bayespo$ if $\inf \bayespo(\mathcal{X}_p) \geq 1/2$ and otherwise:
%    \begin{eqnarray}
%      \bayespo_{p,\delta}(\ve{x}) & \defeq & \left\{
%                                             \begin{array}{ccl}
%                                               \bayespo(\ve{x}) & \mbox{ if } & (\ve{x} \not\in \mathcal{X}_p) \vee \left(\bayespo(\ve{x}) \geq \frac{1}{2}+\delta\right)\\
%                                               \frac{1}{2}+\delta & \multicolumn{2}{l}{\mbox{ otherwise.}}
%                                               \end{array}
%                                             \right. \label{eoo_mapping}
%      \end{eqnarray}
%\end{definition}
%

Fig.~\ref{fig:eta-diff-eoo} (right) presents an example of a pushup map.
Notice that the pushup only changes the predicted probability of example which do not have a ``confident prediction'' (the interval $[\eta(p), 1/2 + \delta]$). 
%Furthermore, on these examples the probability is ``pushed up'' to increase advantage. 
Intuitively, \( p \) controls how many examples are corrected and \( \delta \) controls how much the correction ``pushes up'' advantage, further discussion in \supplement \cref{sec:sup_eoo}.
%
%
%Notice that the transformation can introduce classification mistakes w.r.t. $\bayespo$, but only for examples with (a) small ``edge'' $|1/2 - \bayespo|$ and (b) labeled as negative on $\bayespo$ are susceptible to get positive label on $\bayespo_{p,\delta}$. Notice the tradeoff: (b) is consistent with the fairness objective while (a) limits the degradation in accuracy.
%
We then run \topdown~using as mixture the \textit{positive} measure conditioned to $S = s^\circ$ and $p \defeq \pr_{\X \sim \meas{P}_{s^*}} \left[\predfair(\X) = 1 \right] + \epsilon/(K-1), \delta \defeq K\epsilon/(K-1)$, with $K>1$ user-fixed. Thus, we do
\begin{tcolorbox}[colframe=blue,boxrule=0.5pt,arc=4pt,left=6pt,right=6pt,top=2pt,bottom=4pt,boxsep=0pt]
\begin{equation}
  \textrm{
    Use \topdown~with $\meas{M}_{\mbox{\tiny{\textup{t}}}} \leftarrow \meas{P}_{s^\circ}$ and $\postarget \leftarrow \bayespo_{p,\delta}$
    }.
    \tag{EOO}
\end{equation}
\end{tcolorbox}
%  and we have the following guarantee:
  \begin{theorem}\label{thm-eoo}
    If \topdown~is run until $\totrisk(\posfair; \meas{M}_{\mbox{\tiny{\textup{t}}}}, \postarget) \leq  (\epsilon^4/2) + \expect_{\X \sim \meas{M}_{\mbox{\tiny{\textup{t}}}}}  \left[ \entropy(\postarget(\X))\right]$, then after the run we observe $\pr_{\X \sim \meas{P}_{s^\star}} \left[\predfair(\X) = 1 \right]  - \pr_{\X \sim \meas{P}_{s^\circ}} \left[\predfair(\X) = 1 \right] \leq \epsilon$.
\end{theorem}%
%For the optimization to be carried out properly in the full context of EOO, we should not wait to get the bound on $\totrisk(\posfair; \meas{M}, \postarget)$.
In the full context of EOO, in the optimization we should not wait to get the bound on $\totrisk(\posfair; \meas{M}_{\mbox{\tiny{\textup{t}}}}, \postarget)$.
Rather, we should make sure (a) we update $\arg\min_{s\in S} \pr_{\X \sim \meas{P}_s} \left[\predfair(\X) = 1 \right]$ (and thus $s^\circ$) after each split in the $\alpha$-tree and (b) we keep $\arg\max_{s\in S} \pr_{\X \sim \meas{P}_s} \left[\predfair(\X) = 1 \right]$ as is, to prevent switching targets and eventually composing pushup transformations for the same $S=s^\circ$, which would not necessarily comply with our theory.
Notably, the guarantee presented in \cref{thm-eoo} depends on the mapping \( \predfair \) and not the direct posterior \( \posfair \), as typically considered~\citep{hpsEO}.
%When taking the mapping as a threshold of the posterior (sign of the logit), \( \predfair \) can be interpreted as forcing the original posterior to be extreme values of 0 or 1. 
When taking a threshold (sign of the logit), \( \predfair \) can be interpreted as forcing the original posterior to be extreme values of 0 or 1. 

Unlike the \( \cvar{} \) case, EOO (as per \eqref{eoo_mapping}, \supplement) requires an explicit approximation of \( \bayespo \). In practice, we find that taking a simple approximation of \( \bayespo \) still can yield fairness gains. However, if one does not want to make such an approximation, one can adapt the statistical parity approach (detailed in \supplement, Section \ref{sec-stat-par}).
Similarly, if wants to consider the typical EOO definitions depending on posterior values, the target measure can be replaced (\emph{i.e.}, swapping measure \( \meas{M}_{\mbox{\tiny{\textup{t}}}} \) with the positive examples \( \meas{P} \)).

\csubsec{Explainability}
\( \alpha \)-trees using the initialization proposed in \cref{sec:growing_alpha_trees} (and \cref{fig:bigalpha}) allows for \des{explainability} properties similar to that of decision tree classifiers.
Fixing a sensitive attribute \( \SSS = s \), the corresponding sub-\(\alpha\)-tree \( \alphatree_{s} \) can be examined to scrutinize the correction done for the corresponding subgroup. If the splits of the \( \alpha \)-tree are simple, similarly to standard decision tree classifiers, corresponding partitions of the input domain can be examined. Furthermore, the type of corrections can also be examined, as discussed in \cref{sec-example}; where corrections can be classified as ``sharpening'', ``dampening'', or ``polarity flipping'' depending on the leaves' \( \alpha \)-values.

\csubsec{Usage of sensitive attribute}
Post-processing methods have been flagged in the context of fair classification for the fact that they require explicit access to the sensitive feature at classification time \citep[$\S$ 6.2.3]{zvggFC}. Our basic approach to the induction of $\alpha$-trees falls in this category (Fig. \ref{fig:bigalpha}), but there is a simple way to \textit{mask} the use of the sensitive attribute and the polarity of disparate treatment it induces: it consists in first inducing a decision tree to \textit{predict} the sensitive feature based on the other features and use this decision tree as an alternative initialization to naively splitting on subgroups. We thus also \textit{redefine} sensitive groups based on this decision tree -- thus alleviating the need to use the sensitive attribute in the \( \alpha \)-tree.
The use of \emph{proxy sensitive attributes} in a similar manner has seen ample use in a various domain such as health care~\citep{bUP,bkbkUB} and finance~\citep{fbwbsehrUO}.  We however note that its application in post-process and \( \alpha \)-trees may not be appropriate across all domains~\citep{dfkmsUP}.
%\section{Experiments}
%\label{sec-experiments}
%
%\begin{figure}
%    \centering
%    \includegraphics[width=\textwidth]{Figs/temp_new_plot.png}
%    \caption{\todo{Temp plot (to be exported to better quality). Currently just a draft on presentation. This might be a better alternative than the previous style of plot.}}
%    \label{fig:my_label}
%\end{figure}
%
\section{Experiments}
\label{sec-experiments}
To evaluate \topdown\footnote{Implementation public at: \url{https://github.com/alexandersoen/alpha-tree-fair-wrappers}}, we consider three datasets presenting a range of different size / feature types, Bank and German Credit (preprocessed by \aif~\citep{aif360}) and the American Community Survey (ACS) dataset preprocessed by \folktables\footnote{Public at: \url{https://github.com/zykls/folktables}} \citep{dhmsRA}. The \supplement~(pg \pageref{app-exp-settings}) presents all results at length (including considerations on proxy sensitive attributes, distribution shift, and interpretability), along with the different black-boxes considered (random forests and neural nets). We concentrate in the Section on the ACS dataset for income prediction in the state of CA and evaluate \topdown's application to various fairness criteria (as per \cref{sec-fairness} and \supplement pg \pageref{sec-stat-par}) with Random Forests (RF).
For these experiments, we consider \( \emph{age} \) as a binary sensitive attribute with a bin split at $25$ (a trinary modality is deferred to the \supplement).
For the black-box, we consider a clipped (\cref{assumptionBUNF} with \(B=1\)) random forest (RF) from \sklearn  calibrated using Platt's method \cite{platt1999probabilistic}. The RF consists of an ensemble of 50 decision trees with a maximum depth of 4 and a random selection of 10\% of the training samples per decision tree.
Data is split into 3 subsets for black-box training, post-processing training, and testing; consisting of 40:40:20 splits in 5 fold cross validation.
For \EOO, we utilize an out-of-the-box Gaussian Naive Bayes classifier from \sklearn to approximate \( \bayespo \).
\begin{figure*}[t]
    \centering
    \includegraphics[width=\textwidth,trim={29pt, 775pt, 768pt, 31pt},clip]{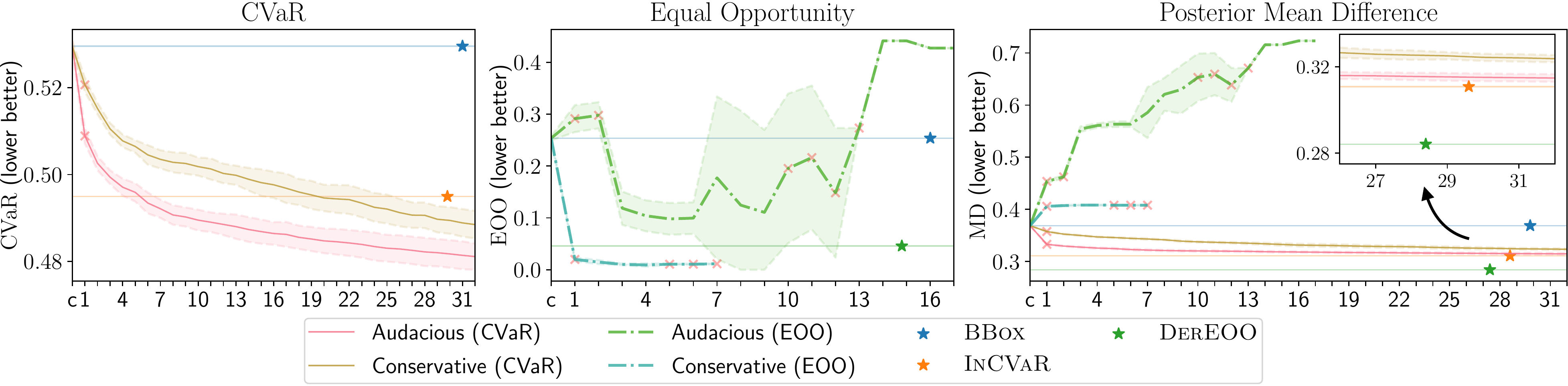}%
    %\vspace{-0.5cm}
    \caption{\textbf{ACS 2015 with Binary Sensitive Attribute and Random Forest Black-box}: Evaluation of \topdown~over boosting iterations (x-axis) for different fairness criteria. `c' on the x-axis denotes the clipped black-box. `\( \times \)' denote when a subgroup's \( \alpha \)-tree is initiated (over any fold). The shade denotes \( \pm \) a standard deviation from the mean, this disappears when folds have early stopping.}%
    \label{fig:main_text_fairness}%
    %\vspace{-0.5cm}
    \fnegativespace
\end{figure*}%

\noindent\textbf{Multiple fairness criteria}
We evaluate \topdown for \( \cvar{} \), equality of opportunity \( \EOO \), and statistical parity \( \SP \).
The complete treatment of SP is pushed to \supplement~(Sections \ref{sec-stat-par}, \ref{app-exp-SP-main}).
%Here we only make references to the results for readability.
%We have pushed to \supplement~(Sections \ref{sec-stat-par}, \ref{app-exp-SP-main}) the complete treatment of SP, that we just summarise here.
\( \SP \) aims to make subgroup's expected posteriors similar and is popular in a various post-processing methods~\citep{wrcOS,alAN}. The definition can be found in \supplement (pg \pageref{sec-stat-par}) along with the strategy used in \topdown.
%For \( \SP \), we present one strategy where we target the largest subgroup posterior (denoted by \( \uparrow \)).
%A symmetric strategy of targeting the smallest subgroup posterior is presented in \supplement. 
Conservative and audacious updates rules are also tested. For each of these \topdown~configurations, we boost for \( 32 \) iterations. The initial \(\alpha\)-tree is initialized as in \cref{fig:bigalpha}.

%To evaluate \topdown, 
We compare against 5 baseline approaches. For \( \cvar{} \) we consider the in-processing approach (\basecvara) presented in \citet{wmFR}. For \EOO, we consider a derived predictor (\baseeooa) \citep{hpsEO}. Our SP baselines include an optimized score transformation approach (\baseeoob)
%with is used for equalized odds (parity for both \(y = \pm 1\) instead of just \( y = 1\) in \EOO)
~\citep{wrcOS}; a derived predictor modified for \SP (\basespa) \citep{hpsEO}; and a randomized threshold optimizer approach (\basespb) \citep{alAN}.
We denote the clipped black-box as \basebb.
%The clipped black-box is also displayed (\basebb).
%
The experiments for \( \cvar{} \) and \( \EOO \) are summarized in \cref{fig:main_text_fairness}; the full plot with \( \SP \) is presented at \supplement \cref{fig:rf_acs_fairness}. For clarity we only plot the baselines and wrappers which are directly associated to each fairness criteria. We also plot the posterior mean difference between the data and debiased posteriors \meandiff (0/1 loss) to examine the effects on accuracy.

\noindent\textbf{For \( \cvar{} \)}, both conservative and audacious approaches decreases \( \cvar{} \), which results in better \( \cvar{} \) values than both the original \basebb and in-processing baseline \basecvara~-- which is good news since \basecvara~directly optimizes $\cvar{}$. We note that there are cases in which the in-processing approach is better than ours (trinary sensitive attributes in \supplement), but this is expected given \basecvara's optimization goal. Interestingly, the audacious update is superior in both \( \cvar{} \) and \meandiff~than the conservative update. This is also consistent for trinary sensitive attributes. Thus, the audacious update is desirable when optimizing \( \cvar{} \). Another observation is that only one sensitive attribute subgroup's \( \alpha \)-tree is initialized (only one `\( \times \)'). This indicates that after \( 32 \) iterations the worse case subgroup does not change in the binary case.

\noindent\textbf{For \EOO}, there is a huge difference between conservative and audacious updates as the former gets to the most fair outcomes of all baselines. Even if we used early stopping or pruning of the $\alpha$-tree (taking an earlier iterations) the audacious update would fail at producing outcomes as fair as its conservative counterpart. Furthermore, the audacious update comes with a significant degradation of accuracy \meandiff. Furthermore, by looking at the iterations in which subgroup \( \alpha \)-trees are initialized, the audacious update causes large (primarily bad) jumps in performance. This rejoins our remark on the interest of having a conservative update in Section \ref{sec-formal}. When compared to \baseeooa, we find that the conservative \topdown approach produces lower \EOO. However, \baseeooa tend to have better accuracy scores in \meandiff. These observations are consistent with the trinary sensitive attribute (\supplement).

\noindent\textbf{For \SP},  we can observe fairness results that can be on par with contenders for the conservative update, but observe a substantial degradation of MD. This, we believe, follows from a simple plug-in instantiation of $\meas{M}, \postarget$ for the fairness notion in \supplement Section \ref{sec-stat-par}, resulting in potentially harsh updates. In \supplement~(pg \pageref{sec-stat-par}), we discuss an alternative approach using ties with optimal transport.
\section{Limitations and Conclusion}

Given the context of fairness, it is important to highlight possible limitations of our approach and the potential social harm from such misuse. We highlight two of these for our \topdown approach. Firstly, our approach has shown to have failure cases in \emph{small data cases}. This can be seen in the experiments on German Credit dataset (\supplement, Section \ref{app-exp-rest}-\ref{app-exp-proxy}). This, we believe, has a formal basis in our approach. For instances, \EOO requires accurate data posterior estimation which may be difficult in small data regimes. Secondly, \topdown is of course not unilaterally better than all other post-processing fairness approaches -- there is \emph{No Free Lunch}. As such, we do not claim that our instantiation of \topdown is optimal in \( \cvar{} \), \EOO, or \SP. However, considering that such different fairness models instantiated in the \textit{same} algorithm can lead to competitive results with the respective state of the art, the avenue for improved instantiations or accurate extensions to new fairness constraints appears promising. We leave these for future work.
\section*{Acknowledgments}
YM received funding from the European Research Council (ERC) under the European Union’s Horizon 2020 research and innovation program (grant agreement No. 882396), the Israel Science Foundation(grant number 993/17), Tel Aviv University Center for AI and Data Science (TAD), and the Yandex Initiative for Machine Learning at Tel Aviv University. AS and LX thank members of the ANU Humanising Machine Intelligence program for discussions on fairness and ethical concerns in AI, and the
NeCTAR Research Cloud for providing computational resources, an Australian research platform
supported by the National Collaborative Research Infrastructure Strategy.

\bibliography{main}
\bibliographystyle{cust}

\newpage
\appendix
\onecolumn
\renewcommand\thesection{\Roman{section}}
\renewcommand\thesubsection{\thesection.\arabic{subsection}}
\renewcommand\thesubsubsection{\thesection.\thesubsection.\arabic{subsubsection}}

\renewcommand*{\thetheorem}{\Alph{theorem}}
\renewcommand*{\thelemma}{\Alph{lemma}}
\renewcommand*{\thecorollary}{\Alph{corollary}}

\renewcommand{\thetable}{A\arabic{table}}

\begin{center}
\Huge{Supplementary Material}
\end{center}

\begin{abstract}
This is the Supplementary Material to Paper "\papertitle". To
differentiate with the numberings in the main file, the numbering of Theorems is letter-based (A, B, ...).
\end{abstract}

\section*{Table of contents}

\noindent \textbf{Supplementary material on proofs and fairness models} \\
\noindent $\hookrightarrow$ Additional \textbf{Equality of Opportunity} Details \hrulefill Pg
\pageref{sec:sup_eoo}\\
\noindent $\hookrightarrow$ Handling \textbf{Statistical parity} \hrulefill Pg
\pageref{sec-stat-par}\\
\noindent $\hookrightarrow$ \textbf{Weak Hypothesis Assumption} Discussion \hrulefill Pg
\pageref{sec-wha-disc}\\
\noindent $\hookrightarrow$ General KL Distortion \ref{sec-kl-dist} \hrulefill Pg
\pageref{sec-kl-dist}\\
\noindent $\hookrightarrow$ Proof of Theorem \ref{thm-distortion-to-blackbox} \hrulefill Pg
\pageref{proof-thm-distortion-to-blackbox}\\
\noindent $\hookrightarrow$ Proof of Theorem \ref{cor-distortion-to-blackbox} and \ref{cor-distortion-to-blackbox-b} \hrulefill Pg
\pageref{proof-cor-distortion-to-blackbox}\\
\noindent $\hookrightarrow$ Proof of Theorem \ref{thm:entropyupper} and \ref{thm-boosting} \hrulefill Pg
\pageref{proof-thm-boosting}\\
\noindent $\hookrightarrow$ Proof of Theorem \ref{thm-alternative-alpha-in-alpha-tree} \hrulefill Pg
\pageref{proof-thm-alternative-alpha-in-alpha-tree}\\
\noindent $\hookrightarrow$ Proof of Lemma \ref{lem-entropies} \hrulefill Pg
\pageref{proof-lem-entropies}\\
\noindent $\hookrightarrow$ Proof of Theorem \ref{thm-eoo} \hrulefill Pg
\pageref{proof-thm-eoo}\\

\noindent \textbf{Supplementary material on experiments} \\
\noindent $\hookrightarrow$ \supplement Experiment Settings \hrulefill Pg
\pageref{app-exp-settings}\\
\noindent $\hookrightarrow$ Experiments on Statistical Parity \hrulefill Pg
\pageref{app-exp-SP-main}\\
\noindent $\hookrightarrow$ Additional Main Text Experiments \hrulefill Pg
\pageref{app-exp-rest}\\
\noindent $\hookrightarrow$ Neural Network Experiments \hrulefill Pg
\pageref{app-exp-nn}\\
\noindent $\hookrightarrow$ Proxy Sensitive Attributes \hrulefill Pg
\pageref{app-exp-proxy}\\
\noindent $\hookrightarrow$ Distribution Shift \hrulefill Pg
\pageref{app-exp-shift}\\
\noindent $\hookrightarrow$ High Clip Value \hrulefill Pg
\pageref{app-exp-clip}\\
\noindent $\hookrightarrow$ Example Alpha-Tree \hrulefill Pg
\pageref{app-exp-example}\\

\newpage

\section{Additional Equality of Opportunity Details}
\label{sec:sup_eoo}

In this section, we present additional details for the equality of opportunity (EOO) strategy presented in \cref{sec-fairness}. In particular, we present the full definition of the  $(p,\delta)$-\textit{pushup} of $\bayespo$.
\begin{definition}\label{def-pushup}
    Fix $p\in [0,1]$ and let $\mathcal{X}_p$ be a subset of $\mathcal{X}$ such that (i) $\inf \bayespo(\mathcal{X}_p) \geq \sup \bayespo(\mathcal{X}\backslash \mathcal{X}_p)$ and (ii) $\int_{\mathcal{X}_p} \dmeas{M} = p$. For any $\delta \geq 0$, the $(p,\delta)$-pushup of $\bayespo$, $\bayespo_{p,\delta}$, is the posterior defined as $\bayespo_{p,\delta} = \bayespo$ if $\inf \bayespo(\mathcal{X}_p) \geq 1/2$ and otherwise:
    \begin{eqnarray}
      \bayespo_{p,\delta}(\ve{x}) & \defeq & \left\{
                                             \begin{array}{ccl}
                                               \frac{1}{2}+\delta & \mbox{ if } \bayespo(\ve{x}) \in [\inf \bayespo(\mathcal{X}_p), 1/2 + \delta] \\
                                               \bayespo(\ve{x}) &  \multicolumn{2}{l}{\mbox{ otherwise.}}
                                               %\mbox{ if } & (\ve{x} \not\in \mathcal{X}_p) \vee \left(\bayespo(\ve{x}) \geq \frac{1}{2}+\delta\right)\\
                                               \end{array}
                                             \right. \label{eoo_mapping}
      \end{eqnarray}
\end{definition}
Following Fig.~\ref{fig:eta-diff-eoo} (right) and discussion in the main-text, we make a couple more observation. From Def.~\ref{def-pushup}, the selection of the set \( \mathcal{X}_{p} \) (\emph{i.e.} choice of \( p \)) presents a tradeoff between accuracy and the fairness objective. As \( p \) increases, the size of \( \mathcal{X}_{p} \) necessarily increases. For instance, taking \( \mathcal{X}_{p} = \{ \ve{x} : \bayespo(\ve{x}) \in > l \} \), the larger \( \mathcal{X}_{p} \) is the more negative prediction are flipped. Thus intuitively, \( p \) measures the size of correct posterior values and \( \delta \) defines the size of the correction.
%
%
%First, the interval where the 
%Notice that the transformation can introduce classification mistakes w.r.t. $\bayespo$, but only for examples with (a) small ``edge'' $|1/2 - \bayespo|$ and (b) labeled as negative on $\bayespo$ are susceptible to get positive label on $\bayespo_{p,\delta}$. Notice the tradeoff: (b) is consistent with the fairness objective while (a) limits the degradation in accuracy.
\section{Handling \textbf{Statistical parity} }\label{sec-stat-par}

\noindent\textbf{Statistical parity} (SP) is a group fairness notion \cite{dhprzFT}, implemented recently in a context similar to ours \cite{alAN} as the constraint that per-group expected treatments must not be too far from each other. We say that $\posfair$ achieves $\varepsilon$-statistical parity (across all groups induced by sensitive attribute $S$) iff
  \begin{eqnarray}
\max_{s\in S} \expect_{\X \sim \meas{M}_s} \left[\posfair(\X)\right] - \min_{s\in S} \expect_{\X \sim \meas{M}_s} \left[\posfair(\X)\right] & \leq & \varepsilon. \label{eqSTATPAR}
    \end{eqnarray}
    Denote $s^\circ \defeq \arg\min_{s\in S} \expect_{\X \sim \meas{M}_s} \left[\posfair(\X)\right], s^* \defeq \arg\max_{s\in S} \expect_{\X \sim \meas{M}_s} \left[\posfair(\X)\right]$. Since the risk we minimize in \eqref{lossgen} involves a proper loss, the most straightforward use of \topdown~is to train the sub-$\alpha$-tree for one of these two groups, giving as target posterior the \textit{expected} posterior of the other group, \textit{i.e.} we use $\postarget(\ve{x}) = \expect_{\X \sim \meas{M}_{s^*}} \left[\posunfair(\X)\right] \defeq \overline{\posunfair}_{s^*}$ if we grow the $\alpha$-tree of $s^\circ$ and thus iterate
\begin{tcolorbox}[colframe=blue,boxrule=0.5pt,arc=4pt,left=6pt,right=6pt,top=6pt,bottom=6pt,boxsep=0pt]
  \begin{center}
\topdown~with $\meas{M}_{\mbox{\tiny{\textup{t}}}} \leftarrow \meas{M}_{s^\circ}$ and $\postarget \leftarrow \overline{\posunfair}_{s^*}$,
    \end{center}
  \end{tcolorbox}
  and we repeat until $s^\circ$ does not achieve anymore the smallest expected posterior. We then update the group and repeat the procedure until a given slack $\varepsilon$ is achieved between the extremes in \eqref{eqSTATPAR}. More sophisticated / gentle approaches are possible, including using the links between statistical parity and optimal transport (OT, \citet[Section 3.2]{dhprzFT}), suggesting to use as target posterior the expected posterior obtained from an OT plan between groups $s^\circ$ and $s^*$.

\section{\textbf{Weak Hypothesis Assumption} Discussion}\label{sec-wha-disc}

In this Section, we discuss the balanced distribution introduced in Definition \ref{balancedDist} and the Weak Hypothesis Assumption (WHA) introduced in Assumption \ref{defWHA}. In particular, we note that these definitions are a generalization of those introduced in ``classical'' decision tree boosting \cite{kmOT}. The primary difference in our case and the ``classical'' case, is that we must consider the posterior values of a base black-box \( \posunfair \), whereas the usual top-down tree induction only cares about minimizing its loss function to produce an accurate \( \postarget \). In-fact, we can even interpret our variants as a case where we replace the labels in a classification task \( \Y \) with ``labels'' (in \( [0, 1]\)) of how well another classifier does in classification \( \Y \normalizedlogit(\posunfair(\X)) \). We will explore various cases to show that our definitions reduce to the classical case; and further provide settings to intuitive describe our more complicated assumption.

\noindent\textbf{The balanced distribution} in Definition \ref{balancedDist} acts as a re-weighting mechanism for particular samples of the original distribution \( {\measdtarget}_\leaf (z) \). Let us first consider the reduction to the balanced distribution introduced by \citet{kmOT}. This comes from the fairness-free case, where we ignore the confidence of the black-box \( \posunfair \) and set \( \normalizedlogit(\posunfair) \leftarrow 1 \). This provides two simplifications. Firstly, the edge value becomes \( \edge(\meas{M}_\leaf, \postarget) = \expect_{(\X, \Y) \sim {\measdtarget}_\leaf} \left[\Y\right] = 2q_{\leaf} - 1 \), where $q_{\leaf}$ the local proportion of positive examples in $\leaf$. That is, we have \( q_{\leaf} = \pr(\Y = +1 | \X \textrm{ at leaf } \leaf) \). The second simplification is in the numerator of Equation \ref{defDMEASTARGET2}. In summary, we get:
\begin{equation} \label{eq:balanced_kmot}
    {\rebmeasdtarget}_\leaf (z) \leftarrow \frac{1 - \edge(\meas{M}_\leaf, \postarget) \cdot y}{1 - \edge(\meas{M}_\leaf, \postarget)^2} \cdot {\measdtarget}_\leaf (z) = \frac{1 - y\cdot(2q_{\leaf} - 1)}{4q_{\leaf}(1 - q_{\leaf})} \cdot {\measdtarget}_\leaf (z) = 
    {\measdtarget}_\leaf (z) \cdot 
    \begin{cases}
        \frac{1}{2q_{\leaf}} & \textrm{if } y = +1 \\
        \frac{1}{2(1 - q_{\leaf})} & \textrm{if } y = -1
    \end{cases},
\end{equation}
which is indeed the balanced distribution of \citet{kmOT}. Intuitively, this balanced distribution simply ensures that regardless of \( \ve{x} \), it is equally likely for \( y \) be either \( -1 \) or \( +1 \) (hence balanced in its prediction value \( \Y \)). This can be seen by calculating the edge with the new distribution \( \expect_{(\X, \Y) \sim {\rebmeasdtarget}_{\leaf}}[\Y] = 0 \).

To consider our balanced distribution, let us start with \eqref{eq:balanced_kmot}. In particular, we will replace the labels \( \Y \in \{ -1, +1 \} \) by the predictive ability of black-box \( \posunfair \). This predictive ability is summarized by \( \Y \normalizedlogit(\posunfair(\X)) \in [-1, 1] \), where larger is better. In particular, taking from the boosting jargon, the  \textit{confidences} ($|\normalizedlogit|$, \citet{ssIBj}) of the black-box \( \posunfair \) are considered. For instance, a highly confident \( \vert \normalizedlogit(\posunfair(\X)) \vert \rightarrow 1 \) and correct \( \mathrm{sign}(\Y) = \mathrm{sign}(\logit(\posunfair(\X))) \) prediction will lead to a value close to \( +1 \) (positive label). If the prediction is confident but incorrect \( \mathrm{sign}(\Y) \neq \mathrm{sign}(\logit(\posunfair(\X))) \), then the value will be close to \( -1 \) (negative label). If the prediction is not confident \( \vert \normalizedlogit(\posunfair(\X)) \vert \rightarrow 0 \), then we get a neutral response \( 0 \).

With this in mind, the balanced distribution in Definition \ref{balancedDist} is ``balanced'' in the prediction score \( \Y \normalizedlogit(\posunfair(\X)) \). Of course, the distribution \( {\rebmeasdtarget}_\leaf (z) \) is with respect to \( \mathcal{X} \times\mathcal{Y} \) and not \( \mathcal{X} \times [-1, 1] \), and thus we are not re-balancing for each possible \( \Y \normalizedlogit(\posunfair(\X)) \in [-1, 1] \). The balance of the distribution takes into account the confidence of the predictions, this can be seen in examination of the balanced edge in this setting:
\begin{equation} \label{eq:bal_new_edge}
    \expect_{(\X, \Y) \sim {\rebmeasdtarget}_{\leaf}}[\Y \cdot \normalizedlogit(\posunfair(\X)) ] = \frac{1}{1 - \edge(\meas{M}_\leaf, \postarget)^{2}} \cdot \edge(\meas{M}_\leaf, \postarget) \cdot \expect_{(\X, \Y) \sim {\measdtarget}_{\leaf}}[1 - \normalizedlogit^{2}(\posunfair(\X))].
\end{equation}
\eqref{eq:bal_new_edge} is in general non-zero. However, in the case that the \( \posunfair \) is confident such that \( \normalizedlogit(\posunfair(\X)) \in \{ -1 , +1 \} \), then the new edge goes to 0. In other words, when we have a maximally confident black-box, the balanced edge will be 0; otherwise, the edge is scaled with respect to a function of the confidence \( \expect_{(\X, \Y) \sim {\measdtarget}_{\leaf}}[1 - \normalizedlogit^{2}(\posunfair(\X))] \).

\noindent\textbf{Our Weak Learning Assumption (WHA)} in Assumption \ref{defWHA} has a similar analysis as that of the balanced distribution. Condition (i) of the assumption is exactly the same \( \gamma \)-witness condition in \citet{kmOT}. Condition (ii), similar to the analysis of \eqref{eq:bal_new_edge}, vanishes when we have a maximally confident black-box \( \posunfair \). In general, condition (ii) ensures that one side of the split induced by \( h \) are skewed to one side.

Let us exemplify how our WHA works if we have a leaf $\leaf$ where local ``treatments due to the black-box" are bad ($y \normalizedlogit(\posunfair(\ve{x})) < 0$ often). In such a case, $\edge(\meas{M}_\leaf, \postarget)  < 0$ so the balanced distribution (Definition \ref{balancedDist}) reweights higher examples whose treatment is better than average, \textit{i.e.} the local minority. Suppose (i) holds as is without the $|.|$. In such a case, the split ``aligns" the treatment quality with $h$, so $h = +1$ for a substantial part of this minority. (ii) imposes $\expect_{(\X, \Y) \sim {\measdtarget}_{\leaf}} \left[h(\X)\right] \geq \expect_{(\X, \Y) \sim {\measdtarget}_{\leaf}} \left[ \normalizedlogit^2(\posunfair(\X)) \cdot h(\X)\right] $: $h = -1$ for a substantial part of large confidence treatment. The split thus tends to separate mostly large confidence but bad treatments (left) and mostly good treatments (right). Before the split, the value $\alphatree(\leaf)$ would be negative \eqref{alphatreeoutputCONS} and thus reverse the polarity of the black-box, which would be good for badly treated examples but catastrophic for the local minority of adequately treated examples. After the split however, we still have the left ($h = -1$) leaf where this would eventually happen, but the minority at $\leaf$ would have disproportionately ended in the right ($h = +1$) leaf, where it would be likely that $\alphatree(.)$ would this time be \textit{positive} and thus preserve the polarity of the treatment of the black-box.

\noindent\textbf{Why are the assumptions states using the balanced distribution?} A point worth clarifying is the ``why's'' of consider a WHA in the way it is formulated and the focus on the balanced distribution. In typical boosting algorithms, a weak learning hypothesis is needed: the assumption that a classifier / split / weak learner exists for \emph{any} distribution which is better than random. In top-down tree induction, it is in-fact sufficient to only have this assumption hold for one type of distribution --- the balanced distribution \cite{kmOT}. As such the WHA presented, is actually a weaker assumption than the ''distributionally global'' weak learning assumption used in other boosting algorithms.
\section{General KL Distortion}\label{sec-kl-dist}

We introduce a general Theorem for upperbounding the distortion an \( \alpha \)-correction can make, as promised in Section~\ref{sec-formal}. Notably, the following result does not rely on any setting.

  \begin{theorem}\label{thm-distortion-to-blackbox}
 For any function $\alpha:\mathcal{X} \rightarrow \mathbb{R}$, any black-box posterior $\posunfair$, and any integer $K\geq 2$, using \eqref{ourposfair} yields the following bound on the KL divergence:
 \begin{equation}
  \kl(\posunfair, \posfair; \meas{M})
  \leq \expect_{\X \sim \meas{M}} \left[\sum_{k= 2}^K \frac{\posunfair(\X)(1-\posunfair(\X)) f^k (\alpha(\X),\posunfair(\X)) }{k(k-1)}\right]
  + o\left(\expect_{\X \sim \meas{M}} \left[(\alpha(\X)-1)^K\right]\right), \label{boundDelta1}
 \end{equation}
%\begin{eqnarray}
%  \lefteqn{\kl(\posunfair, \posfair; \meas{M})}\nonumber\\
%  & \leq & \expect_{\X \sim \meas{M}} \left[\sum_{k= 2}^K \frac{\posunfair(\X)(1-\posunfair(\X)) f^k (\alpha(\X),\posunfair(\X)) }{k(k-1)}\right] \nonumber\\
%  & & + o\left(\expect_{\X \sim \meas{M}} \left[(\alpha(\X)-1)^K\right]\right), \label{boundDelta1}
%\end{eqnarray}
where we have used function $f: \mathbb{R} \times [0,1]\rightarrow \mathbb{R}$ defined as:
  \negativespace
\begin{eqnarray}
f(z,u) & \defeq & \left| \log \left(u/(1-u)\right) \cdot (1-z)\right| = |\logit(u) \cdot (1 - z)|.\label{defFKL}
  \negativespace
  \end{eqnarray}
\end{theorem}
The proof is presented in Section \ref{proof-thm-distortion-to-blackbox}.
In general, one can see that if \( \alpha(\cdot) \) does not differ from \( 1 \) too much, the KL divergence will be small. This intuitively makes sense as \( \alpha = 1 \) causes \( \posfair = \posunfair \) --- there is no untwisting.
Theorem \ref{thm-distortion-to-blackbox} can be further weakened to provide easier to understand bounds, as per Corollary~\ref{cor-distortion-to-blackbox}.

We further present an alternative setting to that in the main-text, which provides another bound on distortion. In this setting \textbf{(S2)}, we ensure that when the predictions of \( \posunfair \) are confident, then the corresponding \( \alpha \) correction are small (close to \( \alpha = 1 \)).
\begin{enumerate}
\item [\textbf{(S2)}] 
$ f(\alpha(\ve{x}), \posunfair(\ve{x})) = \left| \logit(\posunfair(\ve{x})) \cdot (1 - \alpha(\ve{x}))\right| \leq 1$
(a.s.)
%$ \left| \log \left((1-\posunfair(\ve{x}))/\posunfair(\ve{x})\right) \cdot (\alpha(\ve{x})-1)\right| \leq 1$
%$f(\alpha(\ve{x}),\posunfair(\ve{x})) \leq 1$
, $f$ being in \eqref{defFKL}.
\negativespace
\end{enumerate}
This provides the alternative upperbound on the distortion.
\begin{corollary}\label{cor-distortion-to-blackbox-b}
  Under setting \textbf{(S2)}, we have the upperbound
  \begin{eqnarray}
  \kl(\posunfair, \posfair; \meas{M}) & \leq & \pi^2/24 \approx 0.41.\label{boundDelta2}
  \end{eqnarray}
\end{corollary}
The proof of the Corollary is in \supplement, Section \ref{proof-cor-distortion-to-blackbox} and includes a graphical view of the values of \( \posunfair(.) \) and \( \alpha(.) \) complying with \textbf{(S2)}.

\section{Proof of Theorem \ref{thm-distortion-to-blackbox}}\label{proof-thm-distortion-to-blackbox}

We first show two technical Lemmata.
  \begin{lemma}\label{lemTE}
    For any $a \geq 0$, let
    \begin{eqnarray}
h(z) & \defeq & \log\left(\frac{1}{1+a^{1+z}}\right) .
    \end{eqnarray}
    We have
    \begin{eqnarray}
h^{(k)}(z) & = & -\frac{\log^k (a) \cdot a^{1+z}}{(1+a^{1+z})^{k}}\cdot P_{k-1}(a^{1+z}),
      \end{eqnarray}
    where $P_k(x)$ is a degree-$k-1$ polynomial. Letting $c_{k,j}$ the constant factor of monomial $x^j$ in $P_k(x)$, for $j\leq k-1$, we have the following recursive definitions: $c_{1,0} = 1$ ($k=1$) and 
    \begin{eqnarray}
      c_{k+1,k} & = & (-1)^{k},\\
      c_{k+1,j} & = & (j+1) \cdot c_{k,j} - (k+1-j)\cdot c_{k,j-1}, \forall 0<j<k,\\
      c_{k+1,0} & = & 1.
    \end{eqnarray}
    Hence, we have for example $P_1(x) = 1, P_2(x) = -x + 1, P_3(x) = x^2 - 4x + 1, P_4(x) = -x^3 + 11x^2 - 11x + 1, ...$.
  \end{lemma}
  \begin{myproof}
    We let
    \begin{eqnarray}
 f(z) & \defeq & \frac{a^{1+z}}{1+a^{1+z}},
    \end{eqnarray}
    so that $h'(z) = - \log(a) \cdot g(z)$ and we show
    \begin{eqnarray}
f^{(k)}(z) & = & \frac{\log^k (a) \cdot a^{1+z}}{(1+a^{1+z})^{k+1}}\cdot P_k(a^{1+z}).\label{propGder}
    \end{eqnarray}
    We first check
    \begin{eqnarray}
f'(z) & = & \frac{\log(a) \cdot a^{1+z}}{(1+a^{1+z})^2},
    \end{eqnarray}
    which shows $P_1(x) = 1$. We then note that for any $k\in \mathbb{N}_*$,
    \begin{eqnarray}
\frac{\mathrm{d}}{\mathrm{d} z} \frac{a^{1+z}}{(1+a^{1+z})^k} & = & \frac{\log(a) \cdot a^{1+z}}{(1+a^{1+z})^{k+1}}\cdot(-(k-1) a^{1+z} + 1),
    \end{eqnarray}
    so the induction case yields $f^{(k+1)}(z) \defeq {f^{(k)}}'(z)$, that is:
    \begin{eqnarray}
      \lefteqn{f^{(k+1)}(z)}\nonumber\\
      & = & \log^k (a) \cdot \frac{\mathrm{d}}{\mathrm{d} z} \left(\frac{a^{1+z}}{(1+a^{1+z})^{k+1}}\cdot P_k(a^{1+z})\right)\nonumber\\
      & = & \log^k (a) \cdot \left(\frac{\log(a) \cdot a^{1+z}}{(1+a^{1+z})^{k+2}}\cdot(-k a^{1+z} + 1)\cdot P_k(a^{1+z}) + \frac{a^{1+z}\cdot \log(a) }{(1+a^{1+z})^{k+1}}\cdot a^{1+z}\cdot \left.\frac{\mathrm{d} P_k(x) }{\mathrm{d} x}\right|_{x=a^{1+z}}\right)\nonumber\\
      & = & \frac{\log^{k+1} (a) \cdot a^{1+z}}{(1+a^{1+z})^{k+2}}\cdot \underbrace{\left((-k a^{1+z} + 1)\cdot P_k(a^{1+z}) + a^{1+z}(1+a^{1+z})\cdot \left.\frac{\mathrm{d} P_k(x) }{\mathrm{d} x}\right|_{x=a^{1+z}}\right)}_{\defeq P_{k+1}(a^{1+z})},
    \end{eqnarray}
    from which we check that $P_{k+1}$ is indeed a polynomial and its coefficients are obtained via identification from $P_k$, which establishes \eqref{propGder} and yields to the statement of the Lemma.
  \end{myproof}
  \begin{lemma}\label{lemCK}
    Coefficient $c_{k,j}$ admits the following bound, for any $0\leq j\leq k$:
    \begin{eqnarray}
        |c_{k,j}| & \leq & (k-1)! \binom{k-1}{j} .
      \end{eqnarray}
    \end{lemma}
  \begin{myproof}
    First, we have the following recursive definition for the absolute value of the leveraging coefficients in $c_{.,.}$ (we call them $a_{.,.}$ for short): $|c_{.,.}| = a_{.,.}$ with
    \begin{eqnarray}
      a_{k+1,k} & = & 1,\\
      a_{k+1,j} & = & (j+1) \cdot a_{k,j} + (k+1-j)\cdot a_{k,j-1}, \forall 0<j<k,\\
      a_{k+1,0} & = & 1.
    \end{eqnarray}
    We now show by induction that $a_{k+1,j} \leq k! \binom{k}{j} \defeq b_{k+1, j}$. For $j=0$, $b_{k+1,0} = k ! \geq a_{k+1,0}$ ($k\geq 2$) and for $j=k$, $b_{k+1,k} = k! \geq a_{k+1,0}$ as well. We now check, assuming the property holds at all ranks $k$, that for ranks $k+1$, we have
    \begin{eqnarray}
      a_{k+1,j} & = & (j+1) \cdot a_{k,j} + (k+1-j)\cdot a_{k,j-1}\nonumber\\
      & \leq & (j+1) (k-1)! \binom{k-1}{j} + (k+1-j)(k-1)!  \binom{k-1}{j-1},
    \end{eqnarray}
    and we want to check that the RHS is $\leq k! \binom{k}{j}$ for any $0<j<k$. Simplifying yields the equivalent inequality
    \begin{eqnarray}
(j+1)(k-j) + (k+1-j)j & \leq & k^2.
    \end{eqnarray}
    finding the worst case bound for $j$ yields $j = k/2$ (we disregard the fact that $j$ is an integer) and plugging in the bound yields the constraint on $k$: $k \geq 2$, which indeed holds.
\end{myproof}\\
   We also check that $h$ in Lemma \ref{lemTE} is infinitely differentiable. As a consequence, we get from Lemma \ref{lemTE} the Taylor expansion around $g=1$ (for any $a\geq 0$) at any order $K\geq 2$,
  \begin{eqnarray}
    \lefteqn{\log\left(\frac{1}{1+a^g}\right)}\nonumber\\
    & = & \log\left(\frac{1}{1+a}\right) - \frac{a \log a}{1+a}\cdot (g-1) - \underbrace{\sum_{k=2}^K \frac{a \log^k (a)  P_{k-1}(a) }{k! (1+a)^{k}}\cdot (g-1)^k}_{\defeq R_{K,a}(g)} + o((g-1)^K).\label{taylExp1}
  \end{eqnarray}
The choice to start the summation at $k=2$ is done for technical simplifications to come. We thus have
    \begin{eqnarray*}
      \log \posfair(\ve{x}) & = & \log\left(\frac{1}{1+\left(\frac{1-\posunfair(\ve{x})}{\posunfair(\ve{x})}\right)^{\alpha(\ve{x})}}\right)\\
                               & = & \log \posunfair(\ve{x}) - (1-\posunfair(\ve{x})) \log\left(\frac{1-\posunfair(\ve{x})}{\posunfair(\ve{x})}\right) \cdot ({\alpha(\ve{x})}-1) - R_{\frac{1-\posunfair(\ve{x})}{\posunfair(\ve{x})},K}({\alpha(\ve{x})}) \\
      & & + o((\alpha(\ve{x})-1)^K),\\
     \log (1-\posfair(\ve{x})) & = & \log (1-\posunfair(\ve{x})) - \posunfair(\ve{x}) \log\left(\frac{\posunfair(\ve{x})}{1-\posunfair(\ve{x})}\right) \cdot ({\alpha(\ve{x})}-1) - R_{\frac{\posunfair(\ve{x})}{1-\posunfair(\ve{x})},K}({\alpha(\ve{x})})\\
      & & + o((\alpha(\ve{x})-1)^K).
    \end{eqnarray*}
    Define for short $\Delta_{\mbox{\tiny{u}}}(\ve{x}) \defeq \posunfair(\ve{x})\cdot - \log \posfair(\ve{x}) + (1-\posunfair(\ve{x}))\cdot - \log (1-\posfair(\ve{x})) - (\posunfair(\ve{x})\cdot - \log \posunfair(\ve{x}) + (1-\posunfair(\ve{x}))\cdot - \log (1-\posunfair(\ve{x})))$, so that $\kl(\posunfair, \posfair; \meas{M}) = \expect_{\X \sim \meas{M}} \left[\Delta_{\mbox{\tiny{u}}}(\X)\right]$. The Taylor expansion \eqref{taylExp1} unveils an interesting simplification:
    \begin{eqnarray*}
      \Delta_{\mbox{\tiny{u}}}(\ve{x}) & = & -\posunfair(\ve{x})\log \posunfair(\ve{x}) +\posunfair(\ve{x}) (1-\posunfair(\ve{x})) \log\left(\frac{1-\posunfair(\ve{x})}{\posunfair(\ve{x})}\right) \cdot ({\alpha(\ve{x})}-1) \\
      & & +\posunfair(\ve{x})\cdot R_{\frac{1-\posunfair(\ve{x})}{\posunfair(\ve{x})},K}({\alpha(\ve{x})})\\
                             & & - (1-\posunfair(\ve{x}))\log (1-\posunfair(\ve{x})) + (1-\posunfair(\ve{x}))\posunfair(\ve{x}) \log\left(\frac{\posunfair(\ve{x})}{1-\posunfair(\ve{x})}\right) \cdot ({\alpha(\ve{x})}-1) \\
                             & & +(1-\posunfair(\ve{x}))\cdot R_{\frac{\posunfair(\ve{x})}{1-\posunfair(\ve{x})},K}({\alpha(\ve{x})})\\
                             & & - (\posunfair(\ve{x})\cdot - \log \posunfair(\ve{x}) + (1-\posunfair(\ve{x}))\cdot - \log (1-\posunfair(\ve{x})))+ o((\alpha(\ve{x})-1)^K)\\
      & = & \posunfair(\ve{x})\cdot R_{\frac{1-\posunfair(\ve{x})}{\posunfair(\ve{x})}}({\alpha(\ve{x})}) +(1-\posunfair(\ve{x}))\cdot R_{\frac{\posunfair(\ve{x})}{1-\posunfair(\ve{x})},K}({\alpha(\ve{x})}) + o((\alpha(\ve{x})-1)^K), \forall \ve{x} \in \mathcal{X},
      \end{eqnarray*}
so the divergence to the black-box prediction simplifies as well, this time using Lemma \ref{lemCK}: 
  \begin{eqnarray}
    \kl(\posunfair, \posfair; \meas{M}) & = & \expect_{\X \sim \meas{M}} \left[\posunfair(\X)\cdot R_{\frac{1-\posunfair(\X)}{\posunfair(\X)},K}({\alpha(\X)}) +(1-\posunfair(\X))\cdot R_{\frac{\posunfair(\X)}{1-\posunfair(\X)},K}({\alpha(\X)})\right] \nonumber\\
    & & + o\left(\expect_{\X \sim \meas{M}} \left[(\alpha(\X)-1)^K\right]\right).\label{deltaug}
  \end{eqnarray}
  Not touching the little-oh term, we simplify further and bound the term in the expectation: for any $\ve{x} \in \mathcal{X}$,
  \begin{eqnarray}
    \lefteqn{\posunfair(\ve{x})\cdot R_{\frac{1-\posunfair(\ve{x})}{\posunfair(\ve{x})},K}({\alpha(\ve{x})}) +(1-\posunfair(\ve{x}))\cdot R_{\frac{\posunfair(\ve{x})}{1-\posunfair(\ve{x})},K}({\alpha(\ve{x})})} \nonumber\\
    & = & \posunfair(\ve{x})\cdot \sum_{k=2}^K \frac{\frac{1-\posunfair(\ve{x})}{\posunfair(\ve{x})}\cdot \log^k \left(\frac{1-\posunfair(\ve{x})}{\posunfair(\ve{x})}\right)  P_{k-1}\left(\frac{1-\posunfair(\ve{x})}{\posunfair(\ve{x})}\right) }{k! \left(1+\frac{1-\posunfair(\ve{x})}{\posunfair(\ve{x})}\right)^{k}}\cdot ({\alpha(\ve{x})}-1)^k\nonumber\\
                                 & & + (1-\posunfair(\ve{x}))\cdot \sum_{k=2}^K \frac{\frac{\posunfair(\ve{x})}{1-\posunfair(\ve{x})}\cdot \log^k \left(\frac{\posunfair(\ve{x})}{1-\posunfair(\ve{x})}\right)  P_{k-1}\left(\frac{\posunfair(\ve{x})}{1-\posunfair(\ve{x})}\right) }{k! \left(1+\frac{\posunfair(\ve{x})}{1-\posunfair(\ve{x})}\right)^{k}}\cdot ({\alpha(\ve{x})}-1)^k\nonumber\\
    & = & \posunfair(\ve{x})\cdot \sum_{k=2}^K \frac{\frac{1-\posunfair(\ve{x})}{\posunfair(\ve{x})}\cdot \log^k \left(\frac{1-\posunfair(\ve{x})}{\posunfair(\ve{x})}\right)  \cdot \sum_{j=0}^{k-2} c_{k-1,j}\left(\frac{1-\posunfair(\ve{x})}{\posunfair(\ve{x})}\right)^j }{k! \left(1+\frac{1-\posunfair(\ve{x})}{\posunfair(\ve{x})}\right)^{k}}\cdot ({\alpha(\ve{x})}-1)^k\nonumber\\
                                 & & + (1-\posunfair(\ve{x}))\cdot \sum_{k=2}^K \frac{\frac{\posunfair(\ve{x})}{1-\posunfair(\ve{x})}\cdot \log^k \left(\frac{\posunfair(\ve{x})}{1-\posunfair(\ve{x})}\right)  \cdot \sum_{j=0}^{k-2} c_{k-1,j}\left(\frac{\posunfair(\ve{x})}{1-\posunfair(\ve{x})}\right)^j }{k! \left(1+\frac{\posunfair(\ve{x})}{1-\posunfair(\ve{x})}\right)^{k}}\cdot ({\alpha(\ve{x})}-1)^k\nonumber\\
    & = & \sum_{k=2}^K \frac{\log^k \left(\frac{1-\posunfair(\ve{x})}{\posunfair(\ve{x})}\right)  \cdot \sum_{j=0}^{k-2} c_{k-1,j} \cdot \posunfair^{k-j}(\ve{x}) (1-\posunfair(\ve{x}))^{j+1}}{k!}\cdot ({\alpha(\ve{x})}-1)^k\nonumber\\
                                 & & +  \sum_{k=2}^K \frac{\log^k \left(\frac{\posunfair(\ve{x})}{1-\posunfair(\ve{x})}\right)  \cdot \sum_{j=0}^{k-2} c_{k-1,j} \cdot (1-\posunfair(\ve{x}))^{k-j} \posunfair^{j+1}(\ve{x}) }{k!}\cdot ({\alpha(\ve{x})}-1)^k\label{eqBDeltaX}
  \end{eqnarray}
  We now note, using Lemma \ref{lemCK} that for any $\ve{x} \in \mathcal{X}$,
\begin{eqnarray*}
  \lefteqn{\sum_{j=0}^{k-2} |c_{k-1,j}| \cdot \posunfair^{k-j}(\ve{x}) (1-\posunfair(\ve{x}))^{j+1}}\nonumber\\
  & = & \posunfair^{2}(\ve{x})(1-\posunfair(\ve{x}))\cdot \sum_{j=0}^{k-2} |c_{k-1,j}| \cdot \posunfair^{k-2-j}(\ve{x}) (1-\posunfair(\ve{x}))^{j}\\
  & \leq &  \posunfair^{2}(\ve{x})(1-\posunfair(\ve{x}))\cdot \sum_{j=0}^{k-2}(k-2)! \binom{k-2}{j} \posunfair^{k-2-j}(\ve{x}) (1-\posunfair(\ve{x}))^{j}\nonumber \\
  & & = (k-2)!\cdot \posunfair^{2}(\ve{x})(1-\posunfair(\ve{x}))\cdot \underbrace{\sum_{j=0}^{k-2} \binom{k-2}{j} \posunfair^{k-2-j}(\ve{x}) (1-\posunfair(\ve{x}))^{j}}_{=(1-\posunfair(\ve{x}) + \posunfair(\ve{x}))^{k-2} = 1}\nonumber\\
                                                                                   & = & (k-2)! \cdot \posunfair^{2}(\ve{x})(1-\posunfair(\ve{x})),
\end{eqnarray*}
and similarly
\begin{eqnarray*}
  \sum_{j=0}^{k-2} |c_{k-1,j}| \cdot (1-\posunfair(\ve{x}))^{k-j} \posunfair^{j+1}(\ve{x}) & \leq & (k-2)! \cdot \posunfair(\ve{x})(1-\posunfair(\ve{x}))^{2},
\end{eqnarray*}
so plugging the two last bounds on \eqref{eqBDeltaX} yields the bound on $\kl(\posunfair, \posfair; \meas{M})$ from \eqref{deltaug}:
\begin{eqnarray}
  \kl(\posunfair, \posfair; \meas{M}) & \leq & \expect_{\X \sim \meas{M}} \left[\sum_{k= 2}^K \frac{(\posunfair^{2}(\X)(1-\posunfair(\X))+\posunfair(\X)(1-\posunfair(\X))^2) \left| \log \left(\frac{1-\posunfair(\X)}{\posunfair(\X)}\right)\right|^k }{k(k-1)}\cdot |{\alpha(\X)}-1|^k\right] \nonumber\\
  & & + o\left(\expect_{\X \sim \meas{M}} \left[(\alpha(\X)-1)^K\right]\right)\nonumber\\
                                      & = & \expect_{\X \sim \meas{M}} \left[\sum_{k= 2}^K \frac{\posunfair(\X)(1-\posunfair(\X))\left| \log \left(\frac{1-\posunfair(\X)}{\posunfair(\X)}\right)\right|^k }{k(k-1)}\cdot |{\alpha(\X)}-1|^k\right] \nonumber\\
  & & + o\left(\expect_{\X \sim \meas{M}} \left[(\alpha(\X)-1)^K\right]\right),
\end{eqnarray}
which yields the statement of Theorem \ref{thm-distortion-to-blackbox}.

\section{Proof of Theorem \ref{cor-distortion-to-blackbox} and \ref{cor-distortion-to-blackbox-b}}\label{proof-cor-distortion-to-blackbox}

We start by \textbf{(S1)}. We study function
\begin{eqnarray}
f_k(u) & \defeq & u(1-u) \left|\log\left(\frac{1-u}{u}\right)\right|^k, \forall u \in \left[\frac{1}{1+\exp (B)}, \frac{1}{1+\exp (-B)}\right] .\label{defFK}
\end{eqnarray}
$f_k$ being symmetric around $u=1/2$ and zeroing in $1/2$, we consider wlog $u< 1/2$ to find its maximum, so we can drop the absolute value. 
We have
\begin{eqnarray}
f_k'(u) & = & \log^{k-1}\left(\frac{1-u}{u}\right)\cdot\left((1-2u)\cdot \log\left(\frac{1-u}{u}\right)-k\right).
\end{eqnarray}
Function $u \mapsto (1-2u)\cdot \log\left(\frac{1-u}{u}\right)$ is strictly decreasing on $(0,1/2)$ and has limit $+\infty$ on $0^+$, so the unique maximum of $f$ on $[0,1/2)$ (we close by continuity the interval in $0$ since $\lim_{0^+} f = 0$) is attained at the only solution $u_k$ of
\begin{eqnarray}
(1-2u_k)\cdot \log\left(\frac{1-u_k}{u_k}\right) & = & k,\label{eqUK}
\end{eqnarray}
and such a solution always exist for any $k \ll \infty$. It also follows $u_{k+1} < u_k$, so if we denote as $k^*$ the smallest $k$ such that
\begin{eqnarray}
u_{k^*} & \leq & \frac{1}{1+\exp (B)},\label{eqUKStar}
\end{eqnarray}
then we will have the upperbound:
\begin{eqnarray}
  f_k(u) & \leq & \frac{1}{1+\exp (B)} \cdot \frac{1}{1+\exp (-B)} \cdot B^k\nonumber\\
  & & = \frac{B^k}{2 + \exp (B) + \exp (-B)}, \forall k \geq k^*.
  \end{eqnarray}
  We can also compute $k^*$ exactly as it boils down to taking the integer part of the solution of \eqref{eqUK} where $u_k$ is picked as in \eqref{eqUKStar}:
  \begin{eqnarray}
k^* & = & \left\lfloor \frac{\exp(B) - 1}{\exp(B) + 1} \cdot B \right\rfloor,
  \end{eqnarray}
to get $k^* = 2$, it is sufficient that $B\leq 3$, which thus gives:
  \begin{eqnarray}
  \kl(\posunfair, \posfair; \meas{M}) & \leq & \sum_{k= 2}^K \frac{\expect_{\X \sim \meas{M}} \left[(B\cdot |{\alpha(\X)}-1|)^k\right]}{(2 + \exp (B) + \exp (-B))k(k-1)} + G,
  \end{eqnarray}
and if $|{\alpha(\ve{x})}-1| \leq 1/B = 1/3, \forall \ve{x} \in \mathcal{X}$, then we can include all terms for all $k\geq 2$ in the upperbound, which makes the little-oh remainder vanish and we get:
\begin{eqnarray}
  \kl(\posunfair, \posfair; \meas{M}) & \leq & \lim_{K\rightarrow +\infty} \frac{1}{2 + \exp (B) + \exp (-B)} \cdot \sum_{k= 2}^K \frac{1}{k(k-1)} \\
                                   & \leq & \frac{1}{2 + \exp (B) + \exp (-B)} \cdot \sum_{k\geq 1} \frac{1}{k^2} \\
  & = & \frac{\pi^2}{6(2 + \exp (B) + \exp (-B))},
\end{eqnarray}
which is \eqref{boundDelta1} and proves the Corollary for setting \textbf{(S1)}. The proof for setting \textbf{(S2)} is direct as in this case we get:
\begin{eqnarray}
  \kl(\posunfair, \posfair; \meas{M}) & \leq & \lim_{K\rightarrow +\infty} \expect_{\X \sim \meas{M}} \left[\sum_{k= 2}^K \frac{\posunfair(\X)(1-\posunfair(\X)) f^k (\alpha(\X),\posunfair(\X)) }{k(k-1)}\right]\nonumber\\
  & & = \expect_{\X \sim \meas{M}} \left[\sum_{k= 2}^K \frac{\posunfair(\X)(1-\posunfair(\X)) f^k (\alpha(\X),\posunfair(\X)) }{k(k-1)}\right]\nonumber\\
                                         & \leq & \expect_{\X \sim \meas{M}} \left[\sum_{k= 2}^K \frac{\posunfair(\X)(1-\posunfair(\X))}{k(k-1)}\right]\\
  & \leq & \frac{1}{4} \cdot \sum_{k= 2}^K \frac{1}{k(k-1)} \\
                                   & \leq & \frac{1}{4} \cdot \sum_{k\geq 1} \frac{1}{k^2} \\
  & = & \frac{\pi^2}{24}, 
\end{eqnarray}
as claimed.\\

\begin{figure}[h]
    \centering
    \centerline{\includegraphics[trim=20bp 550bp 540bp 10bp,clip,width=0.7\textwidth]{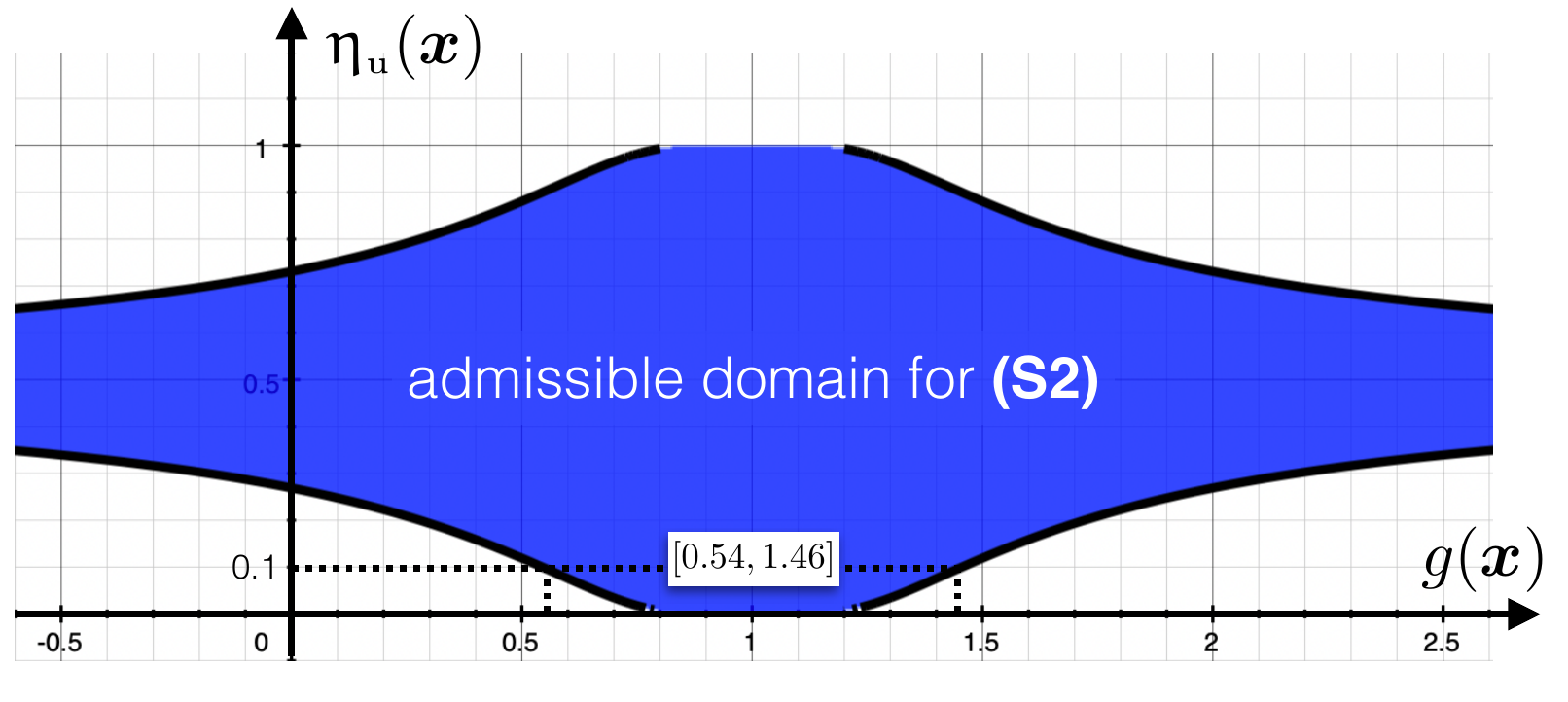}}
    \caption{Admissible couples of values $(g, \posunfair)$ (in blue) complying with setting \textbf{(S2)}. For example, any couple $(g,0.1)$ with $g \in [0.54, 1.46]$ is admissible.}
     \label{fig:authorized-range}
\end{figure}
Figure \ref{fig:authorized-range} provides an idea of the set of \textit{admissible} couples (correction, black-box posterior) that comply with \textbf{(S2)}, from which we see that the range of admissible corrections is quite flexible, even when $\posunfair$ comes quite close to $\{0,1\}$.

\section{Proof of Theorem \ref{thm:entropyupper} and \ref{thm-boosting}}\label{proof-thm-boosting}

\begin{figure}[h]
    \centering
    \centerline{\includegraphics[trim=20bp 580bp 700bp 40bp,clip,width=0.8\textwidth]{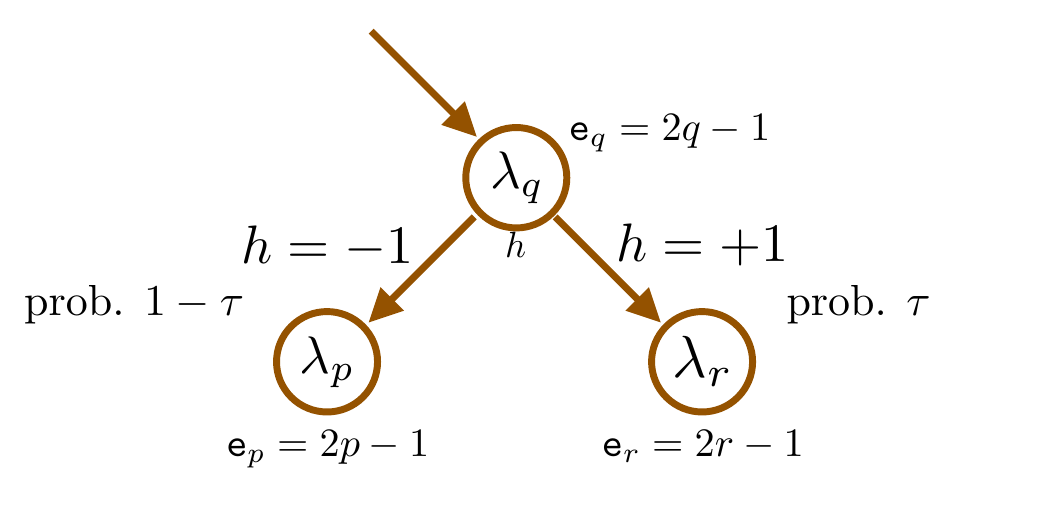}}
    \caption{Main notations used in the proof of Theorem \ref{thm-boosting}, closely following some notations of \citet[Fig. 4]{kmOT}.}
     \label{fig:wha-1}
   \end{figure}

We proceed in two steps, first showing that the loss we care about for fairness \eqref{lossgen} (main file) is upperbounded by the entropy of the $\alpha$-tree $\alphatree$, then developing the boosting result from the minimization of the entropy itself. We thus start with the following Theorem.
\begin{theorem}[Theorem \ref{thm:entropyupper} in Main Paper]\label{thm-alpha-in-alpha-tree}
  Suppose Assumption \ref{assumptionBUNF} holds and the outputs of $\alphatree$ are:
  \begin{eqnarray}
\alphatree(\ve{x}) & \defeq & \normalizedlogit\left(\frac{1+\edge(\meas{M}_{\leaf(\ve{x})}, \postarget)}{2}\right), \forall \ve{x} \in \mathcal{X},\label{alphatreeoutput}
  \end{eqnarray}
where $\leaf(\ve{x})$ is the leaf reached by $\ve{x}$ in $\alphatree$. Then the following bound holds for the risk \eqref{lossgen}:
\begin{eqnarray}
  \totrisk(\posfair; \meas{M}_{\mbox{\tiny{\textup{t}}}}, \postarget) & \leq &  \entropy(\alphatree; \meas{M}_{\mbox{\tiny{\textup{t}}}}, \postarget) \label{cvarcomp-deg3-bound}.
\end{eqnarray}
\end{theorem}
\begin{myproof}
We need a simple Lemma, see \textit{e.g.} \citet{nssBP}.
\begin{lemma}\label{lemGB1-NSS}
$\forall \kappa \in \mathbb{R}, \forall B\geq 0, \forall |z| \leq B$,
\begin{eqnarray}
\log(1+\exp(\kappa z)) & \leq & \log(1+\exp(\kappa B)) - \kappa \cdot \frac{B-z}{2}.\label{boundBZ}
\end{eqnarray}
\end{lemma}
We then note, using $z \defeq
\log\left(\frac{1-\posunfair}{\posunfair}\right)$ (stripping variables
  for readability) and Assumption \ref{assumptionBUNF},
\begin{eqnarray}
  - \log \posfair & = & -\log \left(\frac{\posunfair^\alphatree}{\posunfair^\alphatree+(1-\posunfair)^\alphatree}\right)\nonumber\\
  & = & -\log \left(\frac{1}{1+\left(\frac{1-\posunfair}{\posunfair}\right)^\alphatree}\right)\nonumber\\
                                               & = & \log \left(1+\left(\frac{1-\posunfair}{\posunfair}\right)^\alphatree\right)\nonumber\\
                                               & = & \log\left(1 + \exp\left(\alphatree \log\left(\frac{1-\posunfair}{\posunfair}\right)\right)\right)\nonumber\\ %% CORRECTION
                                               & \leq & \log(1+\exp(\alphatree B)) -\alphatree \cdot \frac{B - \log\left(\frac{1-\posunfair}{\posunfair}\right)}{2}\label{bound1}\\
  & & = \log(1+\exp(\alphatree B)) -\alphatree \cdot \frac{B + \logit(\posunfair)}{2},\label{blog1}
\end{eqnarray}
where in \eqref{bound1} we have used \eqref{boundBZ} with $\kappa \defeq \alphatree$, using Assumption \ref{assumptionBUNF} guaranteeing $| \logit(\posunfair)| \leq B$. We also get, using this time $\kappa \defeq -\alphatree$,
\begin{eqnarray}
 - \log (1-\posfair) & = & \log \left(1 + \exp\left(-\alphatree \log\left(\frac{1-\posunfair}{\posunfair}\right)\right)\right)\nonumber\\
                                & \leq &  \log(1+\exp(-\alphatree B)) +\alphatree \cdot \frac{B + \logit(\posunfair)}{2}\nonumber\\
  & & = \log(1+\exp(\alphatree B)) - \alphatree B +\alphatree \cdot \frac{B + \logit(\posunfair)}{2}\nonumber\\
  & & = \log(1+\exp(\alphatree B)) - \alphatree \cdot \frac{B - \logit(\posunfair)}{2}.\label{blog2}
\end{eqnarray}
Assembling \eqref{blog1} and \eqref{blog2} for an upperbound to $\totrisk(\posfair; \meas{M}_{\mbox{\tiny{\textup{t}}}}, \postarget)$, we get, using the fact that an $\alpha$-tree partitions $\mathcal{X}$ into regions with constant predictions,
\begin{eqnarray}
  \lefteqn{\totrisk(\posfair; \meas{M}, \postarget)}\nonumber\\
  & \defeq & \expect_{\X \sim \meas{M}}  \left[\postarget(\X)\cdot - \log \posfair(\X)+(1-\postarget(\X))\cdot - \log (1-\posfair(\X))\right] \nonumber\\
                                           & \leq & \expect_{\X \sim \meas{M}_{\mbox{\tiny{\textup{t}}}}}  \left[
                                                    \begin{array}{c}
                                                      \postarget(\X)\cdot \left(\log(1+\exp(\alphatree(\X) B)) -\alphatree(\X) \cdot \frac{B + \logit(\posunfair(\X))}{2}\right)\nonumber\\
                                                      +(1-\postarget(\X))\cdot \left(\log(1+\exp(\alphatree(\X) B)) -\alphatree(\X) \cdot \frac{B - \logit(\posunfair(\X))}{2}\right)
                                                    \end{array}\right] \nonumber\\
                                           & & = \expect_{\X \sim \meas{M}_{\mbox{\tiny{\textup{t}}}}}  \left[ \log(1+\exp(\alphatree(\X) B)) - \alphatree(\X) \cdot \left(\begin{array}{c}
                                                                                                                                                  \postarget(\X)\cdot \frac{B + \logit(\posunfair(\X))}{2}\\
                                                                                                                                                  +(1-\postarget(\X))\cdot \frac{B - \logit(\posunfair(\X))}{2}
                                                                                                                                                  \end{array}\right)\right]  \nonumber\\
  & = & \expect_{\leaf \sim \meas{M}_{\leafset(\alphatree)}} \left[ \log(1+\exp(\alphatree(\leaf) B)) - \alphatree(\leaf) \cdot \expect_{\X \sim \meas{M}_\leaf} \left[\left(\begin{array}{c}
                                                                                                                                                  \postarget(\X)\cdot \frac{B + \logit(\posunfair(\X))}{2}\\
                                                                                                                                                  +(1-\postarget(\X))\cdot \frac{B - \logit(\posunfair(\X))}{2}
                                                                                                                                                  \end{array}\right)\right]\right]\nonumber\\
  & = & \expect_{\leaf \sim \meas{M}_{\leafset(\alphatree)}} \left[ \log(1+\exp(\alphatree(\leaf) B)) - \alphatree(\leaf) \cdot \expect_{(\X,\Y) \sim {\measdtarget}_\leaf} \left[\frac{B + \Y \cdot \logit(\posunfair(\X))}{2}\right]\right]\nonumber\\
  & = & \expect_{\leaf \sim \meas{M}_{\leafset(\alphatree)}} \left[ \log(1+\exp(\alphatree(\leaf) B)) - \alphatree(\leaf) \cdot \frac{B + \expect_{(\X,\Y) \sim {\measdtarget}_\leaf} \left[\Y \cdot \logit(\posunfair(\X))\right]}{2}\right]\nonumber\\
  & = & \expect_{\leaf \sim \meas{M}_{\leafset(\alphatree)}} \left[ \log(1+\exp(\alphatree(\leaf) B)) - \alphatree(\leaf) B \cdot \frac{1 + \edge(\meas{M}_\leaf, \postarget)}{2}\right],\label{bENT1}
\end{eqnarray}
where we have used index notation for leaves introduced in the Theorem's statement, used the definition of $\edge(\meas{M}_\leaf, \postarget)$ and let $\alphatree(\leaf)$ denote $\leaf$'s leaf value in $\alphatree$. Looking at \eqref{bENT1}, we see that we can design the leaf values to minimize each contribution to the expectation (noting the convexity of the relevant functions in $\alphatree(\leaf)$), which for any $\leaf \in \leafset(\alphatree)$ we define with a slight abuse of notations as:
\begin{eqnarray}
L(\alphatree(\leaf)) & \defeq & \log(1+\exp(\alphatree(\leaf) B)) - \alphatree(\leaf) B \cdot \frac{1 + \edge(\meas{M}_\leaf, \postarget)}{2}.
\end{eqnarray}
We note
\begin{eqnarray*}
    L'(\alphatree(\leaf)) & = & B \cdot \left(\frac{\exp(\alphatree(\leaf) B)}{1+\exp(\alphatree(\leaf) B)} - \frac{1 + \edge(\meas{M}_\leaf, \postarget)}{2}\right),
\end{eqnarray*}
which zeroes for
\begin{eqnarray*}
\alphatree(\leaf) & = & \frac{1}{B} \cdot \log\left(\frac{1+\edge(\meas{M}_\leaf, \postarget)}{1-\edge(\meas{M}_\leaf, \postarget)}\right) = \normalizedlogit\left(\frac{1+\edge(\meas{M}_{\leaf}, \postarget)}{2}\right), 
\end{eqnarray*}
yielding the bound (we use $\edge(\leaf)$ as a shorthand for $\edge(\meas{M}_\leaf, \postarget)$):
\begin{eqnarray}
  \lefteqn{\totrisk(\posfair; \meas{M}_{\mbox{\tiny{\textup{t}}}}, \postarget)}\nonumber\\
  & \leq & \expect_{\leaf \sim \meas{M}_{\leafset(\alphatree)}} \left[ \log\left(1+\frac{1+\edge(\leaf)}{1-\edge(\leaf)}\right) - \log\left(\frac{1+\edge(\leaf)}{1-\edge(\leaf)}\right) \cdot \frac{1 + \edge(\leaf)}{2}\right]\nonumber\\
                                           & & = \expect_{\leaf \sim \meas{M}_{\leafset(\alphatree)}} \left[ - \log\left(\frac{1-\edge(\leaf)}{2}\right)- \log\left(\frac{1+\edge(\leaf)}{1-\edge(\leaf)}\right) \cdot \frac{1 + \edge(\leaf)}{2}\right]\nonumber\\
  & = &  \expect_{\leaf \sim \meas{M}_{\leafset(\alphatree)}} \left[ - \log\left(\frac{1 - \edge(\leaf)}{2}\right) + \frac{1 + \edge(\leaf)}{2}\cdot \log\left(\frac{1 - \edge(\leaf)}{2}\right) - \frac{1 + \edge(\leaf)}{2} \cdot \log\left(\frac{1+\edge(\leaf)}{2}\right) \right]\nonumber\\
  & = &  \expect_{\leaf \sim \meas{M}_{\leafset(\alphatree)}} \left[ -\frac{1 - \edge(\leaf)}{2}\cdot \log\left(\frac{1 - \edge(\leaf)}{2}\right) - \frac{1 + \edge(\leaf)}{2} \cdot \log\left(\frac{1+\edge(\leaf)}{2}\right) \right]\nonumber\\
  & = &  \entropy(\alphatree; \meas{M}_{\mbox{\tiny{\textup{t}}}}, \postarget),
\end{eqnarray}
which is the statement of Theorem \ref{thm-alpha-in-alpha-tree}.
\end{myproof}\\
Armed with Theorem \ref{thm-alpha-in-alpha-tree}, what we now show is the boosting compliant convergence on the entropy of the $\alpha$-tree. For the informed reader, the proof of our result relies on a generalisation of \citet[Lemma 2]{kmOT}, then branching on the proofs of \citet[Lemma 6, Theorem 9]{kmOT} to complete our result. For this objective, we first introduce notations, summarized in Figure \ref{fig:wha-1}, for the split of a leaf $\leaf_q$ in a subtree with two new leaves $\leaf_p, \leaf_r$. Here, we make use of simplified notation
\begin{eqnarray}
\edge_p & \defeq & \edge(\meas{M}_{\leaf_p}, \postarget),
\end{eqnarray}
and similarly for $\edge_q$ and $\edge_r$. Quantities $p, q, r\in [0,1]$\footnote{Under Assumption \ref{assumptionBUNF}.} are computed from the corresponding $\edge_.$. $\tau$ is the probability, measured from ${\measdtarget}_{\leaf_q}$, that an example has $h(.) = +1$, where $h$ is the split function at $\leaf_q$.
We state and prove our generalisation to \citet[Lemma 2]{kmOT}.    
   \begin{lemma}\label{lemWHA-C}
Assuming notations in Figure \ref{fig:wha-1} for the split $h$ investigated at a leaf $\leaf_q$, and letting $\delta \defeq r - p$, if for some $\upgamma > 0$ the split $h$ $\upgamma$-witnesses the WHA at $\leaf$, then $\tau (1-\tau) \delta \geq \upgamma \cdot q(1-q)$.
\end{lemma}
\begin{myproof}
  Using the definition of the rebalanced distribution, we have:
\begin{eqnarray}
  \lefteqn{\expect_{(\X, \Y) \sim {\rebmeasdtarget}_{\leaf_q}} \left[\Y \normalizedlogit(\posunfair(\X)) h(\X)\right]}\nonumber\\
  & = & \expect_{(\X, \Y) \sim {\measdtarget}_{\leaf_q}} \left[\frac{1 - \edge_q \cdot \Y \cdot \normalizedlogit(\posunfair(\X))}{1 - \edge_q^2} \cdot \Y h(\X) \normalizedlogit(\posunfair(\X))\right] \nonumber\\
  & = & \frac{\expect_{(\X, \Y) \sim {\measdtarget}_{\leaf_q}} \left[\Y h(\X) \normalizedlogit(\posunfair(\X))\right] - \edge_q \cdot \expect_{(\X, \Y) \sim {\measdtarget}_{\leaf_q}} \left[\normalizedlogit^2(\posunfair(\X)) h(\X)\right]}{1 - \edge_q^2}, \label{eqwha1}
\end{eqnarray}
since $y^2 = 1, \forall y \in \mathcal{Y}$. We also have, by definition of the partition induced by $h$ and the definition of $\tau$,
  \begin{eqnarray}
    \tau \edge_r - (1-\tau) \edge_p & = & \tau \cdot \expect_{(\X, \Y) \sim {\measdtarget}_{\leaf_r}} \left[\Y \cdot \normalizedlogit(\posunfair(\X))\right] - (1-\tau) \cdot \expect_{(\X, \Y) \sim {\measdtarget}_{\leaf_p}} \left[\Y \cdot \normalizedlogit(\posunfair(\X))\right]\nonumber\\
    & = & \expect_{(\X, \Y) \sim {\measdtarget}_{\leaf_q}} \left[\Y h(\X) \normalizedlogit(\posunfair(\X))\right].\label{eqwha2}
  \end{eqnarray}
  We can thus write:
  \begin{eqnarray}
    \lefteqn{\expect_{(\X, \Y) \sim {\rebmeasdtarget}_{\leaf_q}} \left[\Y \normalizedlogit(\posunfair(\X)) h(\X)\right]}\nonumber\\
    & = & \frac{\tau \edge_r - (1-\tau) \edge_p - \edge_q \cdot \expect_{(\X, \Y) \sim {\measdtarget}_{\leaf_q}} \left[\normalizedlogit^2(\posunfair(\X)) h(\X)\right]}{1 - \edge_q^2}\label{eqwha3}\\
    & = & \frac{2 \tau \edge_r - \edge_q\cdot \left(1+\expect_{(\X, \Y) \sim {\measdtarget}_{\leaf_q}} \left[\normalizedlogit^2(\posunfair(\X)) h(\X)\right]\right)}{1 - \edge_q^2}\label{eqwha4}\\
    & = & \frac{2 \tau \edge_r - 2 \tau \edge_q}{1 - \edge_q^2} - \edge_q\cdot \frac{\left(1 - 2\tau +\expect_{(\X, \Y) \sim {\measdtarget}_{\leaf_q}} \left[\normalizedlogit^2(\posunfair(\X)) h(\X)\right]\right)}{1 - \edge_q^2}\label{eqwha5}\\
    & = & \frac{2 \tau \edge_r - 2 \tau \edge_q}{1 - \edge_q^2} + \edge_q\cdot \frac{\left(2\tau -1 - \expect_{(\X, \Y) \sim {\measdtarget}_{\leaf_q}} \left[\normalizedlogit^2(\posunfair(\X)) h(\X)\right]\right)}{1 - \edge_q^2}\label{eqwha6}\\
    & = & \frac{2 \tau \edge_r - 2 \tau \edge_q}{1 - \edge_q^2} + \frac{\edge_q\cdot \expect_{(\X, \Y) \sim {\measdtarget}_{\leaf_q}} \left[(1-\normalizedlogit^2(\posunfair(\X)))\cdot h(\X)\right]}{1 - \edge_q^2}\label{eqwha7}.
  \end{eqnarray}
  Here, \eqref{eqwha3} follows from \eqref{eqwha1} and \eqref{eqwha2}, \eqref{eqwha4} uses the fact that $\edge_q = (1-\tau)\edge_p + \tau \edge_r$, \eqref{eqwha5} and \eqref{eqwha6} are convenient reformulations after adding $2 \tau \edge_q-2 \tau \edge_q$ and \eqref{eqwha7} follows from $\expect_{(\X, \Y) \sim {\measdtarget}_{\leaf_q}} \left[h(\X)\right] = 2\tau - 1$ by definition of $\tau$ and $h\in \{-1,1\}$.  Let
  \begin{eqnarray}
    \Delta(h) & \defeq & \edge_q\cdot \expect_{(\X, \Y) \sim {\measdtarget}_{\leaf_q}} \left[(1-\normalizedlogit^2(\posunfair(\X)))\cdot h(\X)\right].
  \end{eqnarray}
We have $p = (1 + \edge_p)/2$ (and similarly for $q = (1 + \edge_q)/2$ and $r = (1 + \edge_r)/2$), so we reformulate \eqref{eqwha6} as:
\begin{eqnarray}
  \expect_{(\X, \Y) \sim {\rebmeasdtarget}_{\leaf_q}} \left[\Y \normalizedlogit(\posunfair(\X)) h(\X)\right]& = & \frac{2\tau(2r-2q)}{4q (1-q)} + \frac{\Delta(h)}{4q (1-q)} \nonumber\\
                                                                                                            & = & \frac{\tau(r-q)}{q (1-q)} + \frac{\Delta(h)}{4q (1-q)} \nonumber\\
                                                                                                            & = & \frac{\tau(1-\tau) \delta}{q (1-q)} + \frac{\Delta(h)}{4q (1-q)}, \label{lastEQ}
\end{eqnarray}
where the last identity comes from the fact that $r = q + (1-\tau) \delta$. We now have two cases depending on what removing the absolute value in the WHA leads to:\\
\noindent \textbf{Case 1} (i) is $\expect_{(\X, \Y) \sim {\rebmeasdtarget}_{\leaf_q}} \left[\Y \normalizedlogit(\posunfair(\X)) h(\X)\right] \geq \upgamma$. We get from \eqref{lastEQ}:
\begin{eqnarray}
  \tau(1-\tau) \delta & \geq & \upgamma \cdot q(1-q) - \frac{\Delta(h)}{4}, \label{propH1}
\end{eqnarray}
and since (ii) brings $\Delta(h) \leq 0$, we obtain $\tau(1-\tau) \delta \geq \upgamma \cdot q(1-q) $, as claimed.\\
\noindent \textbf{Case 2} (i) is $\expect_{(\X, \Y) \sim {\rebmeasdtarget}_{\leaf_q}} \left[\Y \normalizedlogit(\posunfair(\X)) h(\X)\right] \leq -\upgamma$. Since $\mathcal{H}$ is closed by negation we replace $h$ by $h' \defeq -h$, which satisfies $\expect_{(\X, \Y) \sim {\rebmeasdtarget}_{\leaf_q}} \left[\Y \normalizedlogit(\posunfair(\X)) h'(\X)\right] = - \expect_{(\X, \Y) \sim {\rebmeasdtarget}_{\leaf_q}} \left[\Y \normalizedlogit(\posunfair(\X)) h(\X)\right]$. The change switches the sign of $\delta$ by its definition and also $\Delta(h') = -\Delta(h)$ so \eqref{propH1} becomes $-\tau(1-\tau) \delta \leq -\upgamma \cdot q(1-q) + \Delta(h')/4$, \textit{i.e.}
\begin{eqnarray}
  \tau(1-\tau) \delta & \geq & \upgamma \cdot q(1-q) - \frac{\Delta(h')}{4}, \label{propH2}
\end{eqnarray}
which brings us back to Case 1 with the switch $h \leftrightarrow h'$ as $h'$ satisfies $\expect_{(\X, \Y) \sim {\rebmeasdtarget}_{\leaf_q}} \left[\Y \normalizedlogit(\posunfair(\X)) h'(\X)\right] \geq \upgamma$. This ends the proof of Lemma \ref{lemWHA-C}.
\end{myproof}\\
Branching Lemma \ref{lemWHA-C} to the proof of Theorem \ref{thm-boosting} via the results of \cite{kmOT} is simple as all major parameters $p, q, r, \delta, \tau$ are either the same or satisfy the same key relationships (linked to the linearity of the expectation). This is why, if we compute the decrease $\entropy(\alphatree; \meas{M}_{\mbox{\tiny{\textup{t}}}}, \postarget) - \entropy(\alphatree(\leaf, h); \meas{M}_{\mbox{\tiny{\textup{t}}}}, \postarget)$, $\alphatree(\leaf, h)$ being the $\alpha$-tree $\alphatree$ with the split in Figure \ref{fig:wha-1} performed with $h$ at $\leaf$, then we immediately get
\begin{eqnarray}
\entropy(\alphatree; \meas{M}_{\mbox{\tiny{\textup{t}}}}, \postarget) - \entropy(\alphatree(\leaf, h); \meas{M}_{\mbox{\tiny{\textup{t}}}}, \postarget) & \geq & \upgamma^2 q(1-q), \label{eqDEC1}
\end{eqnarray}
which comes from \citet[Lemma 6]{kmOT}, and \eqref{eqDEC1} can be directly used in the proof of \citet[Theorem 9]{kmOT} -- which unravels the local decrease of $\entropy(.; \meas{M}_{\mbox{\tiny{\textup{t}}}}, \postarget)$ to get to the global decrease of the criterion for the whole of $\alphatree$'s induction --, and to get $\entropy(\alphatree; \meas{M}_{\mbox{\tiny{\textup{t}}}}, \postarget) \leq \varepsilon$, it is sufficient that
\begin{eqnarray}
|\leafset(g)| & \geq & \left(\frac{1}{\varepsilon}\right)^{\frac{c \log \left(\frac{1}{\varepsilon}\right)}{\upgamma^2}},\label{eqLAST-th}
\end{eqnarray}
as claimed, for $c>0$ a constant. This ends the proof of Theorem \ref{thm-boosting}.
\begin{remark}
  Lemma \ref{lemGB1-NSS} reveals an interesting property: instead of requesting $\faircorr_{S', \leaf} (h) \leq 0$ in split-fair-compliance, suppose we strengthen the assumption, requesting for some $\beta > 0$ that
  \begin{eqnarray}
\faircorr_{S', \leaf} (h) & \leq & - \upbeta \cdot (1 - \edge_q^2), \label{betaPROP}
  \end{eqnarray}
  then the "advantage" $\upgamma$ becomes an advantage $\upgamma + \upbeta$ in \eqref{eqLAST-th}. Since we have $\faircorr_{S', \leaf} (h) \defeq  \edge_q\cdot \expect_{(\X, \Y) \sim \meas{D}_{S', \leaf_q}} \left[(1-\normalizedlogit^2(\posunfair(\X)))\cdot h(\X)\right]$, constraint \eqref{betaPROP} quickly vanishes as $|\edge_q| \rightarrow 1$, \textit{i.e.} as the black-box gest very good --or-- very bad (in this last case, we remark that $1-\posunfair$ becomes very good, so this is not a surprise). For example, if $\edge_q \geq 1 - \varepsilon'$ for small $\varepsilon'$, then we just need
  \begin{eqnarray}
 \expect_{(\X, \Y) \sim \meas{D}_{S', \leaf_q}} \left[(1-\normalizedlogit^2(\posunfair(\X)))\cdot h(\X)\right] & \leq & -\varepsilon' \upbeta \cdot \frac{2-\varepsilon'}{1-\varepsilon'}.
    \end{eqnarray}
  \end{remark}

\section{Proof of Theorem \ref{thm-alternative-alpha-in-alpha-tree}}\label{proof-thm-alternative-alpha-in-alpha-tree}

The proof is obtained via a generalisation of Lemma \ref{lemGB1-NSS}.
\begin{lemma}\label{lemGB2}
  Fix any $B>0$. For any $\alpha \in \mathbb{R}$, any $\theta, z \in [-B,B]$, if we let
  \begin{eqnarray}
  \vartheta(z) & \defeq & (z-\theta) \cdot \left\{ \begin{array}{ccl}
                                                     \frac{1}{B+\theta} & \mbox{ if } & z < \theta,\\
                                                     0 & \mbox{ if } & z  = \theta,\\
                             \frac{1}{B-\theta}& \mbox{ if } & z > \theta.
                           \end{array}\right.,
\end{eqnarray}
  then we have
  \begin{eqnarray*}
\log(1+\exp(\alpha z)) & \leq & \log\left(\frac{1+\exp(B \alpha)}{1+\exp(\theta \alpha)}\right) \cdot |\vartheta(z)| - B\alpha \max\{0, -\vartheta(z)\} + \log(1+\exp(\theta \alpha)).
  \end{eqnarray*}
\end{lemma}
\textbf{Remark}: Lemma \ref{lemGB1-NSS} is obtained for the choices $\theta = \pm B$.\\
\begin{myproof}
  We fix any $\theta' \in [-1,1]$ and let
  \begin{eqnarray}
    \mathrm{l} & \defeq & (-1, \log(1+\exp(-\alpha))),\\
    \mathrm{c} & \defeq & (\theta', \log(1+\exp(\alpha\theta'))),\\
\mathrm{r} & \defeq & (1, \log(1+\exp(\alpha))).
  \end{eqnarray}
  The equation of the line passing through $\mathrm{l}, \mathrm{c}$ is
\begin{align}
  &f_{\mathrm{l}}(z) \nonumber \\
  & =  \frac{\log\left(\frac{1+\exp(\theta' \alpha)}{1+\exp(-\alpha)}\right)}{1+\theta'} \cdot z + \frac{\log\left(\frac{1+\exp(\theta' \alpha)}{1+\exp(-\alpha)}\right)}{1+\theta'} + \log(1+\exp(-\alpha))\\
  & =  -\frac{\log\left(\frac{1+\exp(\alpha)}{1+\exp(\theta' \alpha)}\right)}{1+\theta'} \cdot z + \frac{\alpha z}{1+\theta'} -\frac{\log\left(\frac{1+\exp(\alpha)}{1+\exp(\theta' \alpha)}\right)}{1+\theta'} + \frac{\alpha}{1+\theta'}+ \log(1+\exp(-\alpha))\\
   & = \frac{\log\left(\frac{1+\exp(\alpha)}{1+\exp(\theta' \alpha)}\right)}{1+\theta'} \cdot (\theta' - z) + \frac{\alpha (z-\theta')}{1+\theta'} - \log\left(\frac{1+\exp(\alpha)}{1+\exp(\theta' \alpha)}\right) + \log(1+\exp(\alpha))\\
   & = \frac{\log\left(\frac{1+\exp(\alpha)}{1+\exp(\theta' \alpha)}\right)}{1+\theta'} \cdot (\theta' - z) + \frac{\alpha (z-\theta')}{1+\theta'} + \log(1+\exp(\theta' \alpha))
\end{align}
and the equation of the line passing through $\mathrm{c},\mathrm{r}$ is
\begin{eqnarray}
  f_{\mathrm{r}}(z) & = & \frac{\log\left(\frac{1+\exp(\alpha)}{1+\exp(\theta' \alpha)}\right)}{1-\theta'} \cdot z - \frac{\log\left(\frac{1+\exp(\alpha)}{1+\exp(\theta' \alpha)}\right)}{1-\theta'}  + \log(1+\exp(\alpha))\\
  & = & \frac{\log\left(\frac{1+\exp(\alpha)}{1+\exp(\theta' \alpha)}\right)}{1-\theta'} \cdot (z-\theta') - \log\left(\frac{1+\exp(\alpha)}{1+\exp(\theta' \alpha)}\right)  + \log(1+\exp(\alpha))\\
  & = & \frac{\log\left(\frac{1+\exp(\alpha)}{1+\exp(\theta' \alpha)}\right)}{1-\theta'} \cdot (z-\theta') + \log(1+\exp(\theta' \alpha)).
\end{eqnarray}
For any $z \in [-1,1]$, define $\vartheta'(z) \in [-1,1]$ to be:
\begin{eqnarray}
  \vartheta'(z) & \defeq & (z-\theta') \cdot \left\{ \begin{array}{ccl}
                                                       \frac{1}{1+\theta'} & \mbox{ if } & z < \theta',\\
                                                       0 & \mbox{ if } & z = \theta',\\
                                 \frac{1}{1-\theta'}& \mbox{ if } & z > \theta'.
                                 \end{array}\right.
  \end{eqnarray}
Function $z\mapsto \log(1+\exp(\alpha z))$ being convex, we thus get the secant upperbound:
\begin{align}
&\log(1+\exp(\alpha z)) \nonumber \\
& \leq \log\left(\frac{1+\exp(\alpha)}{1+\exp(\theta' \alpha)}\right) \cdot |\vartheta'(z)| + \alpha \min\{0, \vartheta'(z)\} + \log(1+\exp(\theta' \alpha)),
\end{align}
and this holds for $z \in [-1,1]$. If instead $z\in [-B,B]$, then letting $\theta \defeq B \theta' \in [-B, B]$, we note:
\begin{align}
  &\log(1+\exp(\alpha z)) \nonumber \\
  & =  \log(1+\exp(\alpha B \cdot (z/B))) \nonumber\\
  & \leq  \log\left(\frac{1+\exp(\alpha B)}{1+\exp(\theta' \alpha B)}\right) \cdot |\vartheta'(z/B)| + \alpha B \min\{0, \vartheta'(z/B)\} + \log(1+\exp(\theta' \alpha B)),
\end{align}
where this time,
\begin{eqnarray}
  \vartheta'\left(\frac{z}{B}\right) & \defeq & \left(\frac{z}{B}-\theta'\right) \cdot \left\{ \begin{array}{ccl}
                                                                                                 \frac{1}{1+\theta'} & \mbox{ if } & z < B\theta',\\
                                                                                                 0 & \mbox{ if } & z = B\theta',\\
                                 \frac{1}{1-\theta'}& \mbox{ if } & z > B\theta'.
                                                     \end{array}\right.\nonumber\\
  & = &  (z-\theta) \cdot \left\{ \begin{array}{ccl}
                                 \frac{1}{B+\theta} & \mbox{ if } & z < \theta,\\
                                                                                                 0 & \mbox{ if } & z = \theta,\\
                                 \frac{1}{B-\theta}& \mbox{ if } & z > \theta.
                                                     \end{array}\right. \defeq \vartheta(z).
\end{eqnarray}
We thus get
\begin{align}
&\log(1+\exp(\alpha z)) \nonumber \\
& \leq \log\left(\frac{1+\exp(B \alpha)}{1+\exp(\theta \alpha)}\right) \cdot |\vartheta(z)| + B\alpha \min\{0, \vartheta(z)\} + \log(1+\exp(\theta \alpha)),
\end{align}
and since $\min\{0,z\} = -\max\{0,-z\}$, we get the statement of the Lemma.
\end{myproof}\\
We use Lemma \ref{lemGB2} with $\theta = 0$, which yields $\vartheta(z) = z/B$; using notations from the proof of Theorem \ref{thm-alpha-in-alpha-tree}, we thus get (using the same notations as in the proof of Theorem \ref{thm-boosting}),
\begin{align}
  &- \log \posfair \nonumber \\
  & =  \log\left(1 + \exp\left(\alphatree \cdot -\logit(\posunfair)\right)\right)\nonumber\\ %% CORRECTION
                             & \leq \frac{1}{B}\cdot \log\left(\frac{1+\exp(B\alphatree)}{2}\right) \cdot |\logit(\posunfair(\X))| + \alphatree \min\{0, -\logit(\posunfair(\X))\}+ \log(2)\nonumber\\
  & =   \frac{1}{B}\cdot \log\left(\frac{1+\exp(B\alphatree)}{2}\right) \cdot |\logit(\posunfair(\X))| - \alphatree \max\{0, \logit(\posunfair(\X))\}+ \log(2)\label{blogb1}\\
  \nonumber \\
 &- \log (1-\posfair) \nonumber \\
 & =  \log \left(1 + \exp\left(\alphatree \cdot \logit(\posunfair(\X))\right)\right)\nonumber\\
                             & \leq   \frac{1}{B}\cdot \log\left(\frac{1+\exp(B\alphatree)}{2}\right) \cdot |\logit(\posunfair(\X))| + \alphatree \min\{0, \logit(\posunfair(\X))\} + \log(2)\nonumber\\
  & =  \frac{1}{B}\cdot \log\left(\frac{1+\exp(B\alphatree)}{2}\right) \cdot |\logit(\posunfair(\X))| - \alphatree \max\{0, -\logit(\posunfair(\X))\} + \log(2).\label{blogb2}
\end{align}
We get that the inequality in \eqref{bENT1} now reads (for \textit{any} values $\{\alphatree(\leaf), \leaf \in \leafset(\alphatree)\}$) $\totrisk(\posfair; \meas{M}_{\mbox{\tiny{\textup{t}}}}, \postarget) = \expect_{\leaf \sim \meas{M}_{\leafset(\alphatree)}} \left[ J(\leaf)\right]$ with $J(\leaf)$ satisfying:
\begin{align}
&J(\leaf) \nonumber \\
& \leq \expect_{\X \sim \meas{M}_{\leaf}} \left[\begin{array}{l}
        \postarget(\X)\cdot\left(\frac{1}{B}\cdot \log\left(\frac{1+\exp(B\alphatree(\leaf))}{2}\right) \cdot |\logit(\posunfair(\X))| - \alphatree(\leaf) \max\{0, \logit(\posunfair(\X))\}+ \log(2)\right)\\
        +(1-\postarget(\X))\cdot \left(\frac{1}{B}\cdot \log\left(\frac{1+\exp(B\alphatree(\leaf))}{2}\right) \cdot |\logit(\posunfair(\X))| - \alphatree(\leaf) \max\{0, -\logit(\posunfair(\X))\} + \log(2)\right)
                                                           \end{array}\right]\nonumber\\
  & =  \log(2) - B\alphatree(\leaf)\cdot \edge^+(\meas{M}_{\leaf}, \postarget) \nonumber \\ & \quad \quad +\log\left(\frac{1+\exp(B\alphatree(\leaf))}{2}\right) \cdot (\edge^+(\meas{M}_{\leaf}, \postarget) + \edge^-(\meas{M}_{\leaf}, \postarget)),
  \end{align}
  and the bound takes its minimum on $\alphatree(\leaf)$ for
  \begin{eqnarray}
\alphatree(\leaf) & = & \frac{1}{B} \cdot \log\left(\frac{\edge^+(\meas{M}_{\leaf}, \postarget)}{\edge^-(\meas{M}_{\leaf}, \postarget)}\right) =  \normalizedlogit\left(\frac{\edge^+(\meas{M}_{\leaf}, \postarget)}{\edge^+(\meas{M}_{\leaf}, \postarget) + \edge^-(\meas{M}_{\leaf}, \postarget)}\right),
    \end{eqnarray}
yielding (using notations from Theorem \ref{thm-alternative-alpha-in-alpha-tree}),
    \begin{eqnarray}
J(\leaf) & \leq & \log(2)\cdot\left(1- \edge^-_{\leaf} -\edge^+_{\leaf}\right) - \edge^+_{\leaf}\cdot \log\left(\frac{\edge^+_{\leaf}}{\edge^-_{\leaf}}\right) + \log\left(\frac{\edge^-_{\leaf} + \edge^+_{\leaf}}{\edge^-_{\leaf}}\right) \cdot (\edge^-_{\leaf} +\edge^+_{\leaf})\nonumber\\
                            & & =  \log(2) \cdot \left(1+(\edge^-_{\leaf} + \edge^+_{\leaf})\cdot \left(H_2 \left(\frac{\edge^+_{\leaf}}{\edge^+_{\leaf} +\edge^-_{\leaf}}\right) - 1\right)\right),
    \end{eqnarray}
    and brings the statement of Theorem \ref{thm-alternative-alpha-in-alpha-tree} after plugging the bound in the expectation.

\section{Proof of Lemma \ref{lem-entropies}}\label{proof-lem-entropies}
We note that $H_2(1/2) = 1$, so we can reformulate:
\begin{eqnarray}
  \frac{\entropy_2(\leaf; \meas{M}, \postarget)}{\log 2} & = & (1- (\edge^+_{\leaf} + \edge^-_{\leaf}))\cdot H_2\left(\frac{1}{2}\right) +(\edge^+_{\leaf} + \edge^-_{\leaf})\cdot H_2 \left(\frac{\edge^+_{\leaf}}{\edge^+_{\leaf}+\edge^-_{\leaf}}\right),\label{propH21}
\end{eqnarray}
and we also have $\edge^+_{\leaf} \leq 0, \edge^-_{\leaf} \geq 0, \edge^+_{\leaf}+\edge^-_{\leaf} \leq 1$, plus
\begin{eqnarray}
  (1- (\edge^+_{\leaf} + \edge^-_{\leaf}))\cdot\left(\frac{1}{2}\right) +(\edge^+_{\leaf} + \edge^-_{\leaf})\cdot \left(\frac{\edge^+_{\leaf}}{\edge^+_{\leaf}+\edge^-_{\leaf}}\right) = \frac{1 + \edge^+_{\leaf} - \edge^-_{\leaf} }{2} = \frac{1+\edge(\meas{M}_\leaf, \postarget)}{2},
\end{eqnarray}
as indeed $\edge(\meas{M}_\leaf, \postarget) = \edge^+_{\leaf} - \edge^-_{\leaf}$ from its definition. Thus, by Jensen's inequality, since $H$ is concave,
\begin{eqnarray*}
  \lefteqn{\log(2) \cdot \left(1+(\edge^+_{\leaf}+\edge^-_{\leaf})\cdot \left(H_2 \left(\frac{\edge^+_{\leaf}}{\edge^+_{\leaf}+\edge^-_{\leaf}}\right) - 1\right)\right)}\nonumber\\
  & = & \log(2) \cdot \left((1- (\edge^+_{\leaf}+\edge^-_{\leaf}))\cdot H_2\left(\frac{1}{2}\right) +(\edge^-_{\rho, \leaf} + \edge^+_{\rho, \leaf})\cdot H_2 \left(\frac{\edge^+_{\leaf}}{\edge^+_{\leaf}+\edge^-_{\leaf}}\right)\right)\\
  & \leq & \log(2) \cdot H_2\left( (1- (\edge^+_{\leaf}+\edge^-_{\leaf})) \cdot \frac{1}{2} + (\edge^+_{\leaf}+\edge^-_{\leaf})\cdot \frac{\edge^+_{\leaf}}{\edge^+_{\leaf}+\edge^-_{\leaf}}\right)\\
  & & = \log(2) \cdot H_2\left(\frac{1+\edge(\meas{M}_\leaf, \postarget)}{2}\right)\\
  & = & H\left(\frac{1+\edge(\meas{M}_\leaf, \postarget)}{2}\right),
\end{eqnarray*}
which, after plugging in expectations and simplifying, yields the statement of Lemma \ref{lem-entropies}.

\section{Proof of Theorem \ref{thm-eoo}}\label{proof-thm-eoo}

We remind that we craft product measures using a mixture and a posterior that shall be implicit from context: we thus note that the KL divergence
\begin{eqnarray}
  \kl(\postarget, \posfair; \meas{M}_{\mbox{\tiny{\textup{t}}}}) & \defeq & \expect_{(\X, \Y)\sim \measdtarget} \left[\log \left(\frac{\dmeasdtarget((\X,\Y))}{\dmeasdfair((\X,\Y))}\right) \right]\\
  & = & -\expect_{\X \sim \meas{M}_{\mbox{\tiny{\textup{t}}}}} \left[\postarget(\X)\cdot \log \left(\frac{\posfair(\X)}{\postarget(\X)}\right) + (1-\postarget(\X))\cdot \log \left(\frac{1-\posfair(\X)}{1-\postarget(\X)}\right) \right]\\
                                      & = &  \totrisk(\posfair; \meas{M}_{\mbox{\tiny{\textup{t}}}}, \postarget) - \expect_{\X \sim \meas{M}_{\mbox{\tiny{\textup{t}}}}}  \left[ H(\postarget(\X))\right],\label{defKL2}
\end{eqnarray}
where $\measdtarget$ (resp. $\measdfair$) is obtained from couple $(\meas{M}_{\mbox{\tiny{\textup{t}}}}, \postarget)$ (resp. $(\meas{M}_{\mbox{\tiny{\textup{t}}}}, \posfair)$). Denote
\begin{eqnarray}
s^\circ & \defeq & \arg\min_s \pr_{\X \sim \meas{P}_s} \left[\predfair(\X) = 1 \right],
\end{eqnarray}
where $\predfair$ is the $+1/-1$ prediction obtained from the posterior $\posfair$ using \textit{e.g.} the sign of its logit. We define the total variation divergence:
\begin{eqnarray}
\tv(\postarget, \posfair; \meas{M}_{\mbox{\tiny{\textup{t}}}}) & \defeq & \int_{\mathcal{X}\times \mathcal{Y}} |\dmeasdtarget((\X,\Y))-\dmeasdfair((\X,\Y))|,
\end{eqnarray}
which, because of the definition of the product measures, is also equal to:
\begin{eqnarray}
  \tv(\postarget, \posfair; \meas{M}_{\mbox{\tiny{\textup{t}}}}) & = & \int_{\mathcal{X}} | \postarget(\X) \dmeas{M}_{\mbox{\tiny{\textup{t}}}}(\X) - \posfair(\X) \dmeas{M}_{\mbox{\tiny{\textup{t}}}}(\X)| \\
  & & + \int_{\mathcal{X}} | (1-\postarget(\X)) \dmeas{M}_{\mbox{\tiny{\textup{t}}}}(\X) - (1-\posfair(\X)) \dmeas{M}_{\mbox{\tiny{\textup{t}}}}(\X)|\\
  & = & 2 \int_{\mathcal{X}} | \postarget(\X) - \posfair(\X) | \dmeas{M}_{\mbox{\tiny{\textup{t}}}}(\X).\label{defTV2}
\end{eqnarray}
We have Pinsker's inequality, $\tv(\postarget, \posfair; \meas{M}_{\mbox{\tiny{\textup{t}}}}) \leq \sqrt{2 \kl(\postarget, \posfair; \meas{M}_{\mbox{\tiny{\textup{t}}}})}$ (see \textit{e.g.} \cite{vehRD}), so if we run \topdown~until
\begin{eqnarray}
  \totrisk(\posfair; \meas{M}_{\mbox{\tiny{\textup{t}}}}, \postarget) & \leq & \frac{\tau^2}{2} + \expect_{\X \sim \meas{M}_{\mbox{\tiny{\textup{t}}}}}  \left[ H(\postarget(\X))\right],\label{condTOPDOWNa}
\end{eqnarray}
then because of \eqref{defKL2} and \eqref{defTV2},
\begin{eqnarray}
 \int_{\mathcal{X}} | \postarget(\X) - \posfair(\X) | \dmeas{M}_{\mbox{\tiny{\textup{t}}}}(\X) & \leq & \tau.\label{firstapprox}
\end{eqnarray}
Denote subgroups $s^\star \defeq \arg\max_s \pr_{\X \sim \meas{P}_s} \left[\predfair(\X) = 1 \right]$ and $s^\circ \defeq \arg\min_s \pr_{\X \sim \meas{P}_s} \left[\predfair(\X) = 1 \right]$. We pick
\begin{eqnarray}
  \meas{M}_{\mbox{\tiny{\textup{t}}}} & \leftarrow & \meas{P}_{s^\circ}
\end{eqnarray}
for $\topdown$ and the $(p, \delta)$-push up posterior $\postarget$, with
\begin{eqnarray}
p & \defeq & \pr_{\X \sim \meas{P}_{s^\star}} \left[\predfair(\X) = 1 \right] + \frac{\delta}{2},\label{defP}
\end{eqnarray}
assuming the RHS is $\leq 1$.

Denote $\mathcal{X}_{p, s^\circ}$ the subset of the support of $\meas{P}_{s^\circ}$ such that $\postarget(\X) \geq (1/2) + \delta$. Notice that by definition,
\begin{eqnarray}
\int_{\mathcal{X}_{p, s^\circ}} \dmeas{P}_{s^\circ}(\X) & = & p.\label{defXCIRC}
  \end{eqnarray}
We have two possible outcomes for $\posfair$ of relevance on $\mathcal{X}_{p, s^\circ}$: (i) $\posfair(\X) \leq 1/2$ and (ii) $\posfair(\X) > 1/2$. Notice that in this latter case, we are guaranteed that $\predfair(\X) = 1$, which counts towards bringing closer $\pr_{\X \sim \meas{P}_{s^\circ}} \left[\predfair(\X) = 1 \right]$ to $\pr_{\X \sim \meas{P}_{s^\star}} \left[\predfair(\X) = 1 \right]$, so we have to make sure that (i) occurs with sufficiently small probability, and this is achieved via guarantee \eqref{firstapprox}.

If the total weight on $\mathcal{X}_{p, s^\circ}$ of the event (i) $\posfair(\X) \leq 1/2$ is more than $\delta$, then
\begin{eqnarray}
  \int_{\mathcal{X}} | \postarget(\X) - \posfair(\X) | \dmeas{P}_{s^\circ}(\X) & \geq & \int_{\mathcal{X}_{p, s^\circ}} | \postarget(\X) - \posfair(\X) | \dmeas{P}_{s^\circ}(\X)\nonumber\\
  & \geq & \left|\frac{1}{2}+\delta - \frac{1}{2}\right| \cdot \int_{\mathcal{X}_{p, s^\circ}} \iver{\posfair(\X) \leq 1/2} \dmeas{P}_{s^\circ}(\X) \nonumber\\
  & > & \left|\frac{1}{2}+\delta - \frac{1}{2}\right| \cdot \delta \nonumber\\
  & & = \delta^2. \label{secondapprox}
\end{eqnarray}
If we have the relationship $\delta = \sqrt{\tau}$, then we get a contradiction with \eqref{firstapprox}. In conclusion, if \eqref{condTOPDOWNa} holds, then
\begin{eqnarray}
\int_{\mathcal{X}_{p, s^\circ}} \iver{\posfair(\X) \leq 1/2} \dmeas{P}_{s^\circ}(\X) & \leq & \delta.\label{boundBAD}
  \end{eqnarray}

In summary, for any $\tau > 0$, if we run \topdown~with the choices $\meas{M}_{\mbox{\tiny{\textup{t}}}} \leftarrow \meas{P}_{s^\circ}$ (which corresponds to the "worst treated" subgroup with respect to EOO) and craft the $(p, \delta)$-push up posterior $\postarget$ with $p$ as in \eqref{defP}, then
\begin{eqnarray}
  \pr_{\X \sim \meas{P}_{s^\circ}} \left[\predfair(\X) = 1 \right] & \geq & \int_{\mathcal{X}_{p, s^\circ}} \iver{\posfair(\X) > 1/2} \dmeas{P}_{s^\circ}(\X)\\
   & & = \int_{\mathcal{X}_{p, s^\circ}} (1-\iver{\posfair(\X) \leq 1/2}) \dmeas{P}_{s^\circ}(\X)\\
   & = & \int_{\mathcal{X}_{p, s^\circ}} \dmeas{P}_{s^\circ}(\X) - \int_{\mathcal{X}_{p, s^\circ}} \iver{\posfair(\X) \leq 1/2} \dmeas{P}_{s^\circ}(\X)\\
  & \geq & p - \delta \label{useb}\\
  &  & = \pr_{\X \sim \meas{P}_{s^\star}} \left[\predfair(\X) = 1 \right] - \frac{\delta}{2},
  \end{eqnarray}
  where \eqref{useb} makes use of \eqref{defXCIRC} and \eqref{boundBAD}. Fixing $\delta \defeq 2 \epsilon$, $\epsilon$ being used in \eqref{eqEOO} (main file), we obtain
  \begin{eqnarray}
\pr_{\X \sim \meas{P}_{s^\star}} \left[\predfair(\X) = 1 \right]  - \pr_{\X \sim \meas{P}_{s^\circ}} \left[\predfair(\X) = 1 \right] & \leq & \epsilon,\label{finEQ}
  \end{eqnarray}
and via relationship $\delta = \sqrt{\tau}$, we check that \eqref{condTOPDOWNa} becomes the following function of $\epsilon$:
  \begin{eqnarray}
  \totrisk(\posfair; \meas{M}_{\mbox{\tiny{\textup{t}}}}, \postarget) & \leq & 8\epsilon^4 + \expect_{\X \sim \meas{M}_{\mbox{\tiny{\textup{t}}}}}  \left[ H(\postarget(\X))\right],\label{condTOPDOWN}
\end{eqnarray}
and we get the statement of the Theorem for the choice \eqref{defP}, which corresponds to $K=2$ and reads
\begin{eqnarray}
p & \defeq & \pr_{\X \sim \meas{P}_{s^\star}} \left[\predfair(\X) = 1 \right] + \epsilon.\label{defPEPS}
\end{eqnarray}
If the RHS in \eqref{defPEPS} is not $\leq 1$, we can opt for an alternative with one more free variable, $K\geq 1$,
\begin{eqnarray}
p & \defeq & \pr_{\X \sim \meas{P}_{s^\star}} \left[\predfair(\X) = 1 \right] + \frac{\delta}{K},\label{defP2}
\end{eqnarray}
where $K$ is large enough for the constraint to hold. In this case, to keep \eqref{finEQ} we must have $\delta(K-1)/K = \epsilon$, which elicitates
\begin{eqnarray}
\delta & = & \frac{K\epsilon}{K-1}
\end{eqnarray}
instead of $\delta \defeq 2 \epsilon$, bringing
\begin{eqnarray}
p & \defeq & \pr_{\X \sim \meas{P}_{s^\star}} \left[\predfair(\X) = 1 \right] + \frac{\epsilon}{K-1},\label{defP3}
\end{eqnarray}
and a desired approximation guarantee for \topdown~of:
\begin{eqnarray}
  \totrisk(\posfair; \meas{M}_{\mbox{\tiny{\textup{t}}}}, \postarget) & \leq & \frac{K^4}{2(K-1)^4} \cdot \epsilon^4 + \expect_{\X \sim \meas{M}_{\mbox{\tiny{\textup{t}}}}}  \left[ H(\postarget(\X))\right].\label{condTOPDOWN2}
\end{eqnarray}
Since $K > 1$, $K^4/(K-1)^4\geq 1$, so we are guaranteed that \eqref{condTOPDOWN2} holds if we ask for
\begin{eqnarray}
  \totrisk(\posfair; \meas{M}_{\mbox{\tiny{\textup{t}}}}, \postarget) & \leq & \frac{\epsilon^4}{2} + \expect_{\X \sim \meas{M}_{\mbox{\tiny{\textup{t}}}}}  \left[ H(\postarget(\X))\right],\label{condTOPDOWN3}
\end{eqnarray}

\newpage

\section{\supplement Experiment Settings} \label{app-exp-settings}

In this \supplement section, we briefly discuss the additional datasets\footnote{Public at: \url{https://github.com/Trusted-AI/AIF360}} and experimental settings included in the subsequent sections. In particular, we highlight the datasets used, the black-boxes post-processed, and specifics of the \topdown algorithm. German Credit and Bank are standard public benchmark datasets in the literature. ACS is a newer dataset with curation details listed in \citep[Section 3]{dhmsRA}.

\subsection*{Datasets}
\begin{itemize}
    \item \textbf{German Credit.} In the \supplement, we additionally consider the German Credit dataset, preprocessed by \aif~\citep{aif360}. The dataset consists of only 1000 examples, which is the smallest of the 3 datasets considered. On the other hand, the dataset provided by \aif contains 57 features, primarily from one-hot encoding.
    \item \textbf{Bank.} Another dataset we consider in the \supplement is the Bank dataset, preprocessed by \aif~\citep{aif360}. The dataset consists 30488 examples, above the German Credit dataset but below the ACS datasets. The dataset also has 57 features which is largely from one-hot encoding.
    \item \textbf{ACS.} The American Community Survey dataset is the dataset we present in the main text. More specifically, we consider the income prediction task (as depicted in the \folktables \python package~\citep{dhmsRA}) over 1-year survey periods in the state of CA. Our of the 3 datasets, ACS provides the largest dataset, with 187475 examples for the 2015 sample of the dataset. Despite this, \folktables only provides 10 features for its prediction task. Through one-hot encoding, this is extended to 29 features.
\end{itemize}
\aif uses a Apache License 2.0 and \folktables uses a MIT License.

Additional \( Z \)-score normalization was used where appropriate. Sensitive attributes are binned into binary and trinary modalities, as specified in the main text (and one-hot encoded for the trinary case).

Each experiment / dataset is used with 5-fold cross-validation and further split such that there are subset partitions for: (1) training the black-box; (2) training a post-processing method; and (3) testing and evaluation.
In particular, we utilize standard cross-validation to split the data into a 80:20 training testing split. The training split is then split randomly equally for separate training of the black-box and post-processing method. The final data splits result in 40:40:20 partitions.

\subsection*{Black-boxes}
\begin{itemize}
    \item \textbf{Random Forest.} As per the main text, we primarily consider a calibrated random forest classifier provided by the \sklearn \python package. The un-calibrated random forest classifier consists of 50 decision trees in an ensemble. Each decision tree has a maximum depth of 4 and is trained on a 10\% subset of the black-box training data. In calibration, 5 cross validation folds are used for Platt scaling.
    \item \textbf{Neural Network.} Additionally to random forests, we consider a calibrated neural network in the \supplement, also provided by \sklearn. The un-calibrated neural network is trained using mostly default parameters provided by \sklearn. The exception to this is the specification of 300 training iterations and the specification of 10\% of the training set to be used for early stopping.
\end{itemize}
The black-boxes are additionally clipped to adhere to \cref{assumptionBUNF} with \( B = 1 \) for all sections except for \cref{app-exp-clip}.

For the criteria we evaluate \topdown and baselines to in the \supplement, we consider those introduced in the main-text alongside \auc as an additional metric for accuracy.

\subsection*{Compute}

Compute was done with no GPUs. Virtual machines were used with RAM 16GB VCPUs 8 VCPU Disk 30GB from [a local HPC facility] (to be named after publication).

\subsection*{Code}
The code used in this submission is attached in the supplementary material, which will be released upon publication.

\subsection*{\topdown Specifics}

The \( \alpha \)-trees learnt by \topdown are initialized as per \cref{fig:bigalpha}. That is, we initialize sub-\( \alpha \)-trees with \( \alpha = 1 \) for each of the modalities of the sensitive attribute. In addition, each split of the \( \alpha \)-tree consists of projects to a specific feature / attribute. The split is either a modality of the discrete feature or a single linear threshold of a continuous feature. In addition, to avoid over-fitting we restrict splits to only those which result in children node that have at least 10\% of the parent node's examples; and at a minimum have at least \( 30 \) examples for each child node.

For \EOO we utilize a Gaussian Naive Bayes classifier from \sklearn with default parameters to fit \( \bayespo \) from data. We note that no fine-tuning was done for this classifier.

In the \supplement, we consider all variants \topdown examined in the main-text. Additionally we consider the symmetric variant of the \SP \topdown approach. We reiterate the two ways of enforcing \SP: as per Section \ref{sec-fairness}, there are two symmetric strategies for \SP. In particular, we aim to match the either the maximum or minimum subgroup to the opposite extreme (conditionally) expected posterior. As such, we can either match to the largest posterior, which we denote as (\( \uparrow \)), or we can match to the smallest posterior (\( \downarrow \)). We already present \SP \( \uparrow \) in the main-text and additionally present evaluation for \SP \( \downarrow \) here.
\newpage
\section{Experiments on Statistical Parity} \label{app-exp-SP-main}

We refer to \supplement, Section \ref{sec-stat-par} for the formal aspects of handling statistical parity. 
% Talk SP Need baselines 
\noindent\textbf{For \SP}, a similar pattern to that of \EOO follows, except the differences are more extreme. In particular, the audacious approach fails to optimize for \SP and instead harms it significantly, but does slightly improve \meandiff.   The audacious update is problematic here as the target \( \postarget \) in the \SP strategy is a constant (and does not take into consideration of the subgroup being updated). As such the audacious approach should not be used for \SP as it will optimize to match the constant \( \postarget \) more harshly. On the other hand, the conservative update variant provides an improvement to \SP whilst antagonizing \meandiff accuracy. Notably, the ``best'' iteration for \SP and \meandiff occurs at its first iteration (which shows interest in early stopping / pruning the \( \alpha \)-tree). This is expected, as there is large shift when changing from an \( \alpha = 1 \) to the initial rooted value (\emph{e.g.}, \eqref{alphatreeoutputCONS}). The pattern of conservative updates being superior to the audacious counterparts is consistent in other datasets and sensitive attribute modalities. Comparing the conservative \SP \topdown to the baselines, discounting \baseeoob for \meandiff, we find that \basespa and \basespb result in lower \SP and \meandiff. This is unsurprising: our \topdown treatment \SP can result in harsh updates; in  \supplement~(pg \pageref{sec-stat-par}), we discuss an alternative approach using ties with optimal transport.

\newpage
\section{Additional Main Text Experiments} \label{app-exp-rest}

In this section, we report the experiments identical to that presented in the main-text, including missing criteria, settings, and the additional German Credit and Bank datasets. Each plot we present provides the binary and trinary sensitive attribute settings over all criteria discussed in the previous setting. 

\begin{sidewaysfigure}[!ht]
    \centering
    \includegraphics[width=\textwidth,trim={0, 6pt, 0, 0},clip]{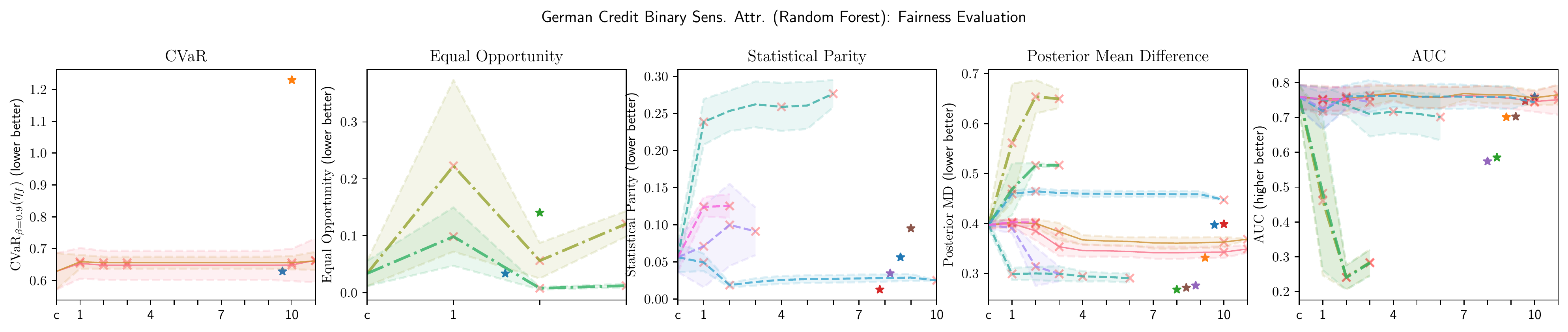}
    \includegraphics[width=\textwidth,trim={0, 6pt, 0, 0},clip]{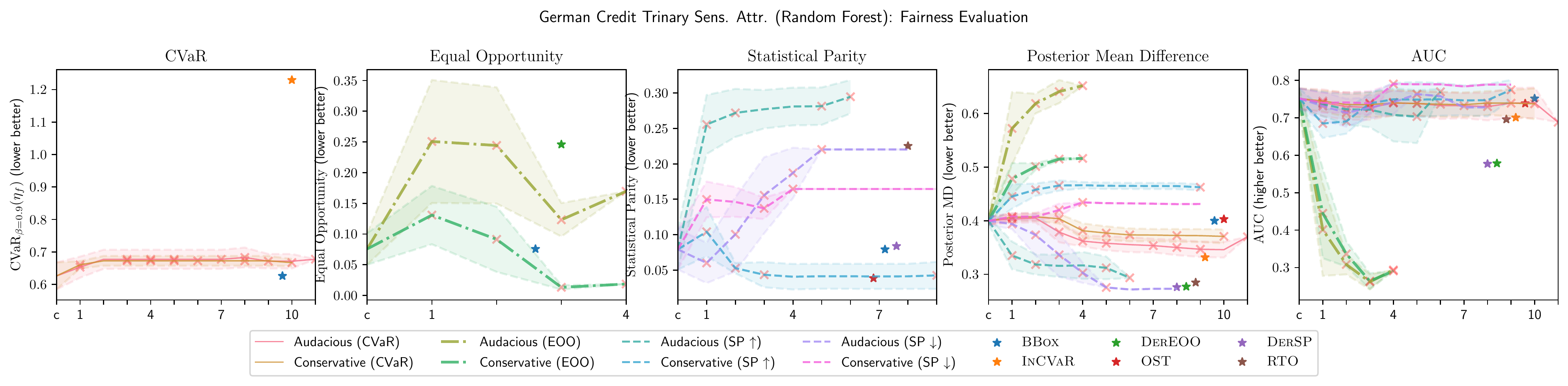}
    \caption{\topdown~optimized for different fairness models evaluated on German Credit with binary (up) and trinary (down) sensitive attributes. Crosses denote when a subgroup's \( \alpha \)-tree is initiated (over any fold). The shade depicts \( \pm \) a standard deviation from the mean. However, this disappears in the case where other folds stop early.}
    \label{fig:rf_german_fairness}
\end{sidewaysfigure}

\begin{sidewaysfigure}[!ht]
    \centering
    \includegraphics[width=\textwidth,trim={0, 6pt, 0, 0},clip]{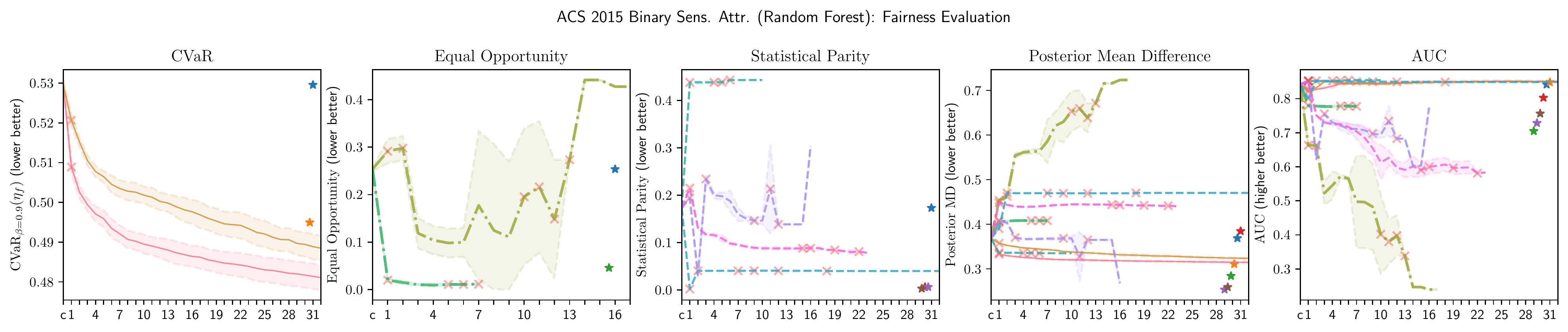}
    \includegraphics[width=\textwidth,trim={0, 6pt, 0, 0},clip]{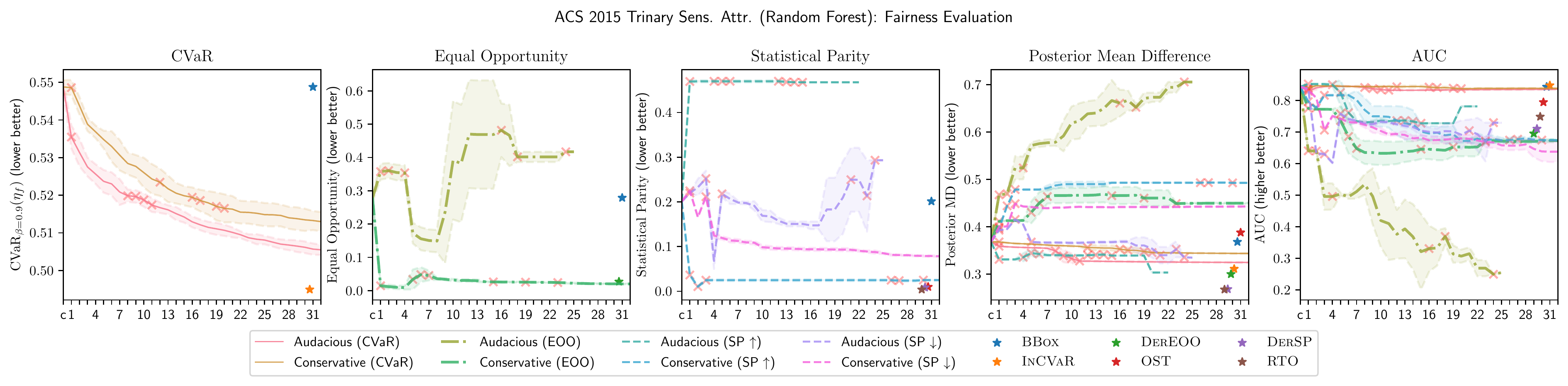}
    \caption{\topdown~optimized over boosting iterations for different fairness models evaluated on ACS 2015 with binary (up) and trinary (down) sensitive attributes. ``c" on the x-axis denotes the clipped black-box. Crosses denote when a subgroup's \( \alpha \)-tree is initiated (over any fold). The shade depicts \( \pm \) a standard deviation from the mean. However, this disappears in the case where other folds stop early.}
    \label{fig:rf_acs_fairness}
\end{sidewaysfigure}

In particular:
\begin{itemize}
    \item \cref{fig:rf_german_fairness} presents the evaluation using a RF black-box with \( B = 1 \) clipping on the German Credit dataset.
    \item \cref{fig:rf_bank_fairness} presents the evaluation using a RF black-box with \( B = 1 \) clipping on the Bank dataset.
    \item \cref{fig:rf_acs_fairness} presents the evaluation using a RF black-box with \( B = 1 \) clipping on the ACS dataset.
\end{itemize}

\subsection*{Fairness Models}
\begin{sidewaysfigure}[!ht]
    \centering
    \includegraphics[width=\textwidth,trim={0, 6pt, 0, 0},clip]{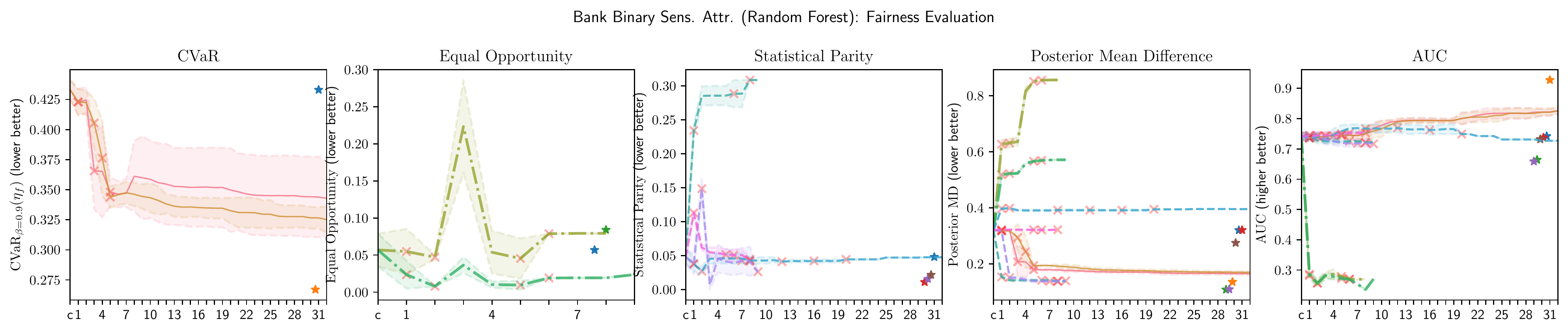}
    \includegraphics[width=\textwidth,trim={0, 6pt, 0, 0},clip]{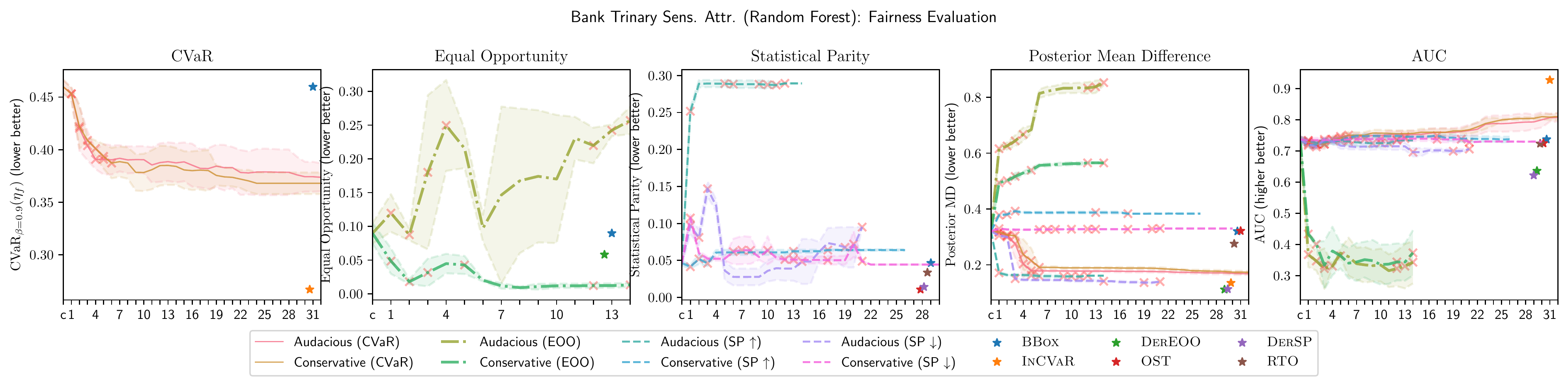}
    \caption{\topdown~optimized for different fairness models evaluated on Bank with binary (up) and trinary (down) sensitive attributes. Crosses denote when a subgroup's \( \alpha \)-tree is initiated (over any fold). The shade depicts \( \pm \) a standard deviation from the mean. However, this disappears in the case where other folds stop early.}
    \label{fig:rf_bank_fairness}
\end{sidewaysfigure}

In comparison to ACS, \cref{fig:rf_bank_fairness} for the Bank dataset performs similarly to the main text figure. There are only slight deviations in the ordering of which \topdown settings perform best. For example, the \( \cvar{} \) optimization of audacious and conservative updates are a lot closer in the Bank dataset than that of the ACS 2015 dataset.

In comparison, the result's of \topdown on the German Credit largely deviate from that of the other experiments. This can be clearly seen in the number of boosting iteration \topdown completes being significantly lower before the entropy stops being decreased (and thus terminating the algorithm). Another major deviation is that \( \cvar{} \) fails to get lowered for both binary and trinary sensitive attribute modalities in the German Credit dataset. Despite this, \EOO and \SP both have slight improvements for the best corresponding \topdown setting (conservative \EOO and conservative \( \SP \uparrow \)), which is consistent with other datasets. This is despite the original classifier's \EOO and \SP being significantly lower than the ACS dataset. However, there is a major cost in the case of \EOO, where the accuracy (both for \meandiff and \auc) is harmed significantly.

A reason for the significantly worse performance, predominantly in \( \cvar{} \) optimization, of \topdown for the German Credit is likely the significantly smaller number of example available in the dataset. Given that there are only 1000 examples and 57 features variables, the 40:40:20 split of the dataset results in the subsets to not be representative of the entire dataset's support. Additionally, \( \cvar{} \) is strongly tied to the cross-entropy loss function and empirical risk minimization~\citep{wmFR,ruOO}. As such, given the nonrepresentative subsets of the dataset used for training \topdown, minimizing the \( \cvar{} \) for low sample inputs is difficult.
Thus for such a failure cases, one should confirm that \topdown is not decreasing fairness to prevent social harms.

\clearpage

\clearpage

\clearpage

\clearpage
\newpage
\section{Neural Network Experiments} \label{app-exp-nn}

\begin{sidewaysfigure}[!ht]
    \centering
    \includegraphics[width=\textwidth,trim={0, 6pt, 0, 0},clip]{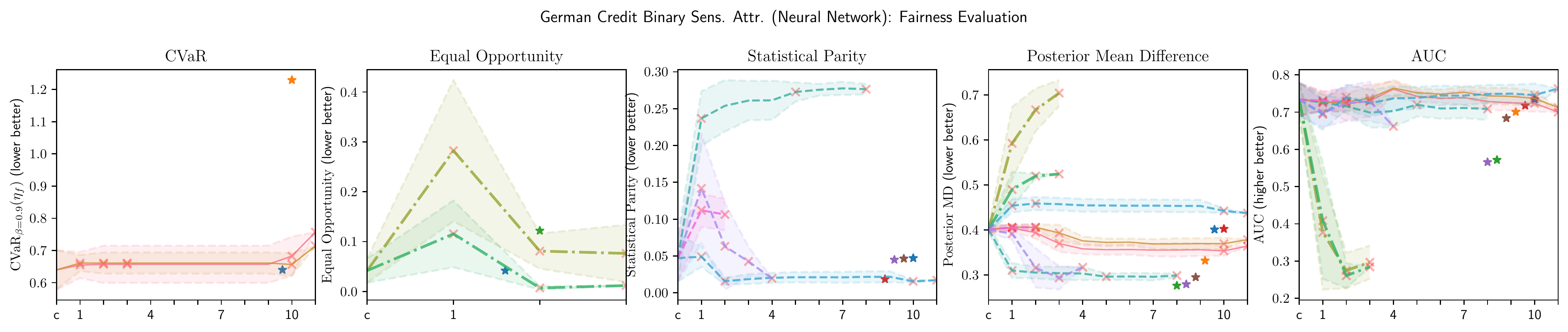}
    \includegraphics[width=\textwidth,trim={0, 6pt, 0, 0},clip]{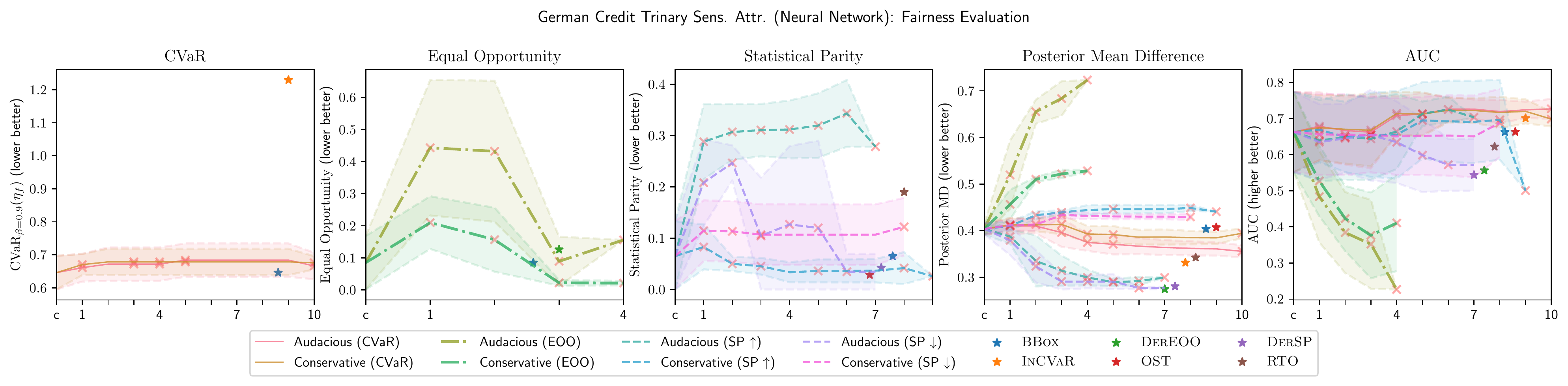}
   \vspace{-0.5cm}
    \caption{\topdown~optimized for different fairness models evaluated on German Credit with binary (up) and trinary (down) sensitive attributes. Crosses denote when a subgroup's \( \alpha \)-tree is initiated (over any fold). The shade depicts \( \pm \) a standard deviation from the mean. However, this disappears in the case where other folds stop early.}
    \label{fig:mlp_german_fairness}
\end{sidewaysfigure}

\begin{sidewaysfigure}[!ht]
    \centering
    \includegraphics[width=\textwidth,trim={0, 6pt, 0, 0},clip]{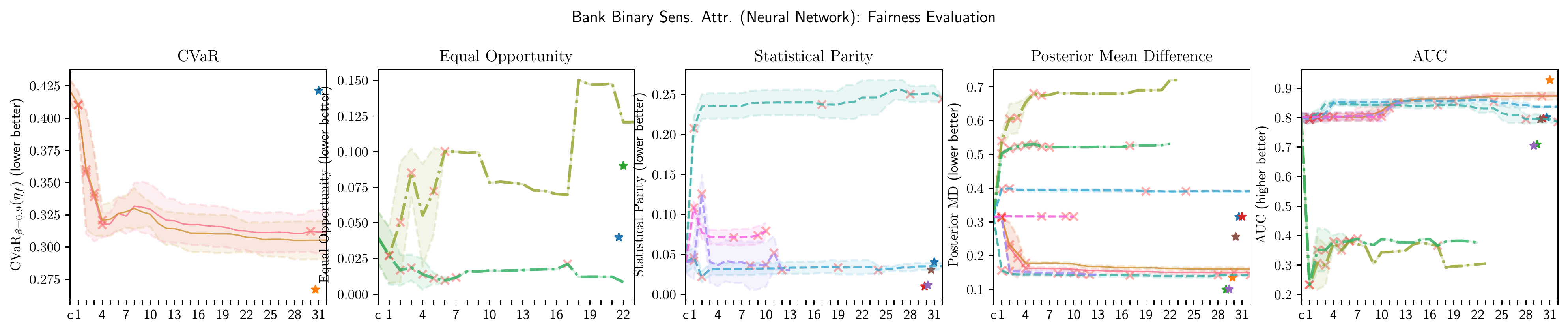}
    \includegraphics[width=\textwidth,trim={0, 6pt, 0, 0},clip]{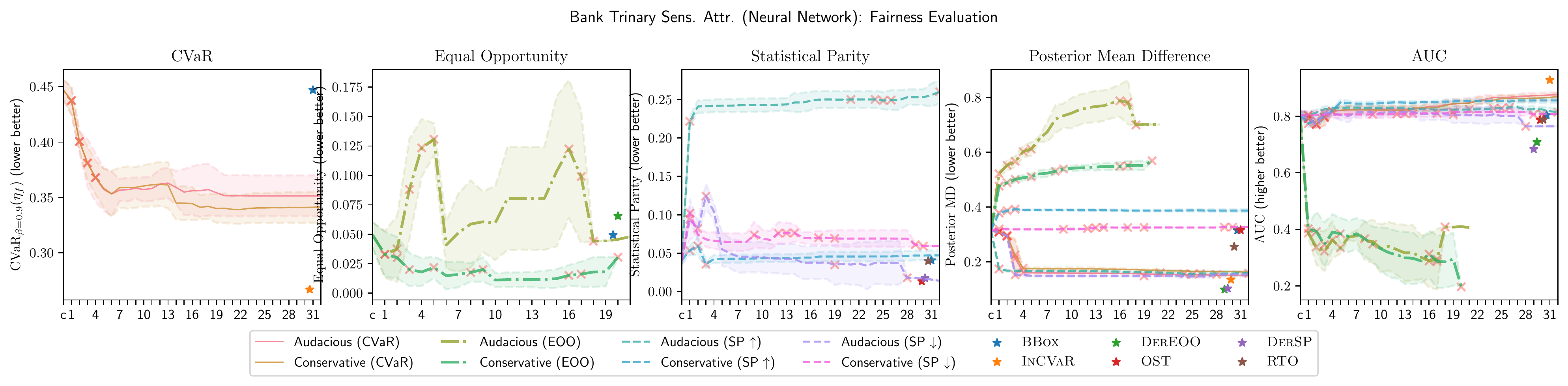}
   \vspace{-0.5cm}
    \caption{\topdown~optimized for different fairness models evaluated on Bank with binary (up) and trinary (down) sensitive attributes. Crosses denote when a subgroup's \( \alpha \)-tree is initiated (over any fold). The shade depicts \( \pm \) a standard deviation from the mean. However, this disappears in the case where other folds stop early.}
    \label{fig:mlp_bank_fairness}
\end{sidewaysfigure}

\begin{sidewaysfigure}[!ht]
    \centering
    \includegraphics[width=\textwidth,trim={0, 6pt, 0, 0},clip]{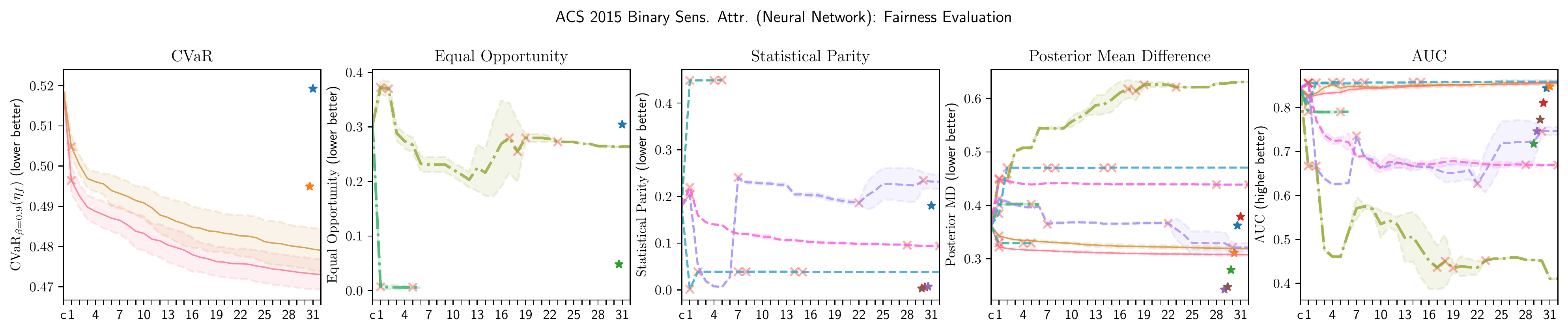}
    \includegraphics[width=\textwidth,trim={0, 6pt, 0, 0},clip]{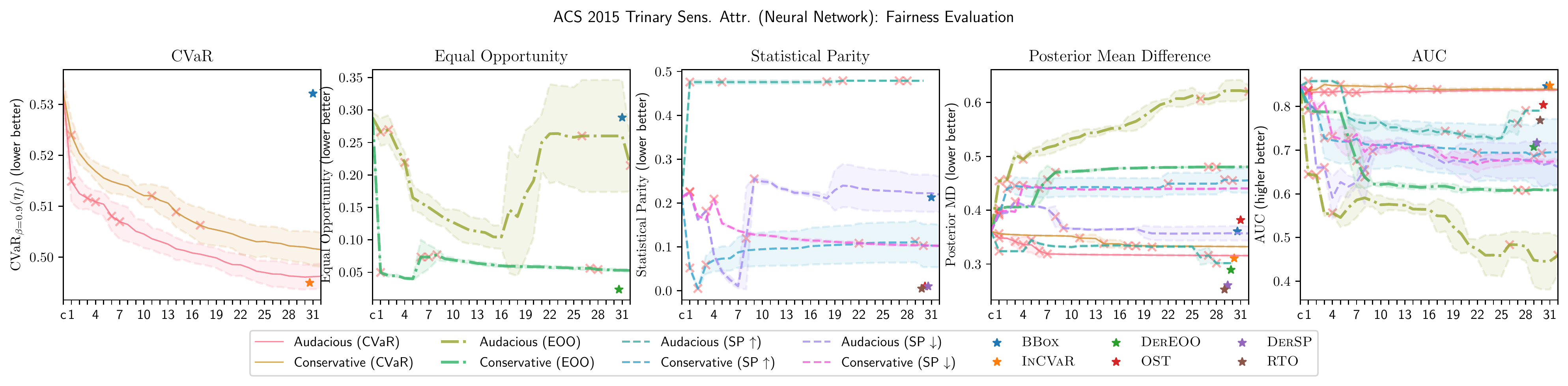}
   \vspace{-0.5cm}
    \caption{\topdown~optimized for different fairness models evaluated on Bank with binary (up) and trinary (down) sensitive attributes. Crosses denote when a subgroup's \( \alpha \)-tree is initiated (over any fold). The shade depicts \( \pm \) a standard deviation from the mean. However, this disappears in the case where other folds stop early.}
    \label{fig:mlp_acs_fairness}
\end{sidewaysfigure}

In this \supplement section, we repeat all experiments evaluating different fairness models and proxy sensitive attributes using the neural network (NN) black-box.
\cref{fig:mlp_german_fairness,fig:mlp_bank_fairness,fig:mlp_acs_fairness} presents neural network equivalent plots to that of \cref{fig:rf_german_fairness,fig:rf_bank_fairness,fig:rf_acs_fairness}.

In particular:
\begin{itemize}
    \item \cref{fig:mlp_german_fairness} presents the evaluation using a NN black-box with \( B = 1 \) clipping on the German Credit dataset.
    \item \cref{fig:mlp_bank_fairness} presents the evaluation using a NN black-box with \( B = 1 \) clipping on the Bank dataset.
    \item \cref{fig:mlp_acs_fairness} presents the evaluation using a NN black-box with \( B = 1 \) clipping on the ACS dataset.
\end{itemize}

When comparing the NN experiments to the experiments corresponding to that of the random forest (RF) black-box experiments, only minor deviation can be seen with most trends staying the same. One consistent deviation is that the \( \cvar{} \) criterion and accuracy measures (\meandiff and \auc) are frequently smaller at the initial and final point of boosting. This comes from the strong representation power of the NN black-box being translated from the initial black-box to the final wrapper classifier. In this regard, switching to a NN did not help the optimization of \( \cvar{} \) for the German Credit dataset, see \cref{fig:mlp_german_fairness}.

\clearpage

\clearpage

\clearpage

\clearpage
\newpage
\section{Proxy Sensitive Attributes}\label{app-exp-proxy}

We examine the use of sensitive attribute proxies to remove sensitive attribute requirements at test time. In particular, we use a decision tree with a maximum depth of \( 8 \) to predict sensitive attributes (from other features) as a proxy to the true sensitive attribute. We present results for both RF and NN black-boxes.

\begin{sidewaysfigure}[!ht]
    \centering
\begin{tabular}{cc}
    \includegraphics[width=0.45\columnwidth,trim={0, 5pt, 0, 0},clip]{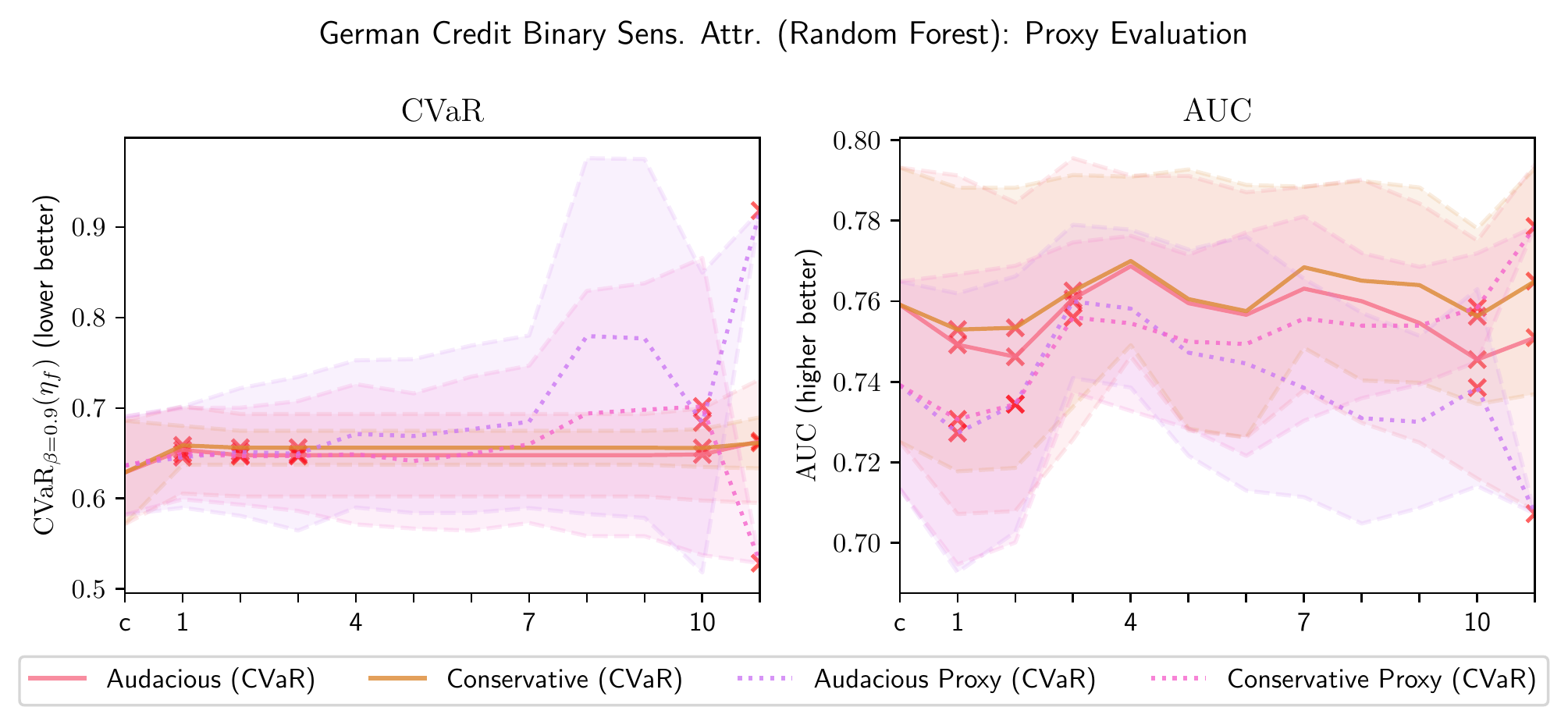}
    &
    \includegraphics[width=0.45\columnwidth,trim={0, 5pt, 0, 0},clip]{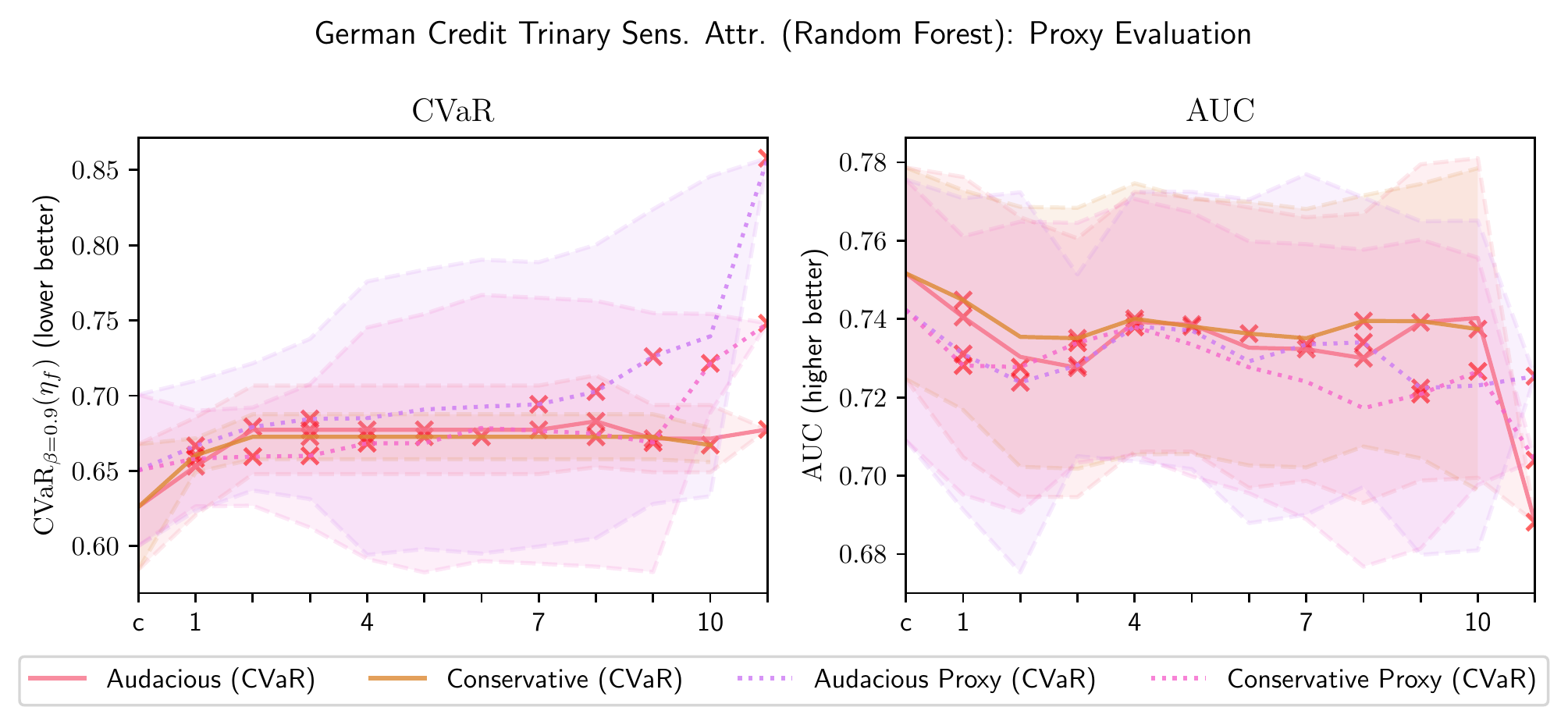}
    \\
    \includegraphics[width=0.45\columnwidth,trim={0, 5pt, 0, 0},clip]{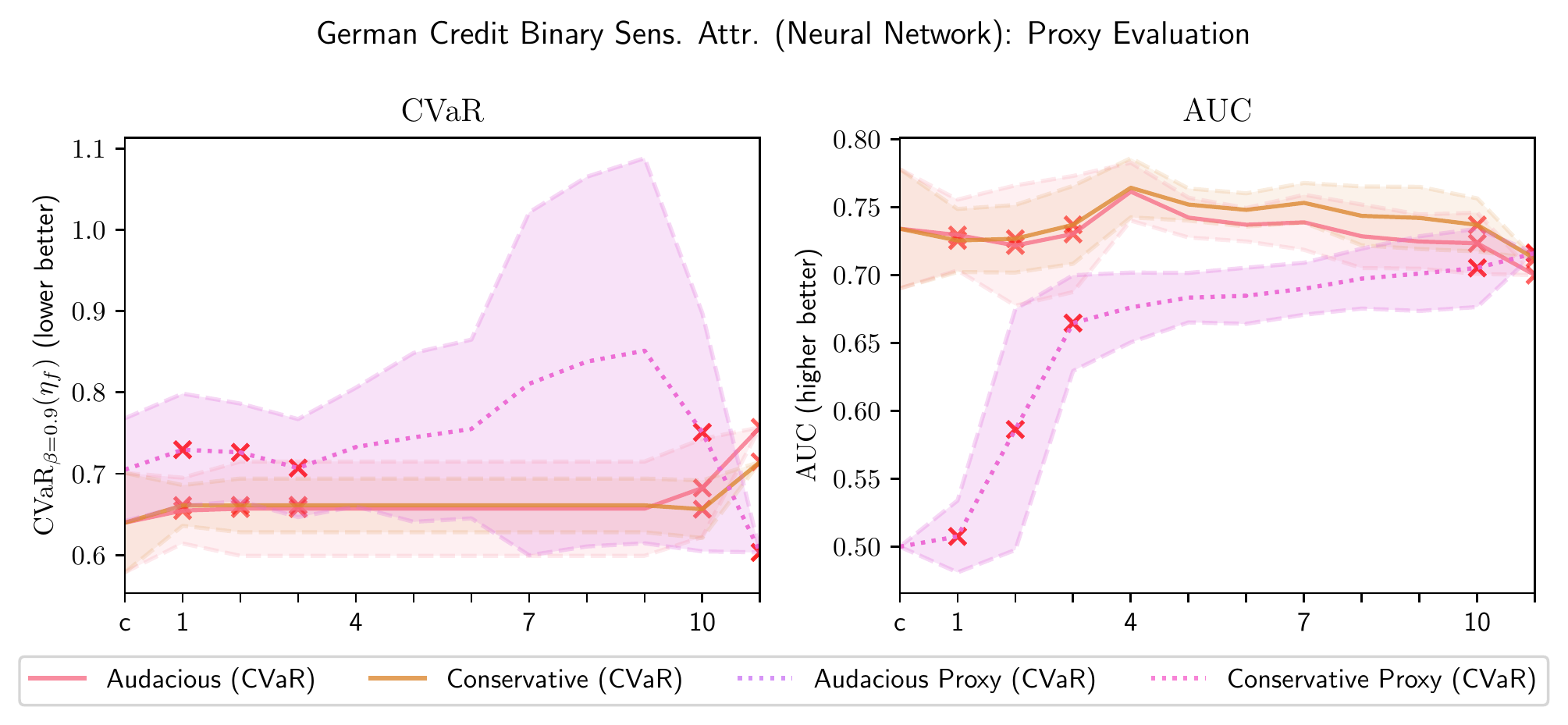}
    &
    \includegraphics[width=0.45\columnwidth,trim={0, 5pt, 0, 0},clip]{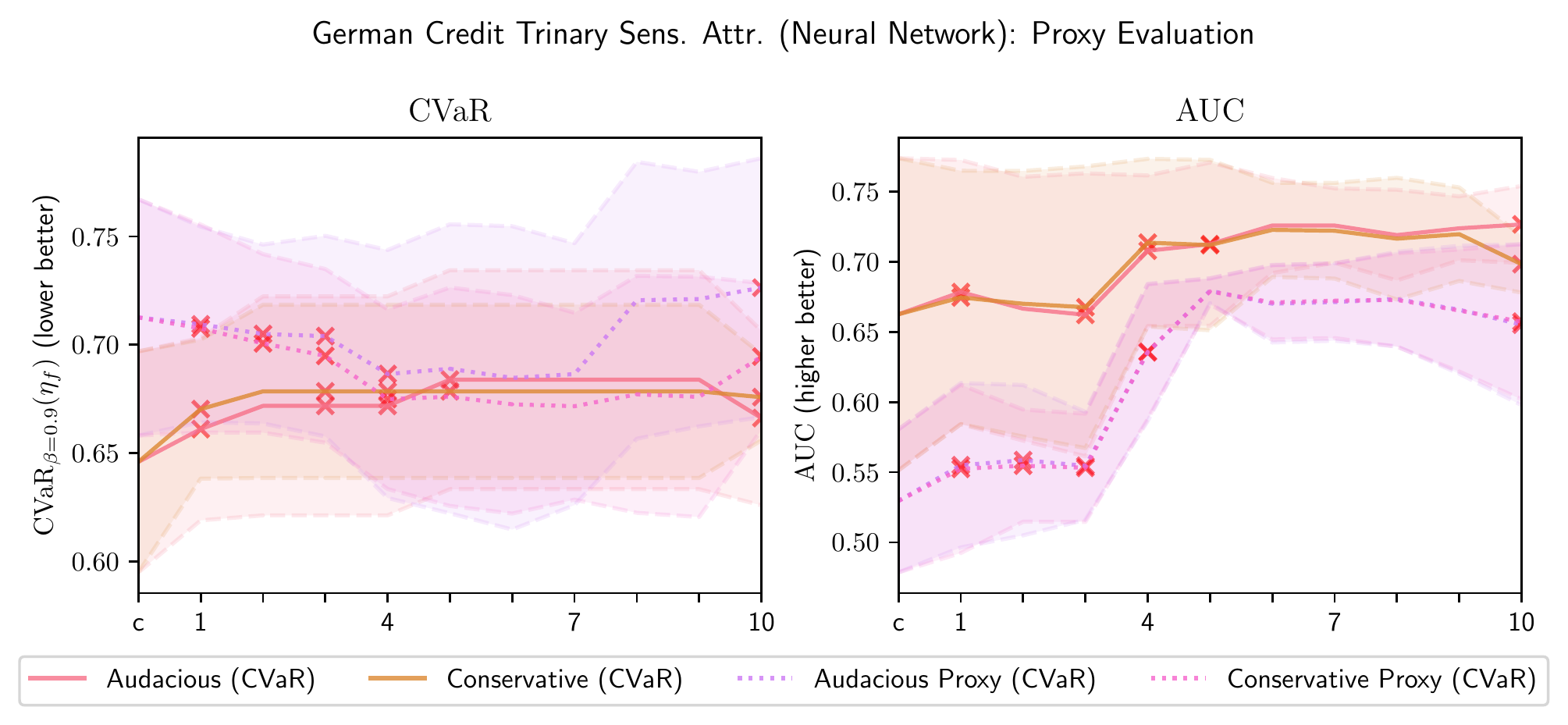}
\end{tabular}
    \caption{RF (top) and MLP (bottom) evaluation of replacing sensitive attributes with a proxy decision tree on the German Credit datasets.}
    \label{fig:german_stree}
\end{sidewaysfigure}

\begin{sidewaysfigure}[!ht]
    \centering
\begin{tabular}{cc}
    \includegraphics[width=0.45\columnwidth,trim={0, 5pt, 0, 0},clip]{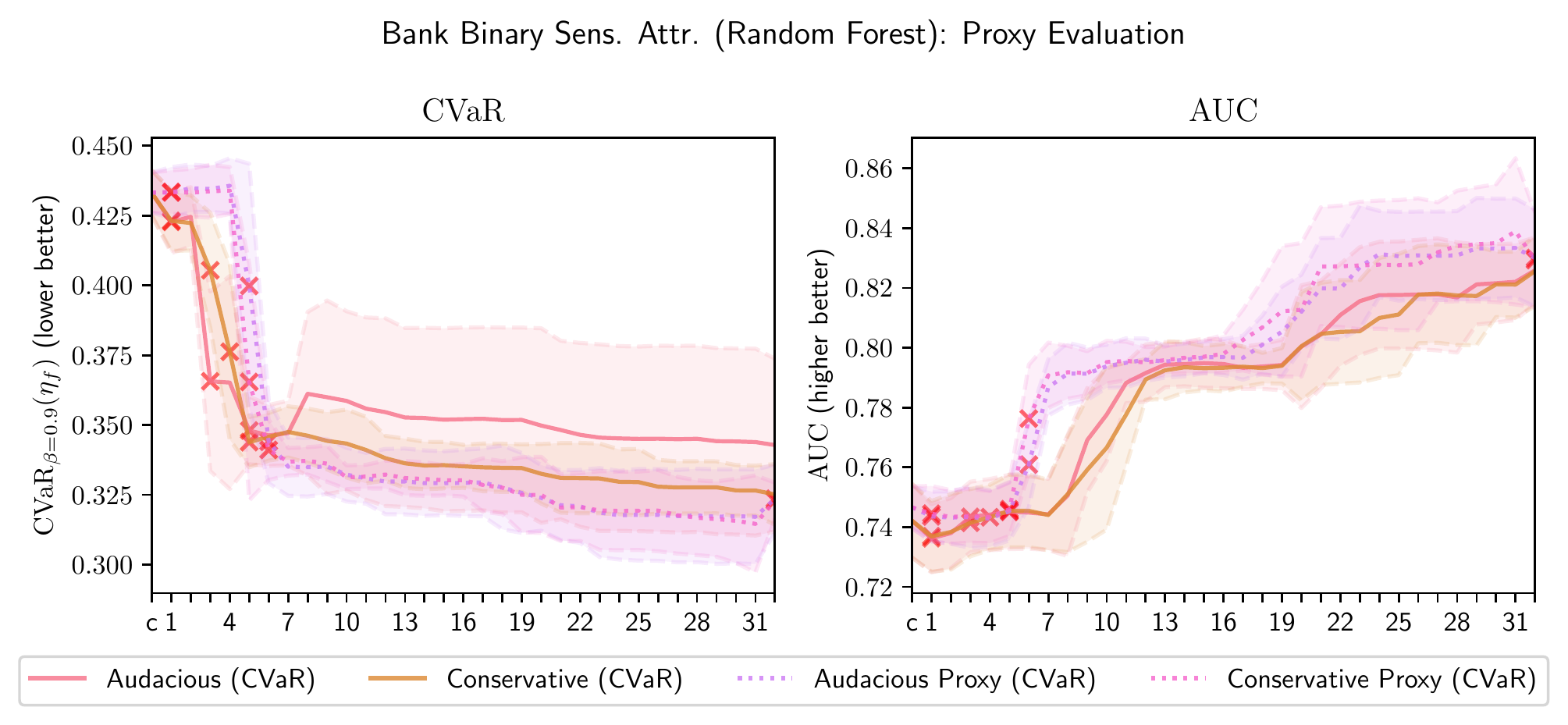}
    &
    \includegraphics[width=0.45\columnwidth,trim={0, 5pt, 0, 0},clip]{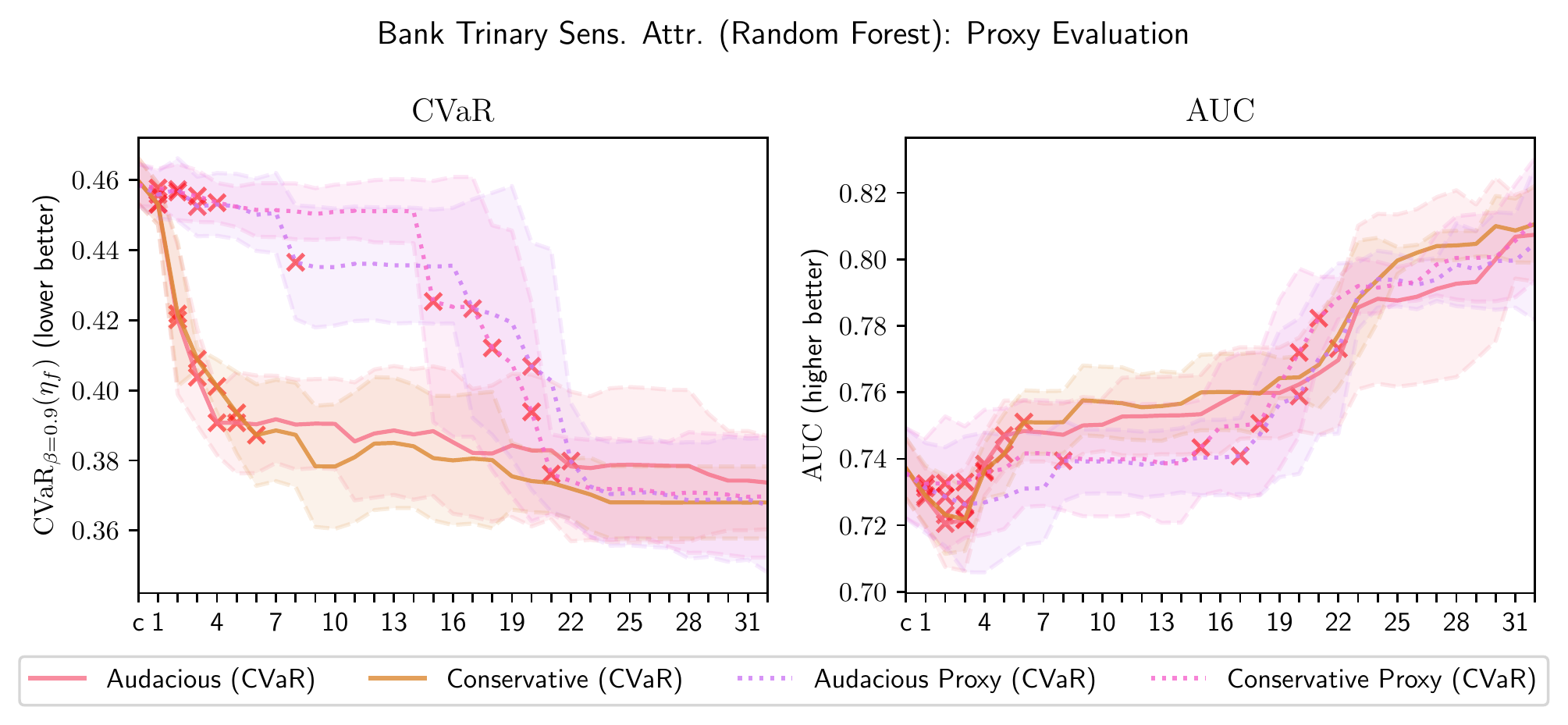}
    \\
    \includegraphics[width=0.45\columnwidth,trim={0, 5pt, 0, 0},clip]{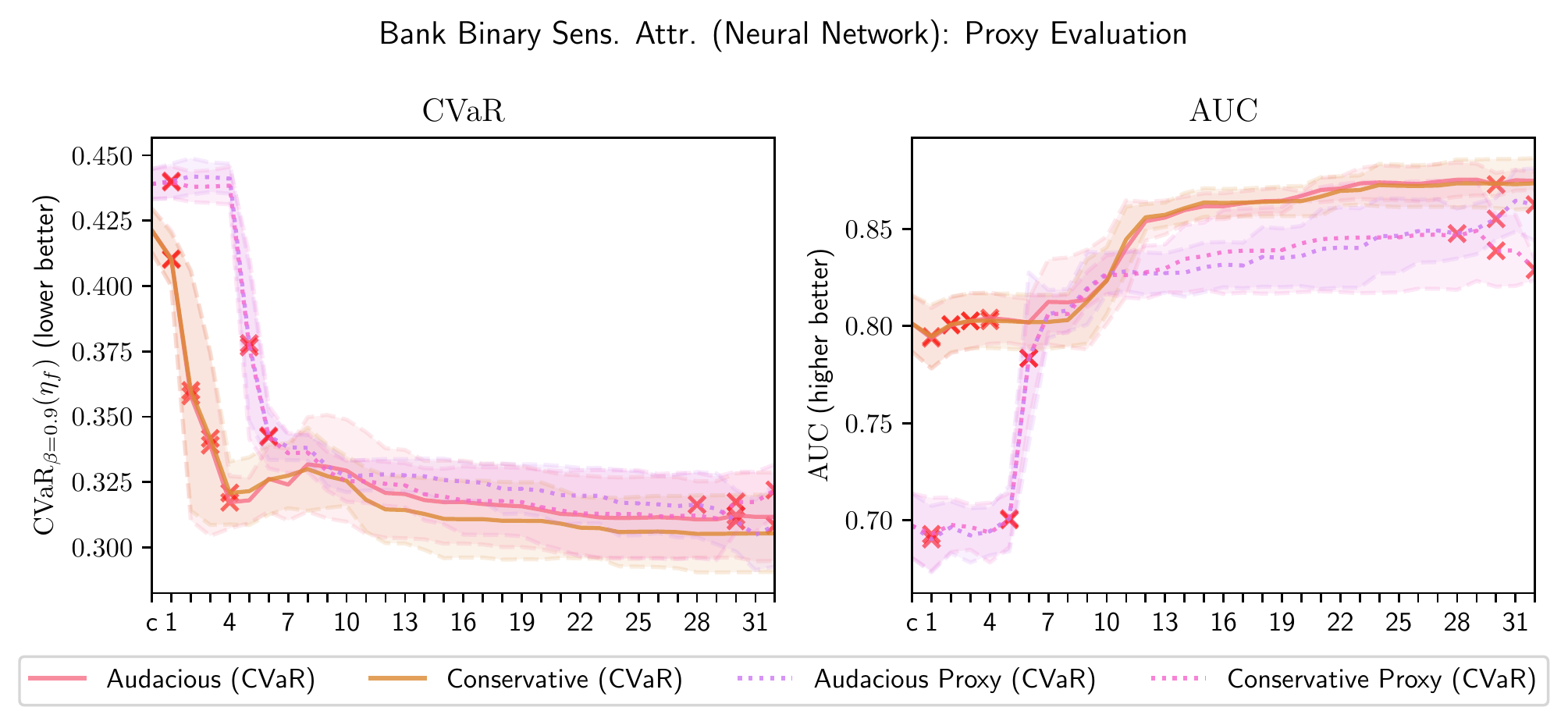}
    &
    \includegraphics[width=0.45\columnwidth,trim={0, 5pt, 0, 0},clip]{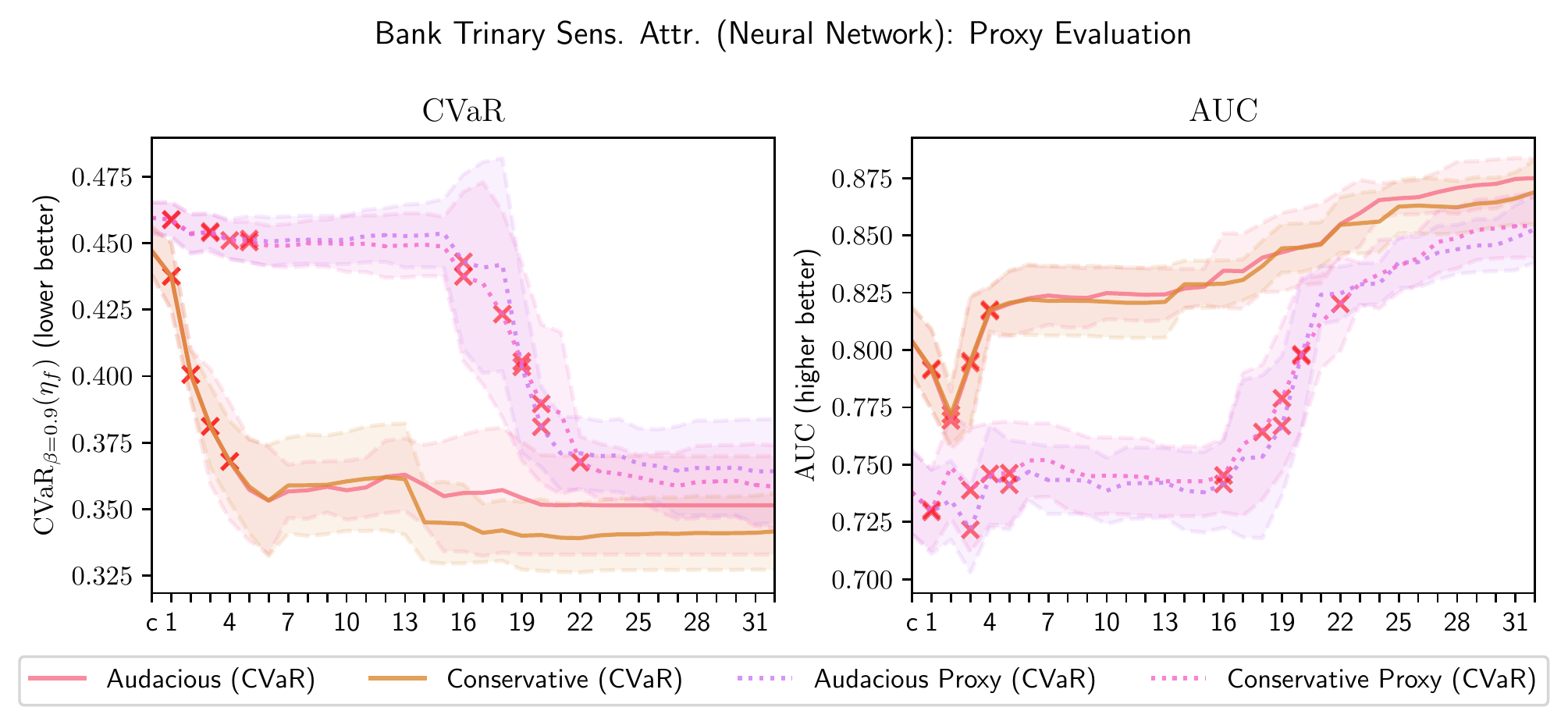}
\end{tabular}
    \caption{RF (top) and MLP (bottom) evaluation of replacing sensitive attributes with a proxy decision tree on the Bank datasets.}
    \label{fig:bank_stree}
\end{sidewaysfigure}

\begin{sidewaysfigure}
    \centering
\begin{tabular}{cc}
    \includegraphics[width=0.45\columnwidth,trim={0, 5pt, 0, 0},clip]{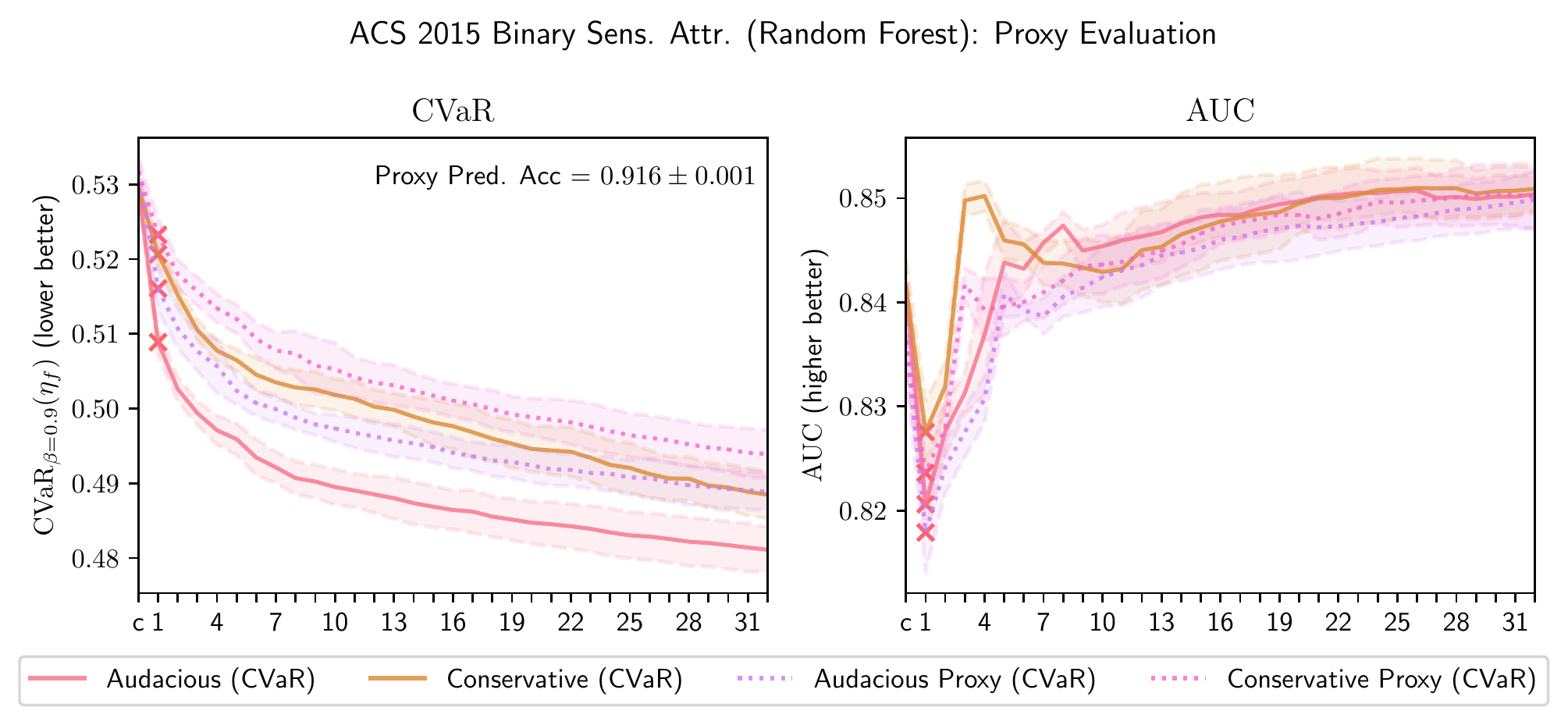}
    &
    \includegraphics[width=0.45\columnwidth,trim={0, 5pt, 0, 0},clip]{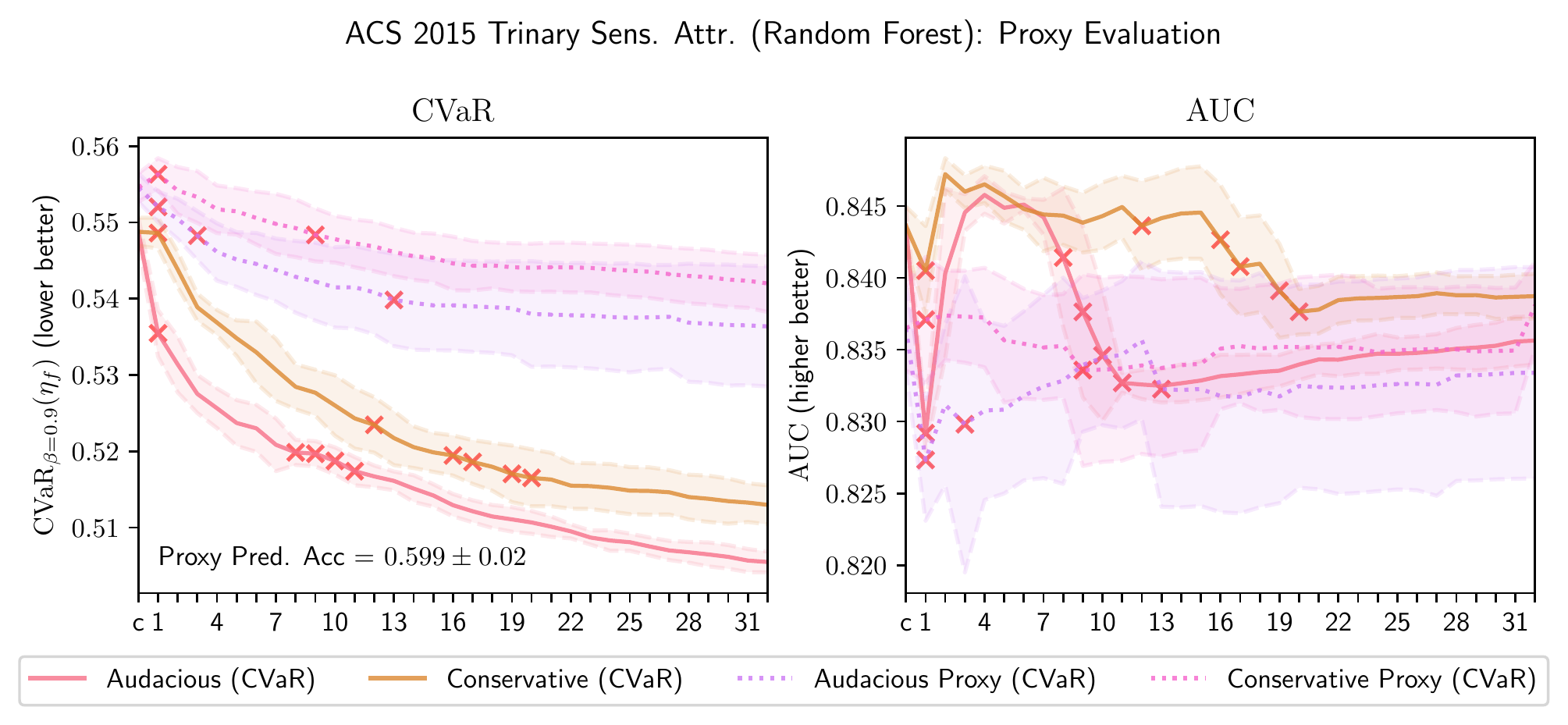}
    \\
    \includegraphics[width=0.45\columnwidth,trim={0, 5pt, 0, 0},clip]{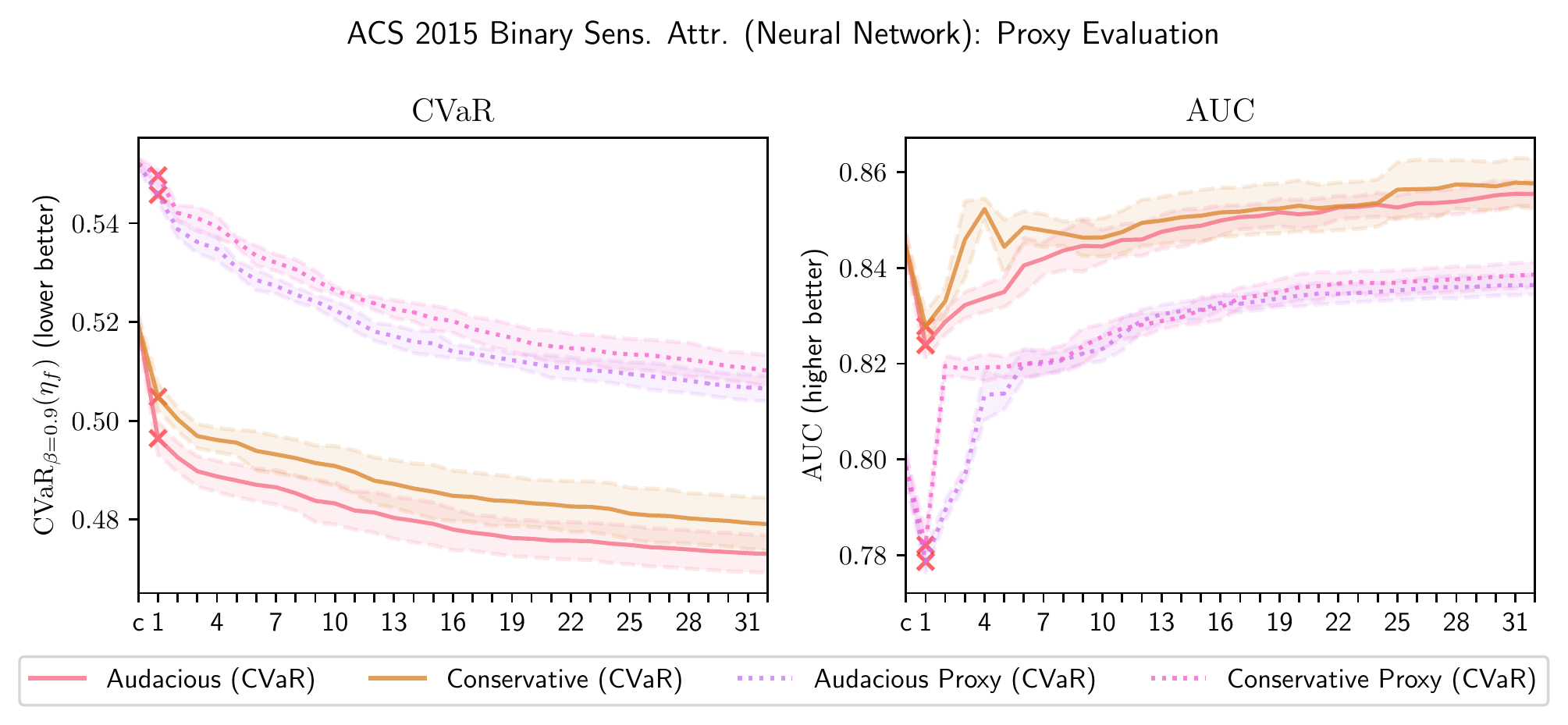}
    &
    \includegraphics[width=0.45\columnwidth,trim={0, 5pt, 0, 0},clip]{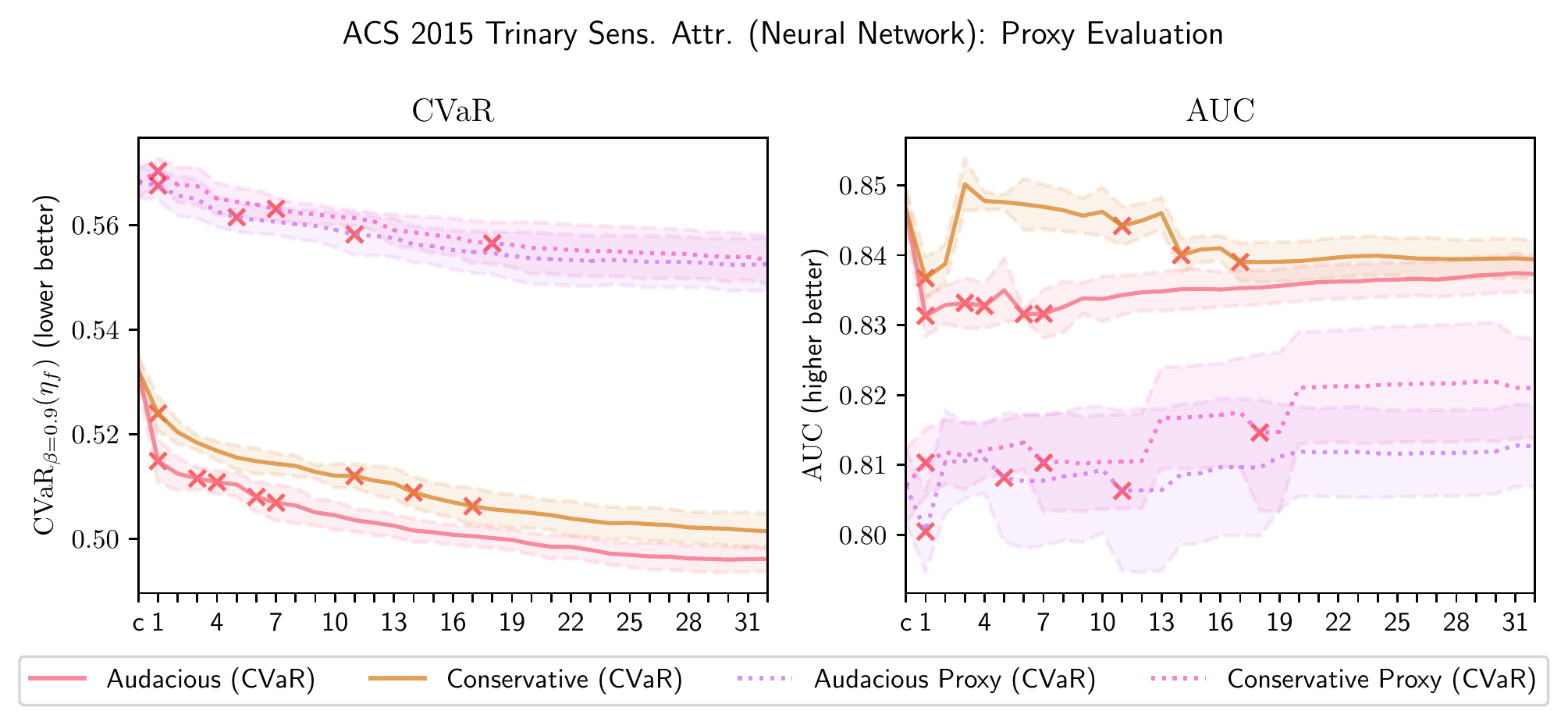}
\end{tabular}
    \caption{RF (top) and MLP (bottom) evaluation of replacing sensitive attributes with a proxy decision tree on the ACS 2015 datasets.}
    \label{fig:acs_stree}
\end{sidewaysfigure}

In particular:
\begin{itemize}
    \item \cref{fig:german_stree} presents the proxy sensitive attribute evaluation using a RF and NN black-box with \( B = 1 \) clipping on the German Credit dataset.
    \item \cref{fig:bank_stree} presents the proxy sensitive attribute evaluation using a RF and NN black-box with \( B = 1 \) clipping on the Bank dataset.
    \item \cref{fig:acs_stree} presents the proxy sensitive attribute evaluation using a RF and NN black-box with \( B = 1 \) clipping on the ACS dataset.
\end{itemize}

\cref{fig:acs_stree} (top) presents the RF \topdown proxy sensitive attribute experiments results of the ACS 2015 dataset not present in the main text. We focus on the binary case (left). Unsurprisingly, the proxy increases the variance of \( \cvar{} \) and \auc whilst also being worse than their non-proxy counterparts; but still manages to improve $\cvar{}$ and \auc at the end (with an initial dip quickly erased for the latter criterion). Remark the non-trivial nature of the proxy approach, as growing the $\alpha$-tree is based on groups learned at the decision tree leaves \textit{but} the $\cvar{}$ computation still relies on the \textit{original} sensitive grouping.

\cref{fig:german_stree,fig:bank_stree} (top) presents the RF \topdown proxy sensitive attribute results of the German Credit and Bank datasets. The ACS and Bank experiments presented here are similar to that presented in the main text. For German Credit, similar degradation in \(\cvar{}\) in the non-proxy case can be seen for \topdown results using proxy attributes.

%\begin{figure}
%    \centering
%\begin{tabular}{cc}
%    \includegraphics[width=0.45\columnwidth,trim={0, 5pt, 0, 0},clip]{Figs/experiments_plots/rf/german_credit_stree.pdf}
%    &
%    \includegraphics[width=0.45\columnwidth,trim={0, 5pt, 0, 0},clip]{Figs/experiments_plots/rf/german_credit_multi_stree.pdf}
%\end{tabular}
%   \vspace{-0.4cm}
%    \caption{RF evaluation of replacing sensitive attributes with a proxy decision tree on the German Credit datasets.}
%    \label{fig:rf_german_stree}
%\end{figure}
%
%\begin{figure}
%    \centering
%\begin{tabular}{cc}
%    \includegraphics[width=0.45\columnwidth,trim={0, 5pt, 0, 0},clip]{Figs/experiments_plots/rf/bank_stree.pdf}
%    &
%    \includegraphics[width=0.45\columnwidth,trim={0, 5pt, 0, 0},clip]{Figs/experiments_plots/rf/bank_multi_stree.pdf}
%\end{tabular}
%   \vspace{-0.4cm}
%    \caption{RF evaluation of replacing sensitive attributes with a proxy decision tree on the Bank datasets.}
%    \label{fig:rf_bank_stree}
%\end{figure}
%
%\begin{figure}
%    \centering
%\begin{tabular}{cc}
%    \includegraphics[width=0.45\columnwidth,trim={0, 5pt, 0, 0},clip]{Figs/experiments_plots/rf/acs_stree.pdf}
%    &
%    \includegraphics[width=0.45\columnwidth,trim={0, 5pt, 0, 0},clip]{Figs/experiments_plots/rf/acs_multi_stree.pdf}
%\end{tabular}
%   \vspace{-0.4cm}
%    \caption{RF replacing sensitive attributes with a proxy decision tree on the ACS 2015 dataset (see text).}
%    \label{fig:rf_acs_stree}
%\end{figure}

When comparing to the MLP variants (\cref{fig:german_stree,fig:bank_stree,fig:acs_stree} bottom), results are quite similar with slight increases in \( \cvar{} \) from the change in black-box. One notable difference can be seen in \cref{fig:acs_stree}. In particular, the proxy and regular curves do not ``cross". This indicates that (given that the sensitive attribute proxy used is the same as RF) the black-box being post-processed is an important consideration in the use of proxies. In particular, as RF has a higher / worse initial \( \cvar{} \), which is highly tied to the loss / cross entropy of the black-box, the robustness of the black-box needs to be considered.

%\begin{figure}
%    \centering
%\begin{tabular}{cc}
%    \includegraphics[width=0.45\columnwidth,trim={0, 5pt, 0, 0},clip]{Figs/experiments_plots/mlp/german_credit_stree.pdf}
%    &
%    \includegraphics[width=0.45\columnwidth,trim={0, 5pt, 0, 0},clip]{Figs/experiments_plots/mlp/german_credit_multi_stree.pdf}
%\end{tabular}
%   \vspace{-0.4cm}
%    \caption{MLP evaluation of replacing sensitive attributes with a proxy decision tree on the German Credit datasets.}
%    \label{fig:mlp_german_stree}
%\end{figure}
%
%\begin{figure}
%    \centering
%\begin{tabular}{cc}
%    \includegraphics[width=0.45\columnwidth,trim={0, 5pt, 0, 0},clip]{Figs/experiments_plots/mlp/bank_stree.pdf}
%    &
%    \includegraphics[width=0.45\columnwidth,trim={0, 5pt, 0, 0},clip]{Figs/experiments_plots/mlp/bank_multi_stree.pdf}
%\end{tabular}
%   \vspace{-0.4cm}
%    \caption{MLP evaluation of replacing sensitive attributes with a proxy decision tree on the Bank datasets.}
%    \label{fig:mlp_bank_stree}
%\end{figure}
%
%\begin{figure}
%    \centering
%\begin{tabular}{cc}
%    \includegraphics[width=0.45\columnwidth,trim={0, 5pt, 0, 0},clip]{Figs/experiments_plots/mlp/acs_stree.pdf}
%    &
%    \includegraphics[width=0.45\columnwidth,trim={0, 5pt, 0, 0},clip]{Figs/experiments_plots/mlp/acs_multi_stree.pdf}
%\end{tabular}
%   \vspace{-0.4cm}
%    \caption{MLP evaluation of replacing sensitive attributes with a proxy decision tree on the ACS datasets.}
%    \label{fig:mlp_acs_stree}
%\end{figure}

%%%

\clearpage

\clearpage

\clearpage

\clearpage
\newpage
\section{Distribution shift}\label{app-exp-shift}
\begin{sidewaysfigure}[!ht]
    \centering
    \includegraphics[width=\textwidth,trim={0, 6pt, 0, 0},clip]{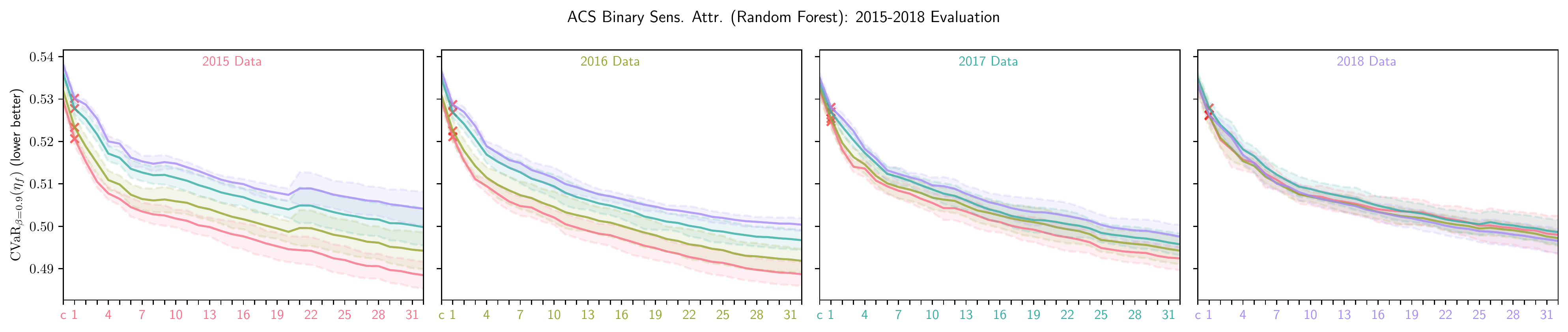}
    \includegraphics[width=\textwidth,trim={0, 6pt, 0, 0},clip]{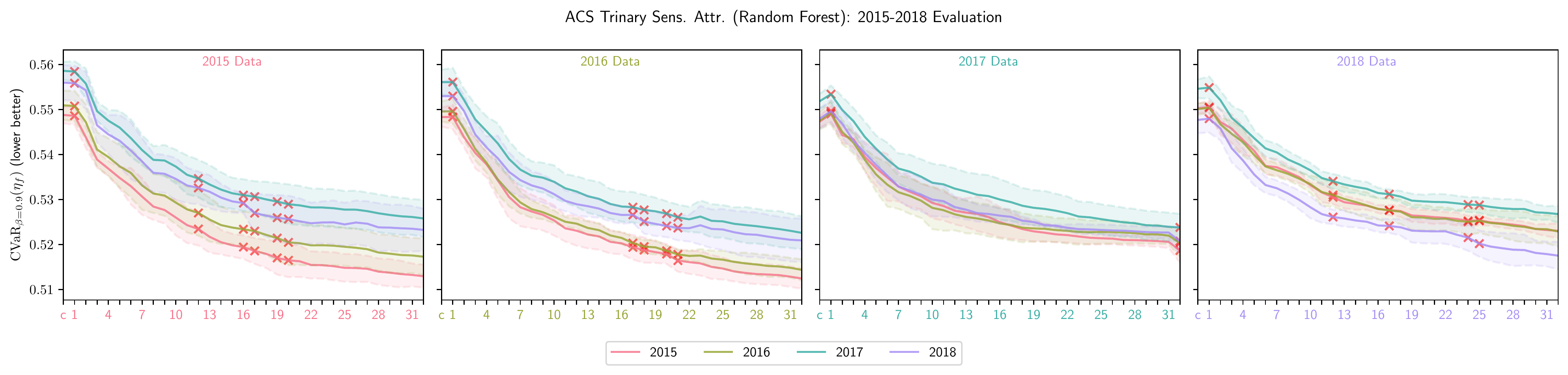}
   \vspace{-0.5cm}
    \caption{Random forest black-box conservative \( \cvar{} \) wrapper trained for ACS 2015 to 2018 datasets Each plot is trained on a different dataset year. Each curve colour, indicates the data being used to evaluate the wrapper.}
    \label{fig:rf_yeareval_conservative}
\end{sidewaysfigure}

\begin{sidewaysfigure}[!ht]
    \centering
    \includegraphics[width=\textwidth,trim={0, 6pt, 0, 0},clip]{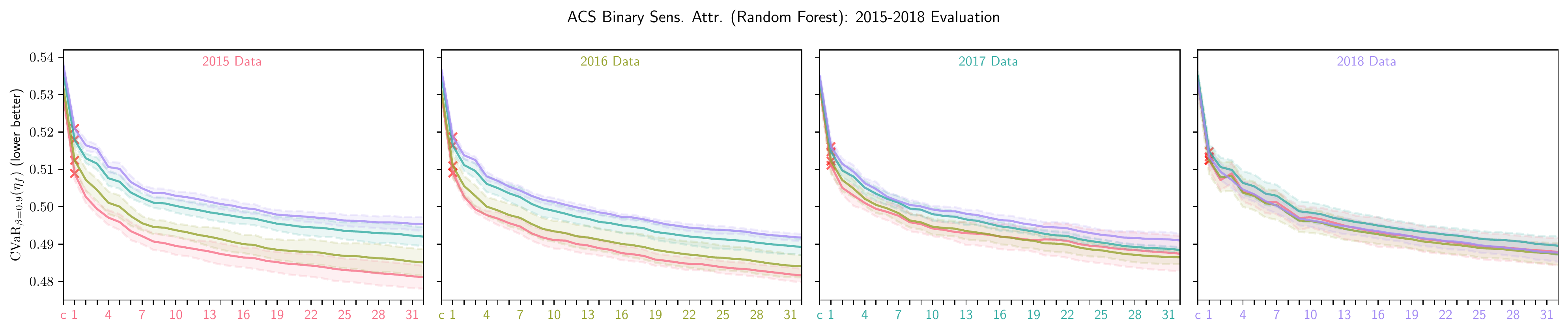}
    \includegraphics[width=\textwidth,trim={0, 6pt, 0, 0},clip]{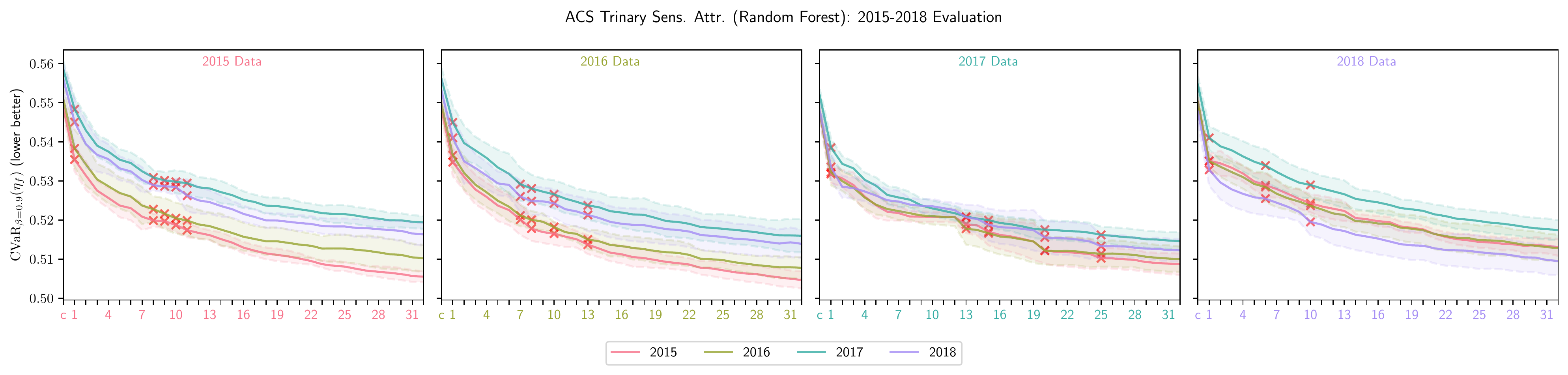}
   \vspace{-0.5cm}
    \caption{Random forest black-box aggressive \( \cvar{} \) wrapper trained for ACS 2015 to 2018 datasets Each plot is trained on a different dataset year. Each curve colour, indicates the data being used to evaluate the wrapper.}
    \label{fig:rf_yeareval_aggressive}
\end{sidewaysfigure}

\begin{sidewaysfigure}[!ht]
    \centering
    \includegraphics[width=\textwidth,trim={0, 6pt, 0, 0},clip]{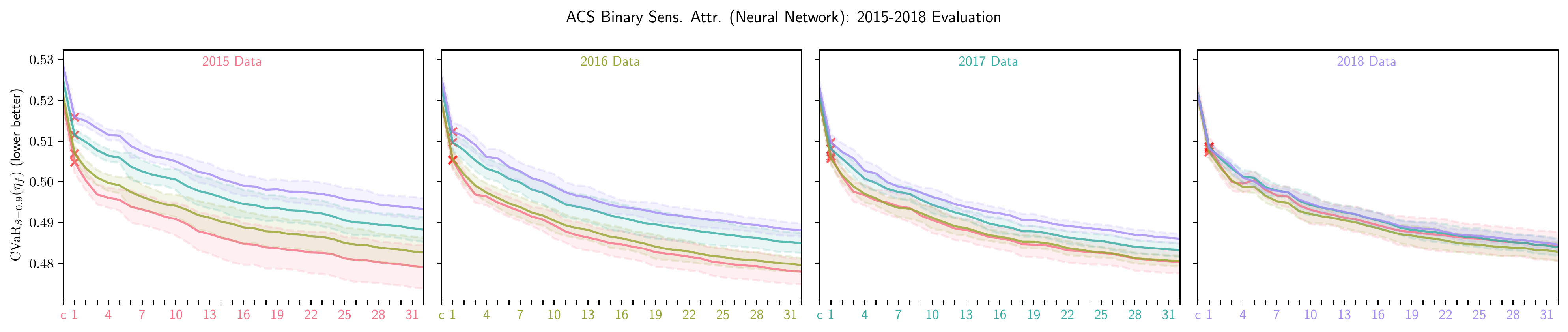}
    \includegraphics[width=\textwidth,trim={0, 6pt, 0, 0},clip]{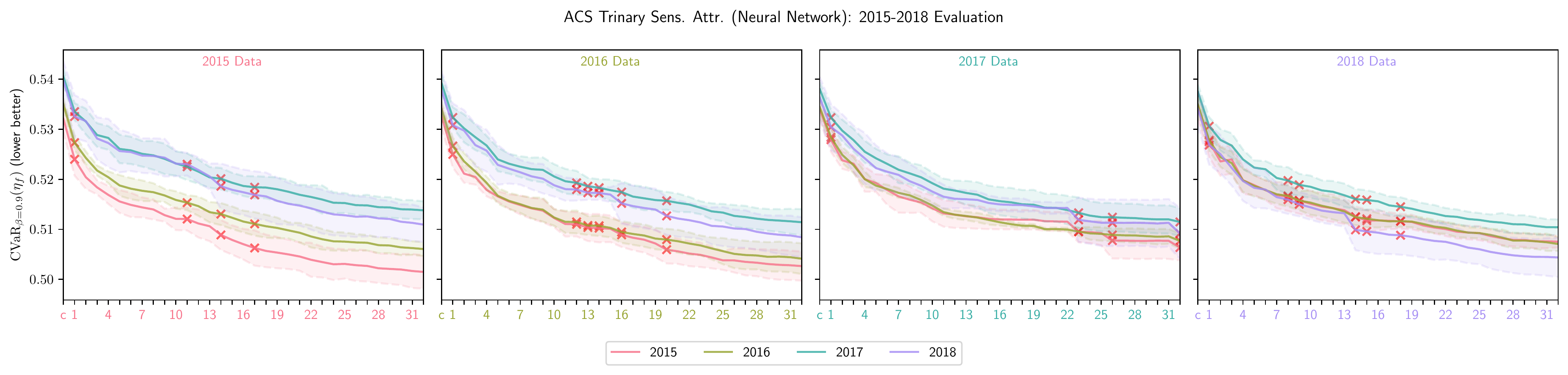}
   \vspace{-0.5cm}
    \caption{Neural Network black-box conservative \( \cvar{} \) wrapper trained for ACS 2015 to 2018 datasets Each plot is trained on a different dataset year. Each curve colour, indicates the data being used to evaluate the wrapper.}
    \label{fig:mlp_yeareval_conservative}
\end{sidewaysfigure}

\begin{sidewaysfigure}[!ht]
    \centering
    \includegraphics[width=\textwidth,trim={0, 6pt, 0, 0},clip]{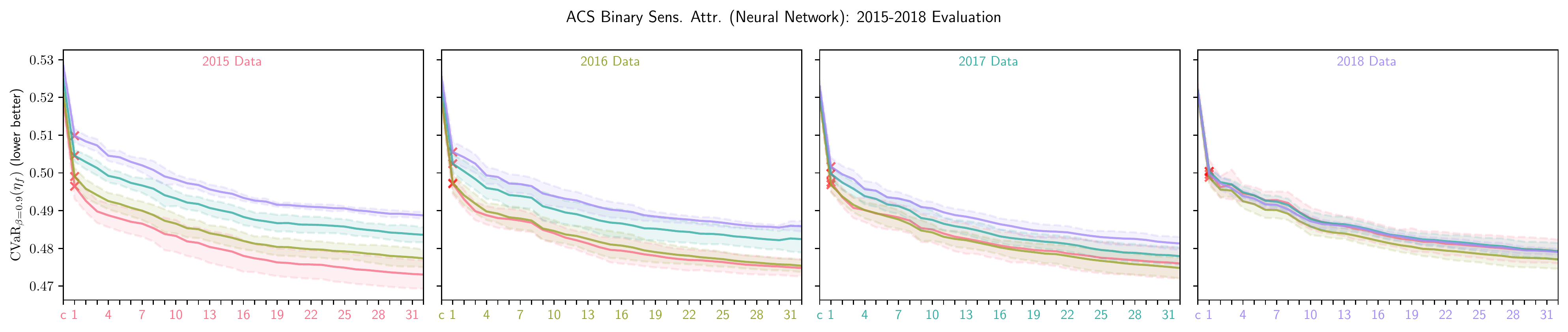}
    \includegraphics[width=\textwidth,trim={0, 6pt, 0, 0},clip]{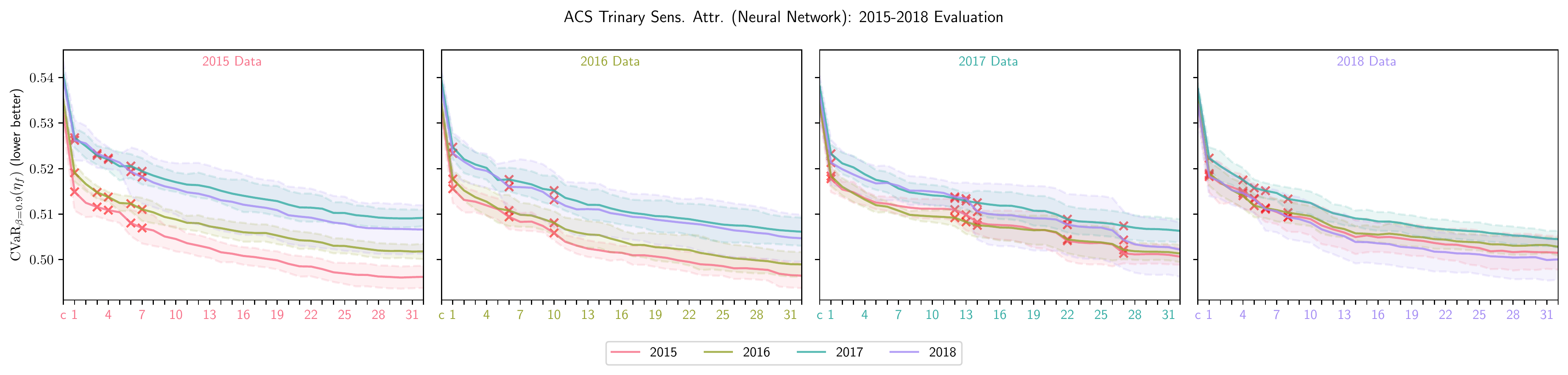}
   \vspace{-0.5cm}
    \caption{Neural Network black-box aggressive \( \cvar{} \) wrapper trained for ACS 2015 to 2018 datasets Each plot is trained on a different dataset year. Each curve colour, indicates the data being used to evaluate the wrapper.}
    \label{fig:mlp_yeareval_aggressive}
\end{sidewaysfigure}

To examine how \topdown is effected by distribution shift, we train various wrappers over multiple years of the ACS dataset. In particular, we train and evaluate \( \cvar{} \) wrappers over the ACS dataset from years 2015 to 2018. \cref{fig:rf_yeareval_conservative,fig:rf_yeareval_aggressive} report the \( \cvar{} \) values over the multiple years for the random forest (RF) black-box. \cref{fig:mlp_yeareval_conservative,fig:mlp_yeareval_aggressive} likewise reports corresponding results for neural network (NN) black-boxes.

In particular:
\begin{itemize}
    \item \cref{fig:rf_yeareval_conservative} presents the conservative update distribution shift evaluation using a RF black-box with \( B = 1 \) clipping on the ACS dataset over years 2015 to 2018.
    \item \cref{fig:rf_yeareval_aggressive} presents the aggressive update distribution shift evaluation using a RF black-box with \( B = 1 \) clipping on the ACS dataset over years 2015 to 2018.
    \item \cref{fig:mlp_yeareval_conservative} presents the conservative update distribution shift evaluation using a MLP black-box with \( B = 1 \) clipping on the ACS dataset over years 2015 to 2018.
    \item \cref{fig:mlp_yeareval_aggressive} presents the aggressive update distribution shift evaluation using a MLP black-box with \( B = 1 \) clipping on the ACS dataset over years 2015 to 2018.
\end{itemize}

As the ACS dataset consists of census data, one could expect that prior years of the data will be (somewhat) represented in subsequent years of the data. This is further emphasised in the plots, where curves become more closely group together as the training year used to train \topdown~increases, \emph{i.e.}, 2018 containing enough example which are indicative of prior years' distributions. Unsurprisingly, we can see that most circumstances the largest decrease in \( \cvar{} \) (mostly) comes from instances where the data matches the evaluation. \emph{i.e.,} the 2015 curve in (top) \cref{fig:rf_yeareval_conservative}.
Nevertheless, we can see that despite the training data, all evaluation curves decrease from their initial values in all plots; where a slight 'break' in `monotonicity' occurs in some instances of miss-matching data --- most prominently in (top) \cref{fig:rf_yeareval_conservative} for the 2015 plot around 21 boosting iterations. We also remark, perhaps surprisingly, that there is no crossing between curves (\textit{e.g.} as could be expected for the test-2015 and test-2016 curves on training from 2016's data in Figure \ref{fig:rf_yeareval_conservative}), but if test-2015 remains best, we also remark that it does become slightly worse for train-2016 while test-2016 expectedly improves with train-2016 compared to train-2015. Ultimately, all test-* curves converge to a 'midway baseline' on train-2018.

In general, there is little change when comparing the two different black-boxes. The only consist pattern in comparison is that the NN approaches start and end with a smaller \( \cvar{} \) value than their RF counter parts. When comparing binary versus trinary results, there is a distinct larger spread between evaluation curves (between each year within a plot) for the trinary counterparts. This is expected as in the trinary sensitive attribute modality, \( \cvar{} \) is sensitive to additional partitions of the dataset. The spread is further strengthened as the final \( \alpha \)-tree in \topdown often does not provide an \( \alpha \)-correction for all subgroups, \emph{i.e.}, at least one subgroup is not changed by the \( \alpha \)-tree with \( \alpha = 1 \). When comparing conservative versus aggressive approaches, it can also be seen that there is a larger spread between evaluation curves for the aggressive variant.

\clearpage

\clearpage

\clearpage

\clearpage

\clearpage
\newpage
\section{High Clip Value} \label{app-exp-clip}

In this section, we consider a higher clipping value than that used in other experiments. In other sections, we consider a \( B = 1 \) clipping value which results in posterior restricted between roughly \( [0.27, 0.73] \). Although this clipping seems harsh, from the prior experiments one can see that \topdown provides a lot of improvement across all fairness criterion (and we will see \( B = 1\) allows \topdown to improve beyond optimization for a large clip value).

\begin{sidewaysfigure}[!ht]
    \centering
    \includegraphics[width=\textwidth,trim={0, 6pt, 0, 0},clip]{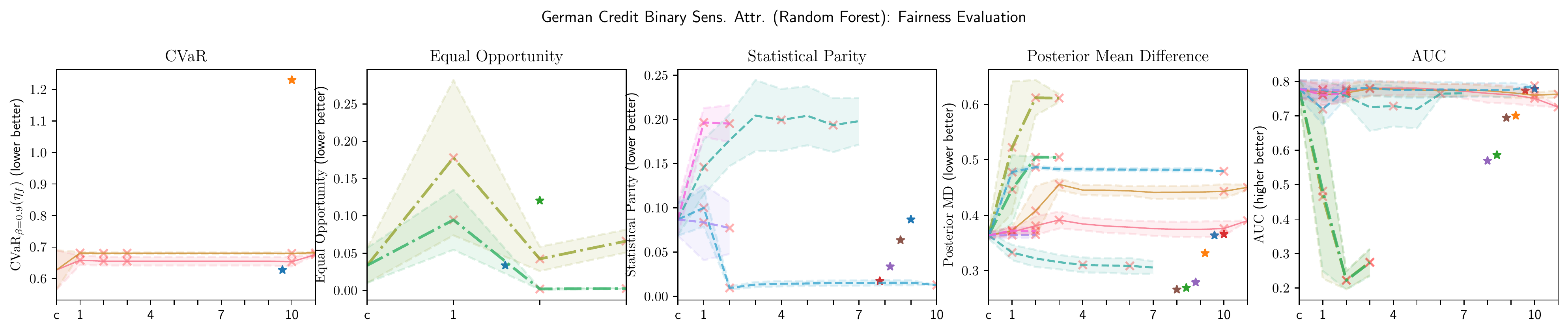}
    \includegraphics[width=\textwidth,trim={0, 6pt, 0, 0},clip]{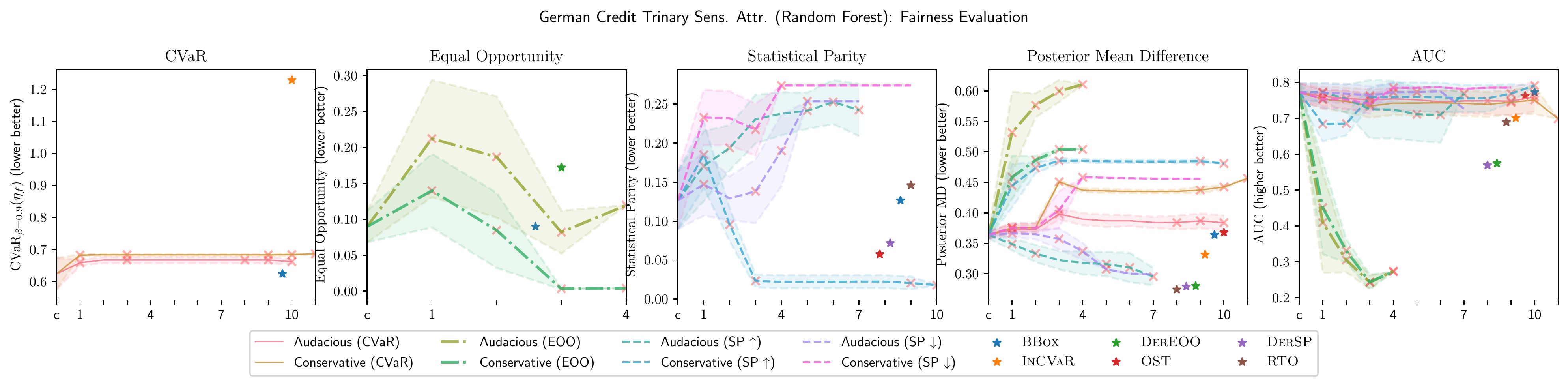}
   \vspace{-0.5cm}
    \caption{RF with \( B = 3 \) \topdown~optimized for different fairness models evaluated on German Credit with binary (up) and trinary (down) sensitive attributes. Crosses denote when a subgroup's \( \alpha \)-tree is initiated (over any fold). The shade depicts \( \pm \) a standard deviation from the mean. However, this disappears in the case where other folds stop early.}
    \label{fig:rf3_german_fairness}
\end{sidewaysfigure}

\begin{sidewaysfigure}[!ht]
    \centering
    \includegraphics[width=\textwidth,trim={0, 6pt, 0, 0},clip]{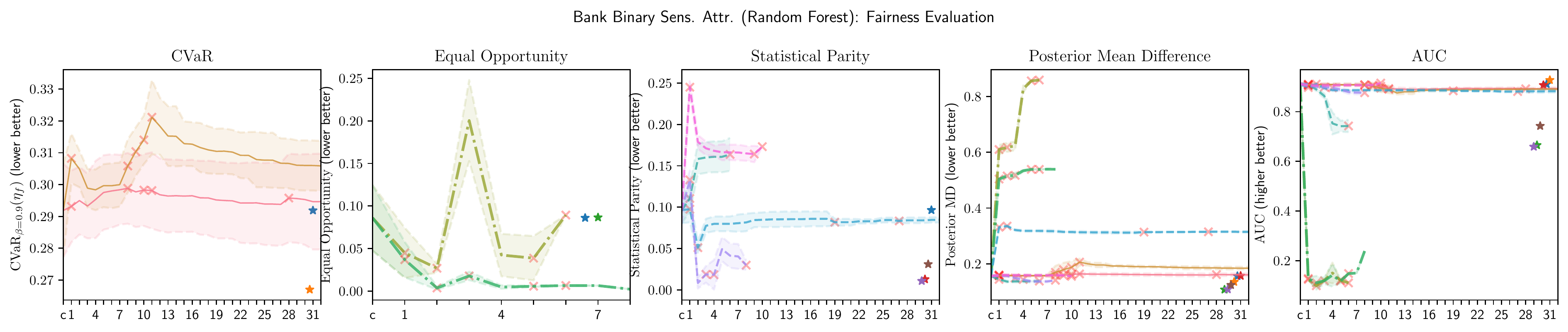}
    \includegraphics[width=\textwidth,trim={0, 6pt, 0, 0},clip]{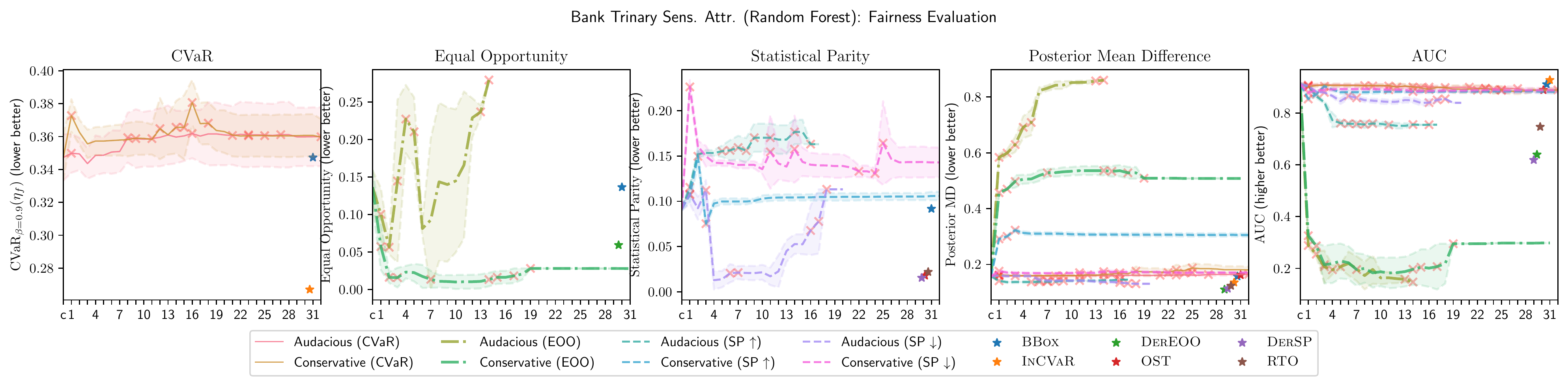}
   \vspace{-0.5cm}
    \caption{RF with \( B = 3 \) \topdown~optimized for different fairness models evaluated on Bank with binary (up) and trinary (down) sensitive attributes. Crosses denote when a subgroup's \( \alpha \)-tree is initiated (over any fold). The shade depicts \( \pm \) a standard deviation from the mean. However, this disappears in the case where other folds stop early.}
    \label{fig:rf3_bank_fairness}
\end{sidewaysfigure}

\begin{sidewaysfigure}[!ht]
    \centering
    \includegraphics[width=\textwidth,trim={0, 6pt, 0, 0},clip]{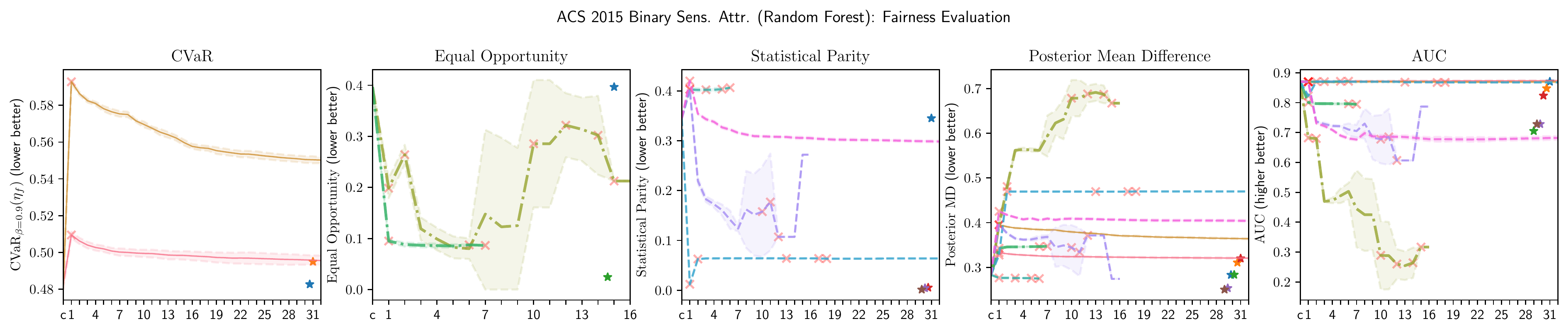}
    \includegraphics[width=\textwidth,trim={0, 6pt, 0, 0},clip]{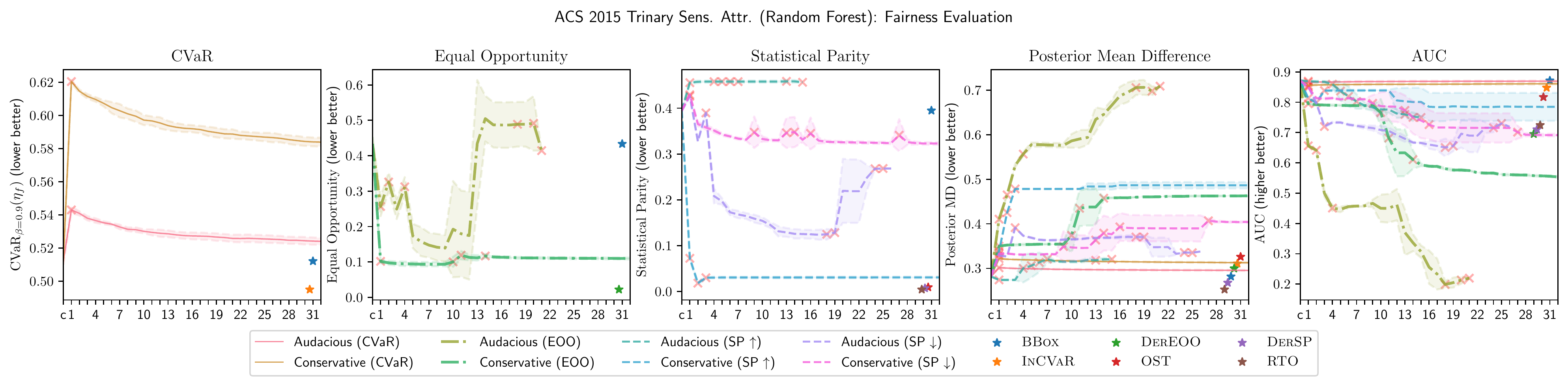}
   \vspace{-0.5cm}
    \caption{RF with \( B = 3 \) \topdown~optimized for different fairness models evaluated on Bank with binary (up) and trinary (down) sensitive attributes. Crosses denote when a subgroup's \( \alpha \)-tree is initiated (over any fold). The shade depicts \( \pm \) a standard deviation from the mean. However, this disappears in the case where other folds stop early.}
    \label{fig:rf3_acs_fairness}
\end{sidewaysfigure}

\begin{sidewaysfigure}[!ht]
    \centering
    \includegraphics[width=\textwidth,trim={0, 6pt, 0, 0},clip]{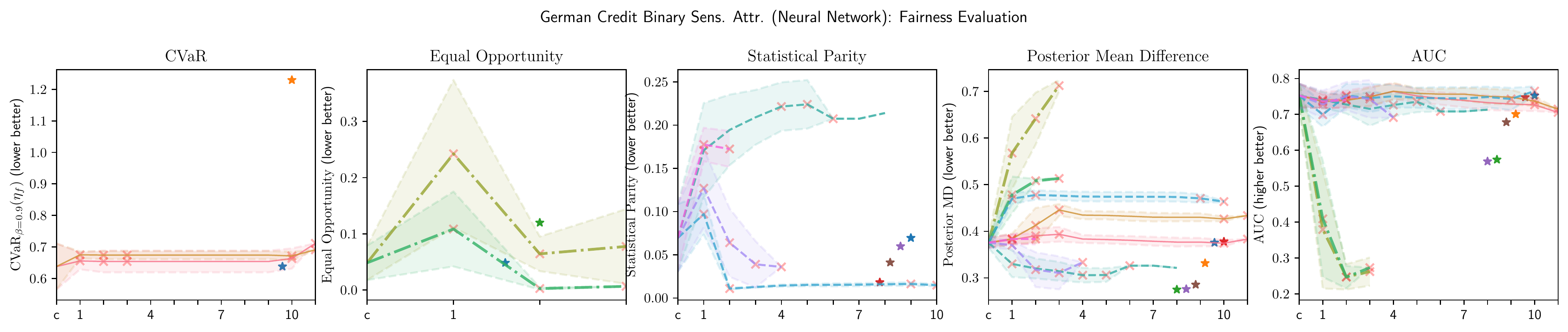}
    \includegraphics[width=\textwidth,trim={0, 6pt, 0, 0},clip]{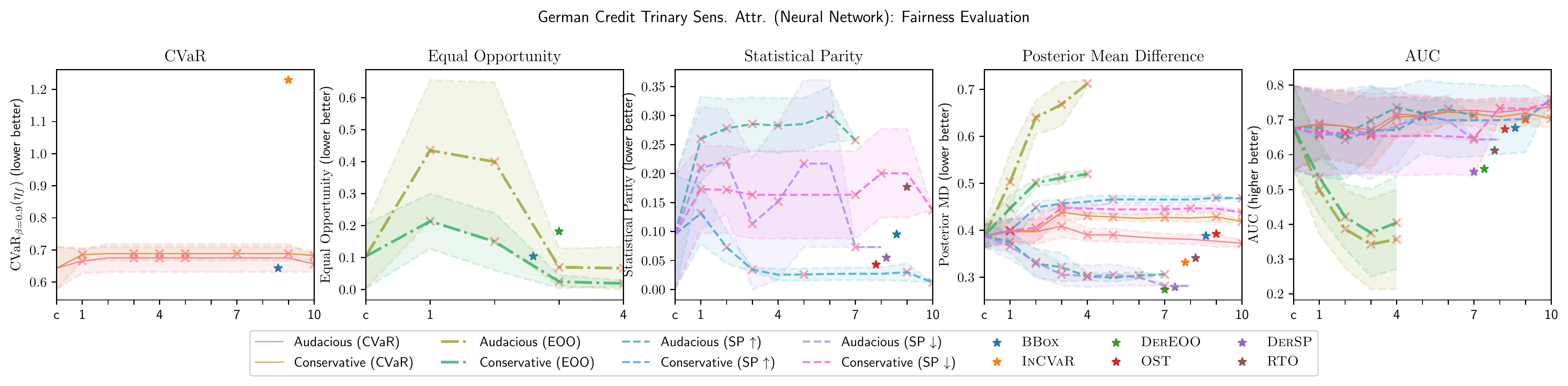}
   \vspace{-0.5cm}
    \caption{MLP with \( B = 3 \) \topdown~optimized for different fairness models evaluated on German Credit with binary (up) and trinary (down) sensitive attributes. Crosses denote when a subgroup's \( \alpha \)-tree is initiated (over any fold). The shade depicts \( \pm \) a standard deviation from the mean. However, this disappears in the case where other folds stop early.}
    \label{fig:mlp3_german_fairness}
\end{sidewaysfigure}

\begin{sidewaysfigure}[!ht]
    \centering
    \includegraphics[width=\textwidth,trim={0, 6pt, 0, 0},clip]{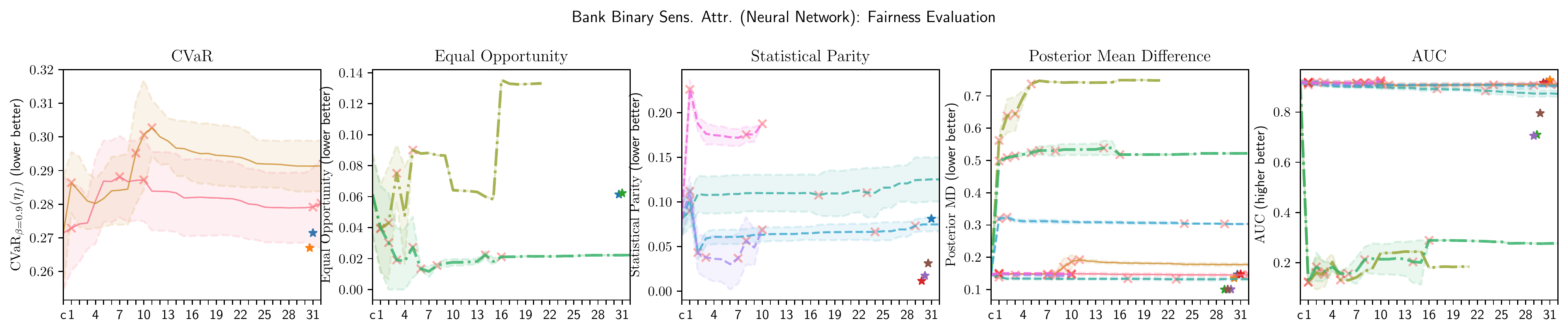}
    \includegraphics[width=\textwidth,trim={0, 6pt, 0, 0},clip]{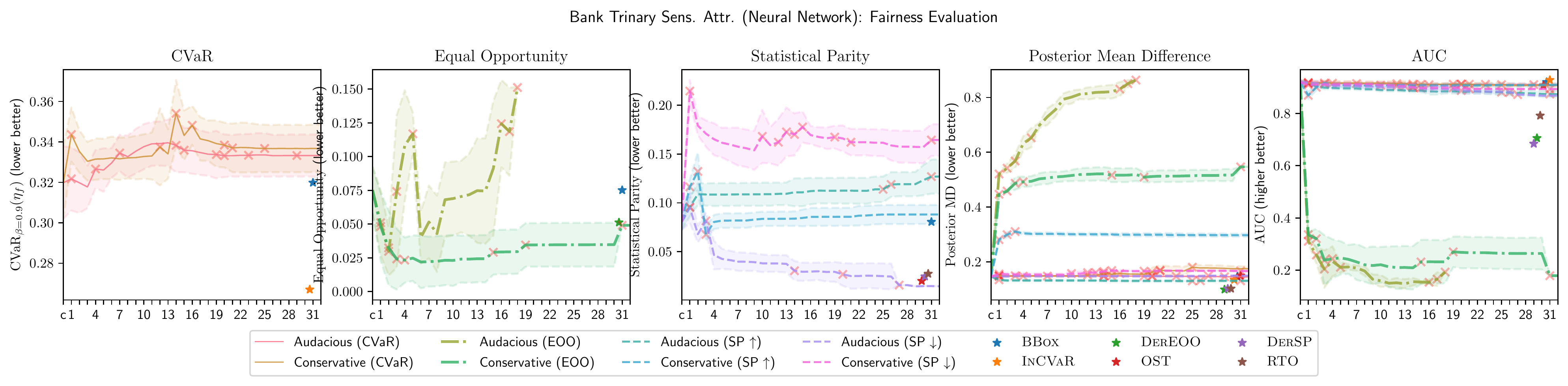}
   \vspace{-0.5cm}
    \caption{MLP with \( B = 3 \) \topdown~optimized for different fairness models evaluated on Bank with binary (up) and trinary (down) sensitive attributes. Crosses denote when a subgroup's \( \alpha \)-tree is initiated (over any fold). The shade depicts \( \pm \) a standard deviation from the mean. However, this disappears in the case where other folds stop early.}
    \label{fig:mlp3_bank_fairness}
\end{sidewaysfigure}

\begin{sidewaysfigure}[!ht]
    \centering
    \includegraphics[width=\textwidth,trim={0, 6pt, 0, 0},clip]{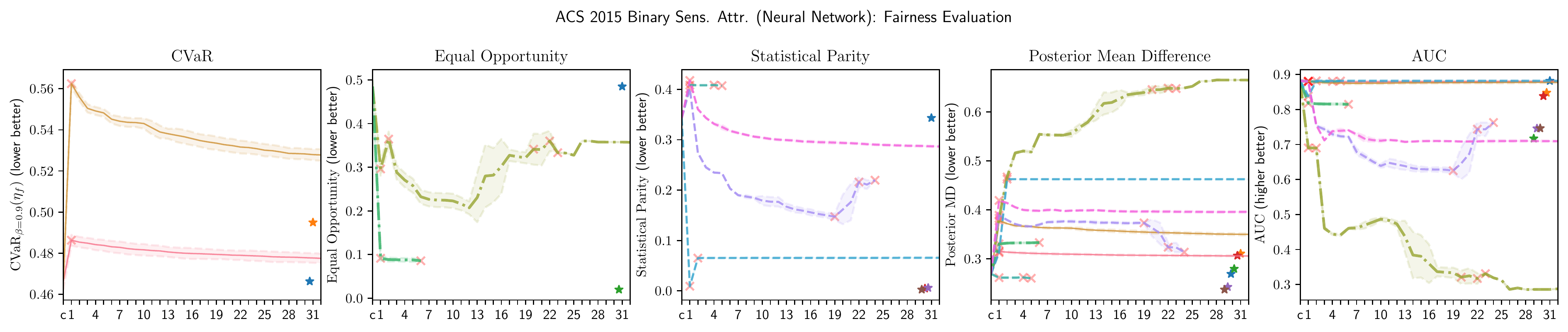}
    \includegraphics[width=\textwidth,trim={0, 6pt, 0, 0},clip]{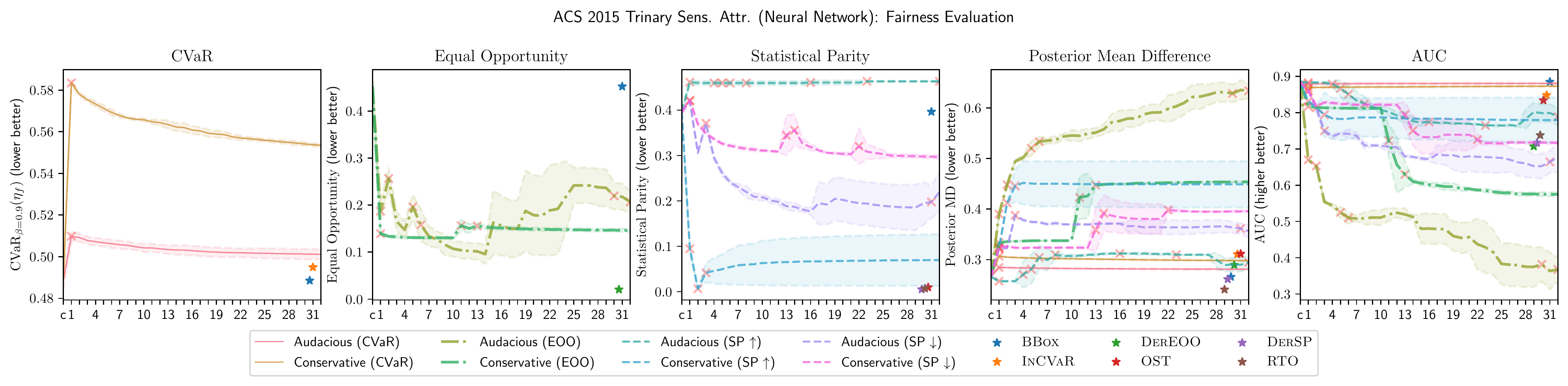}
   \vspace{-0.5cm}
    \caption{MLP with \( B = 3 \) \topdown~optimized for different fairness models evaluated on Bank with binary (up) and trinary (down) sensitive attributes. Crosses denote when a subgroup's \( \alpha \)-tree is initiated (over any fold). The shade depicts \( \pm \) a standard deviation from the mean. However, this disappears in the case where other folds stop early.}
    \label{fig:mlp3_acs_fairness}
\end{sidewaysfigure}

We will now consider \topdown experiments which correspond to evaluation over \( \cvar{} \), \EOO, and \SP criterion with clipping \( B = 3 \) (as discussed in theory sections of the main text). This restricts the posterior to be between roughly \( [0.05, 0.95] \). \cref{fig:rf3_german_fairness,fig:rf3_bank_fairness,fig:rf3_acs_fairness} presents RF plots over German, Bank, and ACS datasets; and \cref{fig:mlp3_german_fairness,fig:mlp3_bank_fairness,fig:mlp3_acs_fairness} presents equivalent MLP plots.

In particular:
\begin{itemize}
    \item \cref{fig:rf3_german_fairness} presents the evaluation using a RF black-box with \( B = 3 \) clipping on the German Credit dataset.
    \item \cref{fig:rf3_bank_fairness} presents the evaluation using a RF black-box with \( B = 3 \) clipping on the Bank dataset.
    \item \cref{fig:rf3_acs_fairness} presents the evaluation using a RF black-box with \( B = 3 \) clipping on the ACS dataset.
    \item \cref{fig:mlp3_german_fairness} presents the evaluation using a NN black-box with \( B = 3 \) clipping on the German Credit dataset.
    \item \cref{fig:mlp3_bank_fairness} presents the evaluation using a NN black-box with \( B = 3 \) clipping on the Bank dataset.
    \item \cref{fig:mlp3_acs_fairness} presents the evaluation using a NN black-box with \( B = 3 \) clipping on the ACS dataset.
\end{itemize}

In general, there is only a slight difference between the RF and MLP plots in this clipping setting.

We focus on the RF ACS plot of the higher clipping value, \cref{fig:rf3_acs_fairness}. The most striking issue is that the minimization of \( \cvar{} \) is a lot worse than when using clipping \( B = 1 \). In particular, \basebb (which in \cref{fig:rf3_acs_fairness} has \( B = 3 \)) is not beaten by the final wrapped classifier produced by either update of \topdown. However, for \EOO and \SP there is still a reduction in criterion, although a lower reduction for some cases, \emph{i.e.}, conservative \EOO. It is unsurprising that \( \cvar{} \) is more difficult to optimize in this case as the black-box would be closer to an optimal accuracy / cross-entropy value without larger clipping. As a result, \( \cvar{} \) would be more difficult to improve on as it depends on subgroup / partition cross-entropy. In particular, the large spike in the first iteration of boosting is striking. This comes from the fact that we are no directly minimizing a partition's cross-entropy directly, but an upper-bound, where the theory specifies that the upper-bound requires that the original black-box is already an \( \alpha \)-tree with correct corrections. However, as the the original black-box is not an \( \alpha \)-tree with correction specified by the update, the initial update can cause an increase in the \( \cvar{} \) (which appears to be more common with higher clipping values).

Despite the initial ``jump" and in-ability to recover, let us compare the \( B = 3 \) plot to the original \( B = 1 \) RF \topdown plot given in \cref{fig:rf_acs_fairness}. From comparing the results, one can see that the final boosting iteration for the \( B = 1 \) aggressive updates beats the \( B = 3 \) black-box classifiers. Thus, even when comparing against \( \cvar{} \) which is highly influenced by accuracy (thus a higher clipping value is desired), a smaller clipping value resulting in a more clipped black-box posterior is potentially more useful in \( \cvar{} \) \topdown. If one looks at the conservative curves in \cref{fig:rf_acs_fairness}, these do not beat the \( B = 3 \) black-box. This further strengthens the argument that the aggressive update is preferred in \( \cvar{} \) \topdown; and is further emphasized by the increase cap between curves with \( B = 3 \) black-boxes, as shown in \cref{fig:rf3_acs_fairness}.

\clearpage

\clearpage

\clearpage

\clearpage

\clearpage

\clearpage

\clearpage
\newpage
\section{Example Alpha-Tree} \label{app-exp-example}

\begin{sidewaysfigure}
    \centering
    \includegraphics[width=\textwidth]{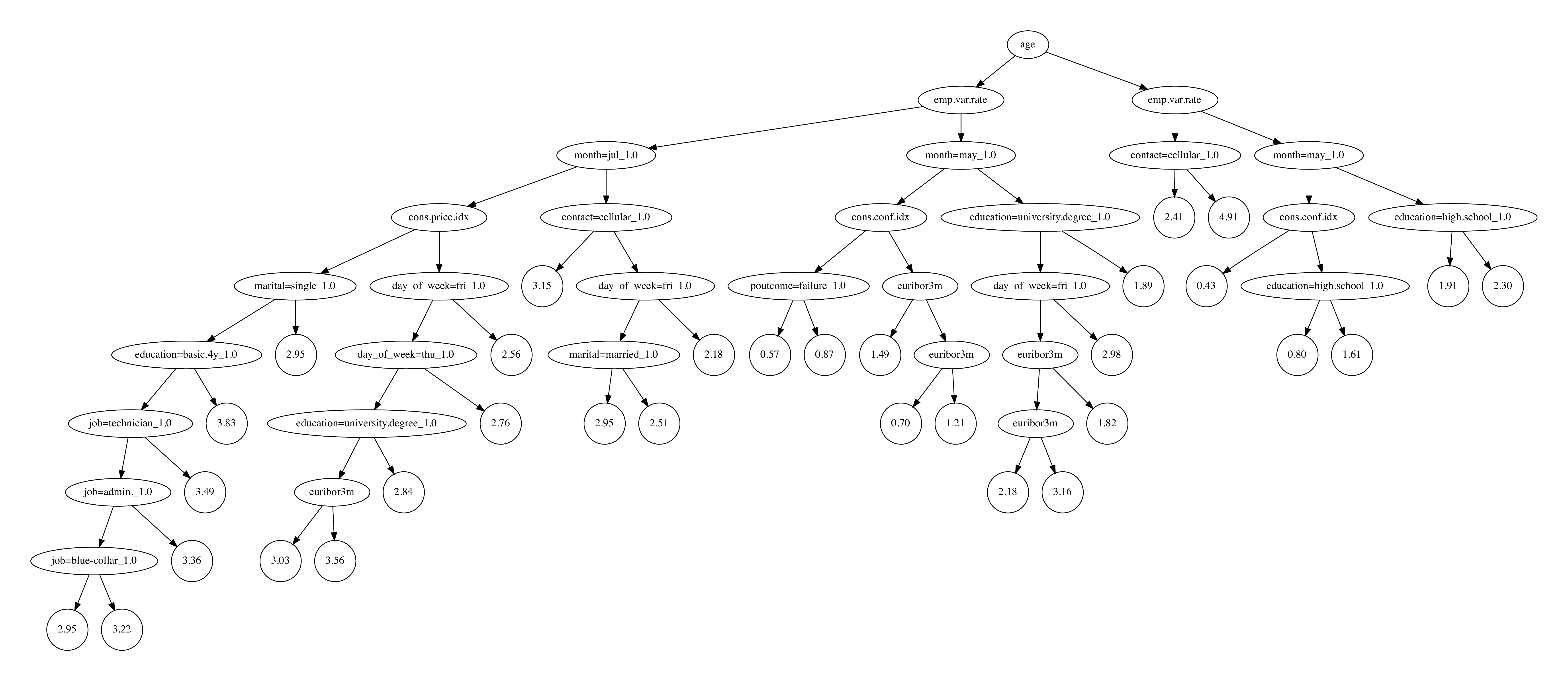}
    \caption{Example tree generated in the optimization of \topdown for \( \cvar{} \) in the Bank dataset with binary sensitive attributes.}
    \label{fig:example_alpha_tree}
\end{sidewaysfigure}

In this section, we provide an example of an \( \alpha \)-tree generated using \topdown. In particular, we look at one example from training \( \cvar{} \) \topdown on the Bank dataset with binary sensitive attributes. \cref{fig:example_alpha_tree} presents the example \( \alpha \)-tree. The tree contains information about the attributes in which splits are made and the \( \alpha \)-correction made at leaf nodes (and their induced partition). In the example, could note that the \( \alpha \) trees for modalities of the age sensitive attribute are imbalanced. The right tree is significantly smaller than the left. One could also note the high reliance on ``education" based attributes for determining partitions. These factors could be used to scrutinise the original blackbox; and eventually, even provide constraints on the growth of an \( \alpha \)-tree which would aim to avoid certain combinations of attribute. We leave these factors for future work.

\end{document}